\documentclass{sigkddExp}

\newcommand\red[1]{\textcolor{black}{#1}}

\newcommand\blue[1]{\textcolor{black}{#1}}
\newcommand\orange[1]{\textcolor{black}{#1}}
\newcommand\green[1]{\textcolor{black}{#1}}

\usepackage{authblk}
\usepackage{comment}
\usepackage{multirow}
\usepackage{color}
\usepackage{footmisc}
\usepackage{bm}
\usepackage{xcolor}
\usepackage{tabularx}
\usepackage{graphicx}
\usepackage{balance}
\usepackage[T1]{fontenc}
\usepackage[hyphens]{url}

\begin{document}
%

\title{Text Style Transfer: A Review and Experimental Evaluation}
%

%


\author[1]{\textbf{Zhiqiang Hu}}
\author[1]{\textbf{Roy Ka-Wei Lee}}
\author[2]{\textbf{Charu C. Aggarwal}}
\author[3]{\textbf{Aston Zhang}}

\affil[1]{Information Systems Technology and Design, Singapore University of Technology and Design}
\affil[2]{IBM T. J. Watson Research Center, Yorktown Heights, New York}
\affil[3]{Amazon Web Services AI, Washington}


\maketitle
\begin{abstract}
The stylistic properties of text have intrigued computational linguistics researchers in recent years. Specifically,  researchers have investigated the text style transfer task (TST), which aims to change the stylistic properties of the text while retaining its independent content of style. Over the last few years, many novel TST algorithms have been developed, while the industry has leveraged these algorithms to enable  exciting TST applications. The field of TST research has developed because of this symbiosis. This article aims to provide a comprehensive review of recent research efforts on text style transfer. More concretely, we create  a taxonomy to organize the TST models, and provide a comprehensive summary of the state of the art. We review existing evaluation methodologies for TST tasks and conduct a large-scale reproducibility study in which we experimentally benchmark 19 state-of-the-art TST algorithms on two publicly available datasets. Finally, we expand on current trends and provide new perspectives on the new and exciting developments in  the TST field.
\end{abstract}

\section{Introduction}
\label{sec:introduction}

The stylistic properties of text have intrigued linguistic researchers for a long time. Enkvist \cite{enkvist2016linguistic} opined that text style is a ``concept that is as common as it is elusive'' and suggested that \textit{style} may be described as a linguistic variation while preserving the conceptual content of the text. To give a practical example, the formality of text will vary across settings for similar content; examples include a  conversation with friends such as  ``let's hang out on Sunday afternoon!'', or a professional email such as  ``We will arrange a meeting on Sunday afternoon.''

In recent years, studies on text style have attracted the attention of not only the linguist, but also many computer science researchers. Specifically, computer science researchers are investigating the text style transfer task (TST), which is an increasingly popular branch of natural language generation \cite{gatt2018survey} that aims to change the stylistic properties of the text while maintaining its \red{independent content of style}. Previous TST studies have mainly attempted to perform TST with parallel corpora \cite{xu2012paraphrasing,jhamtani2017shakespearizing,carlson2018evaluating,shang2019semi,wang2019harnessing,jin2019imat,nikolov2018large,liao2018quase,xu2019formality}. The parallel corpora for TST consist of parallel sentences with different styles but the same semantics. For example, Xu et al. \cite{xu2012paraphrasing} proposed one of the first works to apply a phrase-based machine translation (PBMT) model to perform TST. They generated a parallel corpus of 30K sentence pairs by scraping the modern translations of Shakespearean plays and training a PBMT system to translate from modern English to Shakespearean English. For example, the Shakespearean sentence ``\textit{he slew thy kinsman}'' and the modern English sentence ``\textit{he killed your relative}'' are parallel sentences of different styles but similar semantics in the collected corpus. However, parallel data are scarce in many real-world TST applications, such as dialogue generation of different styles. The scarcity of parallel data motivated a new breed of TST algorithms that attempt to transfer text style without parallel data \cite{li2018delete,xu2018unpaired,zhang2018learning,sudhakar2019transforming,wu2019hierarchical,shen2017style,zhao2018adversarially,fu2018style,chen2018adversarial,logeswaran2018content,zhao2018language,lai2019multiple,john2019disentangled,park2019paraphrase,yin2019utilizing,yang2018unsupervised,hu2017toward,tian2018Structured,lampleSSDRB19,dai2019style,zhang2018shaped,jain2019unsupervised,mueller2017sequence,xu2019variational,wang2019controllable,liu2020revision,luo2019dual,gong2019reinforcement,He2020A}.

\red{This survey aims to thoroughly review the literature on the advances in TST and present experimental comparisons of various algorithms.} It gives a panorama through which readers can quickly understand and step into the field of TST. It should be noted that the literature in the field is rather disparate and a unified comparison is critical to aid in understanding the strengths and weaknesses of various methods.  This survey lays the foundations for future innovations in TST and taps into the richness of this research area. To summarize, the key contributions of this survey are threefold: (i) we investigate, classify, and summarize recent advances in the field of TST; (ii) we present several evaluation methods and experimentally compare different TST algorithms; (iii) we discuss the challenges in this field and propose possible directions on how to address them in future works.

The organization of this paper is as follows: We begin our discussion on the related research areas that inspire the commonly used TST techniques in Section \ref{sec:related}. Section \ref{sec:definition} provides the preliminary information about TST. In Section \ref{sec:taxonomy}, we categorize and explain the existing TST algorithms. The methodologies for evaluating TST algorithms are presented in Section \ref{sec:evaluation}. In Section \ref{sec:experiment}, we present experiments on publicly available datasets to benchmark the existing TST algorithms. \red{Next, we explore and demonstrate some of the commercial applications of TST in Section \ref{sec:application}. We also discuss the ethical considerations of TST applications and algorithms in Section \ref{sec:ethics}.} In Section \ref{sec:challanges}, we outline the open issues in TST research and offer possible future TST research directions. Finally, we conclude in Section \ref{sec:conclusion}.

\section{Related Research Areas}
\label{sec:related}

TST finds its roots in the field of natural language generation and is a relatively new research area. Many of the earlier TST works are also heavily influenced by two related research areas: \textit{neural machine translation} \cite{cho2014learning,sutskever2014sequence,DBLP:journals/corr/BahdanauCB14} and \textit{neural style transfer}, i.e., transferring styles in images~\cite{jing2019neural,gatys2015neural}. We found that a substantial number of TST techniques were adapted from the common methods used in natural language generation, neural machine translation, and neural style transfer. Furthermore, some of the evaluation metrics used in TST are also ``inherited'' from the natural language generation and neural machine translation tasks. In this section, we will provide some background on natural language generation and briefly introduce these two related research areas. Specifically, we will highlight some of the common techniques and evaluation metrics that are transferred or adapted for the TST task.

\subsection{Natural Language Generation}
Natural language generation involves a wide range of natural language processing tasks that aim to generate coherent and semantically meaningful text from input data or machine representations. Such tasks include machine translation \cite{cho2014learning,brown1990statistical}, dialogue generation \cite{liu-etal-2020-impress,DBLP:conf/aaai/SerbanKTTZBC17}, text summarization \cite{hovy-lin-1998-automated,miao2016language,takase-okazaki-2019-positional,9257174}, paraphrase generation \cite{ma-etal-2018-query,NEURIPS2019_5e2b6675}, visual storytelling \cite{huang-etal-2016-visual,wang-etal-2018-metrics}, and text style transfer \cite{li2018delete,xu2018unpaired,zhang2018learning,sudhakar2019transforming,wu2019hierarchical,shen2017style}. Recently, academia and industry have actively advanced natural language generation research, especially with the wide-adoption of large pre-trained language models \cite{devlin-etal-2019-bert,DBLP:conf/nips/BrownMRSKDNSSAA20}.

Many tasks of natural language generation share common characteristics and goals. For example, most of these tasks aim to generate fluent texts with minimal grammatical errors, and the generated texts should contain specific intended content. TST inherits these common goals and adds another objective: generating text in specific styles. Hence, TST models often extend natural language generation techniques to manipulate the style attribute in texts. For example, generative adversarial networks (GANs) have been applied to generate realistic and natural sentences \cite{DBLP:conf/nips/GoodfellowPMXWOCB14,lamb2016professor,hu2018unifying}.  Fu et~al.~\cite{fu2018style}
extended the GAN-based approach to disentangle the semantic and stylistic aspects of a text for TST.

The evolution of TST methods closely follows the advancements in natural language generation techniques. For instance, \cite{zhu2018texygen} introduced a benchmarking platform for text generation models named \textit{Texygen} which implemented a majority of text generation models and covered a set of metrics that evaluate the diversity, the quality, and the consistency of the generated texts. However, TST also shared similar limitations as other natural language generation tasks. Specifically, there are shortcomings in existing evaluation methods used to assess the output generated by the TST and natural language generation models \cite{mir2019evaluating,pang2019towards,celikyilmaz2020evaluation}. For example, there are currently no standard automatic evaluation metrics to assess the contextual quality or informativeness of generated texts for specific natural language generation tasks. There is also little consensus on how human evaluations should be conducted.

\subsection{Controllable Text Generation}
Controllable text generation, which offers the possibility of controlling various aspects of generated textual content, has drawn much attention in the natural language generation research community in recent years. The aspects commonly controlled include context~\cite{sordoni-etal-2015-neural,voita-etal-2018-context,serban2016building,xing2018hierarchical}, topic~\cite{dziri-etal-2019-augmenting,feng2018topic,ijcai2018-619,xing2017topic}, emotion~\cite{fu2018style,kong2019adversarial,DBLP:journals/inffus/SunLWLT20,zhou2018emotional}, user preferences~\cite{li-etal-2016-persona,luan-etal-2017-multi,yang2018investigating,yang2017personalized}. Consequently, studies have also proposed controllable text generation techniques to control the stylistic proprieties of the text and perform TST. Techniques such as GAN \cite{goodfellow2020generative}, VAE \cite{kingma2013auto}, and Transformer \cite{vaswani2017attention}, which are employed in controllable text generation, have been been also used in TST models. We will discuss some of these techniques in more detail in Section~\ref{sec:taxonomy}.

\subsection{Neural Machine Translation}
Neural machine translation, a deep learning-based approach to machine translation, is a well-studied research area \cite{cho2014learning,sutskever2014sequence,DBLP:journals/corr/BahdanauCB14}. Unlike the traditional statistical machine translation techniques \cite{brown1990statistical,koehn2007moses}, neural machine translation can perform end-to-end training of a machine translation model without the need to deal with word alignments, translation rules, and complicated decoding algorithms. Both TST and neural machine translation are branches of natural language generation \cite{gatt2018survey}. Naturally, the two research areas share a few similarities. First, neural machine translation aims to change the language of a text sequence while preserving the content, and TST aims to modify the stylistic properties of a text sequence while also preserving the content. Second, most of the TST models have ``borrowed'' the most commonly used neural machine translation technique: the sequence-to-sequence encoder-decoder model \cite{DBLP:journals/corr/BahdanauCB14,cho2014learning,sutskever2014sequence}. Other TST studies have also adopted the back-translation technique originally proposed for neural machine translation \cite{sennrich2015improving} to transfer styles in texts \cite{prabhumoye2018style,dos2018fighting}. For example, Prabhumoye et al. \cite{prabhumoye2018style} used a back-translation model to extract content features in text and then generate text in different styles using the extracted content features and multiple decoders. Third, TST works have inherited some of the automatic quantitative evaluation metrics that were generally used to evaluate the neural machine translation task. For example, the bilingual evaluation understudy (BLEU) metric \cite{papineni2002bleu}, which is used to evaluate the quality of machine-translated text by computing the modified n-gram precision between the generated and reference text, is also widely used in the TST task. 

Despite the close similarities between TST and neural machine translation, it is also worth noting their subtle differences. The text attribute transferred in neural machine translation is the text language, which is easily observed. TST, on the other hand, focuses on transferring text style that is abstract and subtle. Although both neural machine translation and TST are also concerned with preserving the semantics of the original text and ensuring the fluency of the transferred text, most neural machine translation methods are independent of style information. 
These differences motivated TST researchers to explore other techniques such as controllable generation \cite{hu2017toward,tian2018Structured,lampleSSDRB19,dai2019style,zhang2018shaped,zhang2018shaped,jain2019unsupervised} to ensure that the style of a text sequence is modified during TST. The need to evaluate whether the style is effectively transferred has also encouraged the development of evaluation metrics to evaluate TST models \cite{fu2018style,mir2019evaluating,pang2019unsupervised}. 

\subsection{Neural Style Transfer}
Gatys et al. \cite{gatys2015neural} first explored using a convolutional neural network (CNN) to extract content and style characteristics from images separately. Their experimental results demonstrated that CNNs could extract content information from an arbitrary photograph and style information from a well-known artwork. Based on this finding,  Gatys et al. \cite{gatys2015neural} further experimented with the idea of exploiting CNN feature activation to recombine the content of a given photo with the style of famous artwork. Their proposed algorithms' underlying idea is to minimize the loss between the synthesized images' CNN latent representation and the desired CNN feature distributions, which is the combination of the photos' content feature representation and artworks' style feature representation. Interestingly, the algorithm also does not have any explicit restriction on the style of images and does not require ground truth for training. The seminal work of Gatys et al. opened the new field of neural style transfer, which is the process of using neural networks to render content images in different styles \cite{jing2019neural}. 

The burgeoning research in the emerging field of neural style transfer has attracted wide attention from both academia and industry. In particular, researchers in natural language processing are motivated to adopt similar strategies to implicitly disentangle content and style features in text and transfer the learned style features to another textual content \cite{prabhumoye2018style,zhang2019machine,shen2017style,zhao2018adversarially,fu2018style,chen2018adversarial,logeswaran2018content,yin2019utilizing,zhao2018language,lai2019multiple,john2019disentangled,park2019paraphrase,yin2019utilizing,lai2019multiple,yang2018unsupervised,hu2017toward,tian2018Structured}. For example, Fu et al. \cite{fu2018style} proposed two TST models, which adopted an adversarial learning approach to implicitly disentangle content and style in text. The first method used multiple decoders for each type of style to generate texts of different styles from a common content embedding. In the second approach, the style embeddings are learned and augmented to a content embedding, and a single decoder is used to generate output in different styles. 

Although this line of TST works shares similar objectives as neural style transfer approaches, disentangling content and style in texts has proven to be much harder than in the case of images \cite{lampleSSDRB19}. First, the styles in images are easier to visualize and differentiate styles in two images in terms of patterns that can be easily modeled by a neural network. In contrast, text styles are somewhat more subtle, making it challenging to differentiate and define styles in two given pieces of text. Second, unlike the image's content and style, which are easily separated in the different CNN layers, the content and styles in the texts are tightly coupled and could not be easily separated even with the style labels. Third, in an image, arbitrarily changing a pixel may hardly alter the viewer's perception. However, randomly changing a word in a sentence will probably alter its meaning. This makes preserving the input content of text more difficult than for images. Hence, some of the recent TST works have proposed a new direction to transfer text style without disentanglement of texts' content and style \cite{lampleSSDRB19,dai2019style,zhang2018shaped,zhang2018shaped,jain2019unsupervised,mueller2017sequence,xu2019variational,wang2019controllable,liu2020revision,luo2019dual,gong2019reinforcement,He2020A}.

\section{Style and Notation Definition}
\label{sec:definition}

Before we dive into the details of this survey, we first introduce the basic terminology and concepts used throughout this survey. We will also describe the TST task formulation and summarize the notions commonly used in TST techniques.

\subsection{Definition of Style}
\red{We discuss the definitions of text style from two perspectives: (a) the definitions of text style discussed in linguistic studies and (b) the definitions adopted in text style transfer literature.}


\textbf{Linguistic View.} Style is an intuitive notion \red{that involves} the manner in which some semantics are delivered \cite{mcdonald-pustejovsky-1985-computational}. The semantics (or `\textit{content}' used in this survey) of a text refers to the subject matter or the argument that the author wants to deliver. Text style is the literary element that describes the ways the author uses language, including word choice, sentence structure, figurative language, and sentence arrangement, and works together to establish tone, image, and semantics in the text. The style indicates how the author describes events, objects, and ideas. Through style, the author can \red{provide} additional information that the reader can interpret and respond to. It is impossible to list all possible styles. Every speaker has an idiosyncratic set of techniques, often tailored to particular hearers, for using language to achieve his or her interpersonal goals \cite{hovy1987generating}.

\textbf{Style in Text Style Transfer.}
Unlike linguistics studies, which provide a more theoretical rule-based definition of text style, TST studies adopt a more data-driven approach \red{to define} text style. Specifically, TST studies often consider `\textit{style}' \red{as} text style attributes or labels dependent on the style-specific corpora. These style-specific corpora usually contain text style attributes that are not difficult for neural machine learning techniques to model. \red{For example, sentiment transfer tasks (that is, \textit{positive} and \textit{negative} as style attributes) and formality transfer (that is, \textit{formal} and \textit{informal} as style attributes)} are two of the most popular benchmarks for the evaluation of TST performance. We will provide a more comprehensive list \red{of the} evaluation tasks and available corpora in Section \ref{sec:evaluation}. However, \red{it is arguable that the sentiment of a text can be regarded as its style.} The main goal of TST is to modify the text's style while preserving its semantics. One could argue that sentiment transfer would have also modified the semantics of the sentence. However, from the perspective of the current deep learning methods for TST, most of the existing TST models in this survey can be applied to any datasets labeled with different stylistic attributes, including sentiment, formality, \red{gender, and political slant.}

\subsection{Task Formulation}

\begin{table*}[t]
\centering
\caption{Notation of each variable and its corresponding meaning.}
\label{tbl:notation}
\begin{tabular}{ll}
\hline
Not. & Meaning \\
\hline
$s$         & The source attribute value, e.g., the formal style\\
$t$         & The target attribute value different from $s$, e.g., the informal style \\
$\mathcal{A}$         & A predefined set of attribute values, $s,t \in \mathcal{A}$ \\
$\bm{x}$         & A sentence with the source attribute value $s$ \\
$\bm{x'}$        & The transferred sentence of $\bm{x}$ with the target attribute value $t$ \\
$\bm{X}$         & The corpus of sentences with different attribute values \\
$E$         & Encoder of a TST model \\
$G$         & Generator of a TST model \\
$D$         & style classifier or discriminator \\
$\Theta_E$  & Parameters of the encoder \\
$\Theta_G$  & Parameters of the generator \\
$\Theta_D$  & Parameters of the style classifier \\
$\bm{z}$    & Latent representation of text, i.e., $\bm{z} \triangleq E(\bm{x})$ \\
$\bm{a}$    & Latent representation of the attribute value in text \\
\hline
\end{tabular}
\end{table*}

The TST task aims to change the stylistic properties of any given text while preserving its style-independent content. 
The input includes a set of attributes $\mathcal{A}$ with text for each attribute in the corpus $\bm{X}$. 
For example, for the formality transfer task, there are two attributes: formal and informal. \red{The task consists of} taking the sentence $\bm{x}$ with the source attribute $s$ (e.g., formal) and generating the sentence $\bm{x'}$ with the target attribute $t$ (e.g., informal) while preserving the style-independent content. The style corpus $\bm{X}$ can be parallel or non-parallel. In the parallel setting, for each sentence with the source attribute $s$, a counterpart sentence with the same style-independent content with the target attribute $t$ is contained in $\bm{X}$. For the non-parallel setting, there is no alignment information among sentences with different attributes. Table \ref{tbl:notation} shows the notation used in this survey.

\section{A Taxonomy of Text Style Transfer Methods}
\label{sec:taxonomy}

This section first proposes a taxonomy to organize the most notable and promising advances in TST research in recent years. Then we discuss each category of TST models in greater detail.

\subsection{Categories of Text Style Transfer Models}
\begin{figure*}[t]
    \centering
    \includegraphics[width=\textwidth]{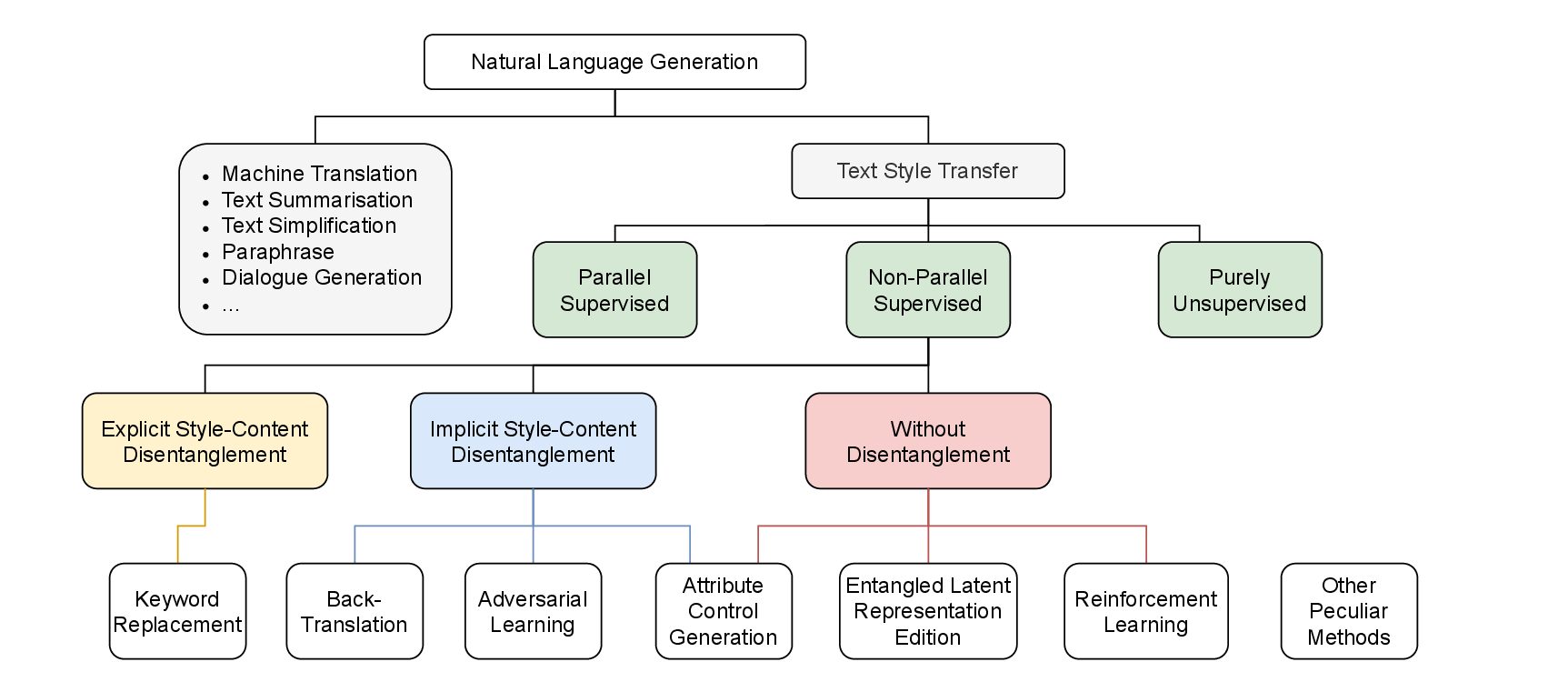}
    \caption{A taxonomy of text style transfer methods}
    \label{fig:taxonomy}
\end{figure*}

To provide an overview of this field, we classify the existing TST models based on types of (1) data settings, (2) strategies, and (3) techniques. Fig. \ref{fig:taxonomy} summarizes the taxonomy of the TST methods. We consider TST as one of the many natural language generation tasks. The TST techniques developed can be broadly categorized by the data settings used in training, i.e., parallel supervised, non-parallel supervised, and purely unsupervised. As most recent developments focus on designing non-parallel supervised TST techniques, we scope our survey to cover these techniques in greater detail. Specifically, we discuss the broad strategies applied by the non-parallel supervised TST techniques and the various purely unsupervised TST techniques

\subsubsection{Data Settings}
\label{sec:tax_datasetting}
We \red{generally classify existing TST studies} into three categories based on the data settings used for model training. 

\begin{itemize}
    \item \textbf{Parallel Supervised.} In this data setting, the TST models are trained with known pairs of text with different styles. Commonly, NMT methods such as {\em sequence-to-sequence} (Seq2Seq) models \cite{xu2012paraphrasing,cho2014learning,sutskever2014sequence,DBLP:journals/corr/BahdanauCB14} are applied to transfer the style of text. For example, Jhamtani et al. \cite{jhamtani2017shakespearizing} trained a Seq2Seq model with a pointer network on a parallel corpus and applied the model to translate modern English phrases to Shakespearean English. Details of the techniques applied on parallel datasets will be discussed in Section \ref{sec:tax-parallel}
    \item \textbf{Non-Parallel Supervised.} TST models in the non-parallel supervised setting aim to transfer the style of text without any knowledge of matching text pairs in different styles. Most of the existing TST studies fall into this category due to the scarcity of parallel datasets in real-world TST applications.
    \item \textbf{Purely Unsupervised.} In both parallel and non-parallel supervised data settings, the style labels are available to enable supervised training of the TST models. A more challenging setting is purely unsupervised where only an unlabeled text corpus is available, and the TST models need to be trained in an unsupervised fashion to perform text style transfer without any knowledge of style labels.
\end{itemize}

\subsubsection{Strategies}
\label{sec:strategy}
In order to perform TST in the \textit{non-parallel supervised} setting, existing studies have proposed to disentangle the style and content in text, which is a strategy commonly used in NST \cite{gatys2015neural}. 

The TST research community has widely discussed the possibility of disentangling style and content, and \red{it} remains an open research question. As discussed in Section \ref{sec:definition}, the semantics or content \red{of the} text refers to the subject matter or the argument that the author wants to communicate, while the text style describes the literary elements in which the text is communicated. The literary includes word choice, syntactic structures, figurative language, and sentence arrangement. While the content and style have precise definitions, it is unclear if it can disentangle text style and content. Intuitively, an author's stylistic choices could be influenced by the content communicated. For instance,  considering the sentiment of texts about food reviews, some food may be universally considered tastier than others, e.g., most people love pizza. Thus, a food review that discusses pizza is likely associated with positive sentiment. It becomes difficult to disentangle style from the content with confounding factors like this without some additional assumptions or domain knowledge.

Lample et al. \cite{lampleSSDRB19} also argued that it is challenging to perform style-content disentanglement and demonstrated that the adversarial method proposed by Fu et al. \cite{fu2018style} did not separate style and content successfully. \red{Nevertheless}, Lample et al. \cite{lampleSSDRB19}'s evaluation was only performed on the Fu et al. \cite{fu2018style}'s model, and it is still an ongoing research question on whether we can disentangle style from content and how we can achieve this goal. We encourage researchers in the TST community to explore the method to disentangle style and content more effectively and demonstrate whether the disentanglement has been achieved from different perspectives.

In this survey, we discuss three types of style-content disentanglement strategies and categorize the existing non-parallel supervised TST methods into one of these three strategies:

\begin{itemize}
    \item \textbf{Explicit Style-Content Disentanglement.} In this strategy, TST models adopt a straightforward text replacement approach to generate texts of a target style. For example, Li et al. \cite{li2018delete} first explicitly \red{identified} parts of the text that are associated with the original style and then \red{replaced} them with new phrases associated with the target style. The text with the new replaced phrases \red{was} then fed into a Seq2Seq model to generate a fluent text sequence in the target style. Details of the techniques applied to disentangle content and style will be explicitly discussed in Section \ref{sec:tax-explicit}.
    \item \textbf{Implicit Style-Content Disentanglement.} To disentangle style and content in text implicitly, TST models aim first to learn the latent representations of content and style of a given text sequence. Subsequently, the original text's content latent representation is combined with the latent representation of the target style to generate new text in the target style. Multiple techniques such as back-translation, adversarial learning, and controllable generation \cite{shen2017style,zhao2018adversarially,fu2018style,prabhumoye2018style,hu2017toward} have been proposed to disentangle \red{latent representations in content and style}. 
    \item \textbf{Without Style-Content Disentanglement.} Recent studies have suggested that it is difficult to judge the quality of text style and content disentanglement, and disentanglement is also unnecessary for TST \cite{lampleSSDRB19}. Therefore, more recent research explored performing TST without disentangling the text's style and content. Techniques such as adversarial learning, controllable generation, reinforcement learning, probabilistic modeling, and /red{pseudo-parallel corpus constructing} \cite{lampleSSDRB19,dai2019style,li2019domain,luo2019dual,He2020A} have been applied to perform TST without disentanglement of the text's content and style.
\end{itemize}

\subsubsection{Techniques}
Table~\ref{tbl:taxonomy} lists various types of techniques that are commonly used to perform TST. We organize them following the previously mentioned taxonomy and review each in detail in the following subsections. Additionally, we also present literature that adopted such techniques.


\begin{table*}[h]
\centering
\caption{Publications Based on Different Text Style Transfer Techniques}
\label{tbl:taxonomy}
\begin{tabular}{|c|c|c|p{2.5cm}|}
\hline
\textbf{Data Setting} & \textbf{\begin{tabular}[c]{@{}c@{}} Strategy (Content-Style \\ Disentanglement)\end{tabular}} & \textbf{Technique} & \multicolumn{1}{c|}{\textbf{Literature}} \\ \hline
Parallel Supervised & - & Sequence-to-Sequence  & \cite{jhamtani2017shakespearizing,carlson2018evaluating,shang2019semi,wang2019harnessing,jin2019imat,nikolov2018large,liao2018quase,xu2019formality,zhang2020parallel}\\ \hline
{Non-Parallel} Supervised & Explicit  & Explicit Style Keyword Replacement & \cite{li2018delete,xu2018unpaired,zhang2018learning,sudhakar2019transforming,wu2019hierarchical} \\ \cline{2-4} & {Implicit}  & Back-Translation & \cite{prabhumoye2018style,zhang2019machine} \\ \cline{3-4} 
 & & Adversarial Learning  &  \cite{shen2017style,zhao2018adversarially,fu2018style,chen2018adversarial,logeswaran2018content,yin2019utilizing,zhao2018language,yang2018unsupervised,lai2019multiple,john2019disentangled,park2019paraphrase}\\ \cline{3-4} 
 &  & Attribute Control Generation & \cite{hu2017toward,tian2018Structured} \\ \cline{3-4}  & & Other Peculiar Methods  &  \cite{DBLP:conf/acl/ChengMSMZLC20} \\ \cline{2-4} 
 & {Without} & Attribute Control Generation &  \cite{lampleSSDRB19,dai2019style,zhang2018shaped,jain2019unsupervised,zhou2020exploring} \\ \cline{3-4} 
 & & Entangled Latent Representation Edition & \cite{mueller2017sequence,xu2019variational,wang2019controllable,liu2020revision}\\ \cline{3-4} 
 & & Reinforcement Learning & \cite{luo2019dual,gong2019reinforcement} \\ \cline{3-4} 
 & & Other Peculiar Methods  &  \cite{He2020A,DBLP:conf/acl/ChengMSMZLC20,syed2020adapting} \\ \hline
 Purely Unsupervised & - & Purely Unsupervised  & \cite{radford2017learning,xu2019variational, shen2020educating,DBLP:conf/iclr/DathathriMLHFMY20} \\ \hline
\end{tabular}
\end{table*}

\subsection{Sequence-to-Sequence Model with Parallel Data}
\label{sec:tax-parallel}
The Sequence-to-Sequence model (Seq2Seq) \cite{xu2012paraphrasing,cho2014learning,sutskever2014sequence,DBLP:journals/corr/BahdanauCB14} based on the encoder-decoder architecture is \red{the core} to many natural language generation tasks \cite{gatt2018survey}, where TST is no exception. Generally, a Seq2Seq model is trained on a parallel corpus, where the text of the original style is fed into the encoder, and the decoder outputs the corresponding text according to the target style. \red{Various Seq2Seq TST models have been proposed that are trained on parallel datasets}. Jhamtani et al. \cite{jhamtani2017shakespearizing} extended the work of Xu et al. \cite{xu2012paraphrasing} by adding a pointer network \cite{vinyals2015pointer} to the Seq2Seq model to selectively copy word tokens from the input text directly to transfer modern English to Shakespearean English. The direct copy mechanism \red{was} motivated by the intuition that Shakespearean and modern English have significant overlap in vocabulary \red{and there are rare words, infrequent nouns, that are harder to generate by the Seq2Seq model}. Carlson \cite{carlson2018evaluating} proposed a Seq2Seq model with attention mechanisms and evaluated it with a collected parallel Bible-prose-style corpus.


However, applying the Seq2Seq approach to the TST task is challenging \red{due to} the shortage of parallel data. To this end, the researchers explored various data augmentation methods to enhance the parallel datasets used to train the Seq2Seq TST models. For instance, many existing works have attempted to generate larger pseudo-parallel datasets to train Seq2Seq TST models. A commonly-used approach is to generate a pseudo-parallel dataset is to retrieve pseudo-references of a given text from a large text corpus \cite{shang2019semi,jin2019imat,nikolov2018large,liao2018quase}. For instance, Liao et al. \cite{liao2018quase} generated a pseudo-parallel dataset by searching pseudo-references containing similar content but different sentiments from Yelp reviews. For example, given the sentence ``\textit{the food is terrible}'', a matching pseudo-reference could be ``\textit{the food is delicious}'', where ``\textit{the food}'' is similar content and the two sentences have very different review ratings, i.e., different sentiment. The generated pseudo-parallel dataset \red{was} subsequently used to train a Variational Autoencoder (VAE) \cite{kingma2013auto} based framework to perform TST.

Besides retrieving pseudo-references from existing large text corpus to construct a pseudo-parallel dataset,  Zhang et al. \cite{zhang2020parallel} had explored simply generating pseudo-references using a machine translation approach. Specifically, they first collected a set of potentially informal English sentences (e.g., online forums) and subsequently translated them into a pivot language (e.g., French), then translated them back into English. The back-translated English texts \red{were} further filtered using a pre-trained formality classifier, where the predicted formal text would be retained as the pseudo-reference for the informal texts.

The aforementioned data augmentation methods have demonstrated their effectiveness in improving parallel-supervised TST models' training. Furthermore, the generated pseudo-parallel datasets could also be leveraged to train and test other types of TST models. However, there remains a lack of systematic evaluation of the generated pseudo-parallel datasets; it is uncertain if the pseudo-references are indeed in a different style and \red{retain} the given text's content. Therefore, future TST  pseudo-parallel data augmentation methods will need to consider evaluating the augmented data.  

Another interesting data augmentation approach is leveraging and fine-tuning large pre-trained language models to perform TST. For instance, Wang et al. \cite{wang2019harnessing} fine-tuned pre-trained GPT-2 model~\cite{radford2019language} using the text formality transfer rules harnessed from analyzing the GYAFC parallel dataset~\cite{rao2018dear}. The fine-tuned GPT-2 model was subsequently used to transfer the formality of text (e.g., informal to formal text). This is a promising research direction, as language models capture many facets of language relevant for many downstream natural language processing tasks, and they could be pre-trained with a large text corpus. Future TST works may consider designing fine-tuning approaches that could better adapt the pre-trained language models for the TST task.

\subsection{Explicit Style Keyword Replacement}
\label{sec:tax-explicit}
Text style attributes, such as sentiment, are often marked by distinctive keywords and phrases. For example, words such as ``nice'' and ``good'' would infer positive sentiment, while words \red{such as} ``bad'' and ``nasty'' would infer negative sentiment. Therefore, an intuitive solution to transfer a text's sentiment would be to replace the specific keywords and phrases associated with specific sentiments. Motivated by this simple intuition, researchers have proposed TST methods that explicitly disentangle content and style in text by replacing keywords attributed to a specific style \cite{li2018delete,xu2018unpaired,zhang2018learning,sudhakar2019transforming,wu2019hierarchical}.

\begin{figure*}[h]
    \centering
    \includegraphics[scale = 0.6]{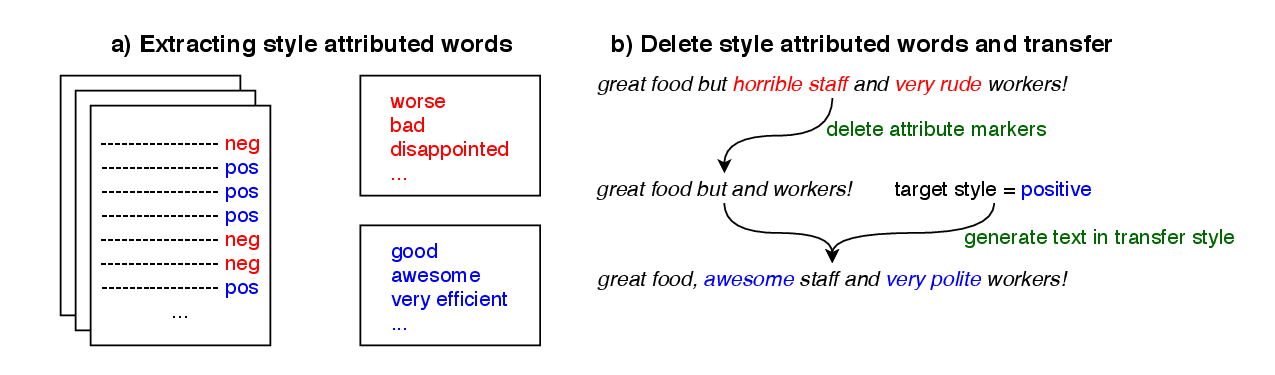}
    \caption{An overview of the \textit{Delete-Retreive-Generate} framework proposed by Li et al. \protect\cite{li2018delete}}
    \label{fig:explicit}
\end{figure*}

A representative work of the explicit style keyword replacement approach is the \textit{Delete-Retrieve-Generate} model~\cite{li2018delete}. Fig.~\ref{fig:explicit} provides an overview of the \textit{Delete-Retrieve-Generate} method. Given a text in source style (e.g., text with negative sentiment), the model first identifies the style-attributed words such as ``\textit{horrible, very rude}'' in the text by computing the relative frequency of each word. The underlying intuition is that the style-attributed words are likely to be frequently used in sentences of a specific style. Next, the model removes the style-attributed words from the text. The resulting text at this stage is assumed to contain only the content information such as ``\textit{food, staff}''. Subsequently, the model retrieves a reference text similar to the source text with only content-related words. The reference text would be retrieved from the target style corpus (i.e., text corpus with positive sentiment). The model then extracts the style-attributed words in the reference text using a similar approach. Finally, the style-attributed words of the retrieved reference text are combined with the content words of the source text to generate a text in target style using a rule-based approach or a Seq2Seq model. 

Existing explicit style keyword replacement TST methods adopted a similar framework and focused on innovating the approach to identify and replace the style-attributed keywords. For example, Transformer~\cite{vaswani2017attention} and deep learning text classifiers with an attention mechanism~\cite{zhang2018learning} have been explored to identify and replace \red{keywords attributed to style}~\cite{sudhakar2019transforming,wu2019mask,LeeftinkS19}. These studies have shown that the advancement of the style-attributed keyword replacement methods would improve the TST task's performance. \red{Recent work} has also combined explicit style keyword replacement with cycled reinforcement learning to iteratively replace style-attributed keywords while maintaining the content in the text \cite{xu2018unpaired,wu2019hierarchical}.

The strength of the explicit style keyword replacement TST approach lies in its simplicity; the models are relatively less complex with a shorter training time. Furthermore, \red{the replacement of keywords and phrases also provides some explainability to the TST model}; we can explicitly observe which part of the text is modified to transfer its style. However, this approach has several weaknesses. First, explicit style keyword replacement TST methods are mostly restricted to transferring the text's sentiment and cannot be applied to transfer other text style attributes such as formality. This is due to the nature of the sentiment transfer task, where particular lexicons often encode sentiment information, and replacing these lexicons is a viable approach to modify the text's sentiment. Transferring text's formality goes beyond simple keywords or phrases replacement because its formality could be encoded in its syntax. Second, it is challenging to apply explicit style keyword replacement TST to transfer beyond polar text attributes, i.e., positive to negative sentiment and vice versa.

\subsection{Adversarial Learning}
\label{sec:tax-adversarial}

To overcome the limitations of explicit style keyword replacement methods, researchers have investigated another research direction to disentangle text's content and style information implicitly to perform TST. Adversarial learning is \red{a} popular technique applied in many implicit content-style disentanglement TST methods~\cite{shen2017style,zhao2018adversarially,fu2018style,chen2018adversarial,logeswaran2018content,yin2019utilizing,zhao2018language,lai2019multiple,john2019disentangled,park2019paraphrase,yang2018unsupervised}. Broadly, these TST methods apply adversarial learning for two purposes: (a) generate text output that is indistinguishable from real data~\cite{shen2017style,logeswaran2018content,tian2018Structured,chen2018adversarial,zhao2018language,yin2019utilizing,park2019paraphrase}; (b) remove the style attributes in the latent representation of text~\cite{zhao2018adversarially,fu2018style,yang2018unsupervised, lai2019multiple,john2019disentangled}. A representative work for later purposes is the seminal framework proposed by Fu et al. \cite{fu2018style}.

\begin{figure*}[h]
    \centering
    \includegraphics[scale = 0.6]{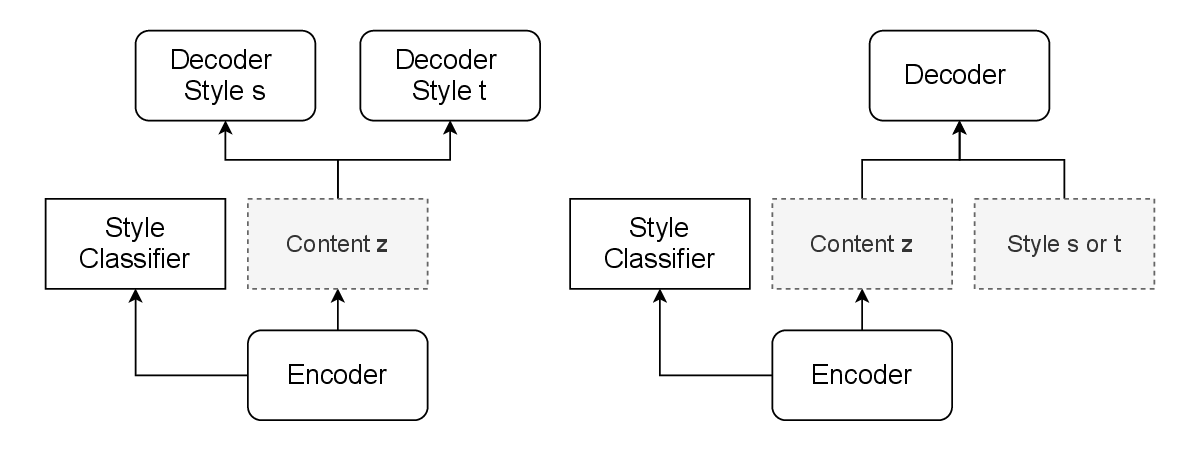}
    \caption{Two common adversarial-learning-based TST models: multi-decoder (left) and style-embedding (right) proposed by Fu et al. \protect\cite{fu2018style}. \red{The content representation \textit{z} is the output of the encoder}. The style classifier aims at distinguishing the style of the input. An adversarial network is used to ensure that content \textit{z} does not have the style representation. In multi-decoder, multiple decoders are used to generate the text in specific styles. In style embedding, content representation \textit{z} and style-embedding \textit{s} are concatenated and fed into the decoder.}
    \label{fig:adv_learning}
\end{figure*}

Fig. \ref{fig:adv_learning} illustrates two models included in the \red{framework proposed by et al. \cite{fu2018style}}. In both models, an encoder is trained to generate intermediate latent representation of input text sequence $\bm{X}=(\bm{x}_1, \cdots, \bm{x}_T)$ of length $T$. An adversarial network is used to separate \red{the representation $\bm{z}$ of content from the style}. The adversarial network is composed of two main components. The first component aims to \red{classify the input style} $\bm{x}$ given the representation learned by the encoder. The loss function to minimize is the negative log-likelihood of the style labels in the training data:
\begin{align*}
    L_{\text{adv1}}(\Theta_D) = -\sum^M_{i=1}\log p(l_i|\mathrm{Encoder}(\bm{x}_i;\Theta_E);\Theta_D),
\end{align*}
where $\Theta_D$ and $\Theta_E$ are the parameters of the classifier and encoder, respectively. $M$ denotes the size of the training dataset, and $l_i$ refers to the style label. The second component aims to make the classifier unable to identify the style of input $\bm{x}$ by maximizing the entropy (i.e., minimizing the negative entropy) of the predicted style labels:
\begin{align*}
    L_{\text{adv2}}(\Theta_E) = -\sum^M_{i=1}\sum^N_{j=1} H(p(j|\mathrm{Encoder}(\bm{x}_i;\Theta_E);\Theta_D)),
\end{align*}
where $N$ is the number of styles. Note that the two parts of the adversarial network update different sets of parameters. They work together to ensure that the output of $\mathrm{Encoder}(\bm{x}_i;\Theta_E)$ does not contain style information. 

Once the encoder is trained to produce the content representation, two generative approaches generate the text in the target style. The first approach involves training multiple decoders (shown in Fig. \ref{fig:adv_learning}-left) to take in the content representation and generate output in different styles.  
The second approach involves training a style embedding (shown in Fig. \ref{fig:adv_learning}-right) and concatenating it into the content representation to output in the target style using a decoder. 

\begin{figure*}[h]
    \centering
    \includegraphics[scale = 0.6]{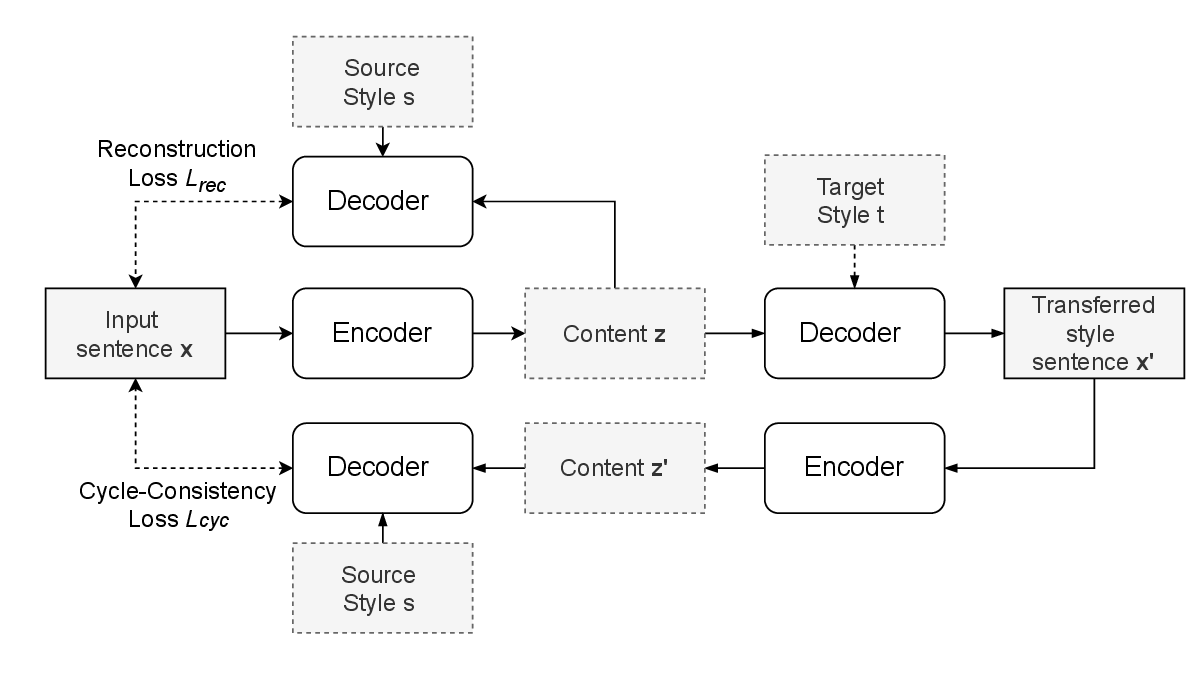}
    \caption{Reconstruction loss $L_{rec}$ and cycle-consistency loss $L_{cyc}$ used in text style transfer.}
   \label{fig:content_preservation_loss}
\end{figure*}

While the proposed adversarial learning-based TST framework has the potential to disentangle the style from the content information in text implicitly, there remains uncertainty if the reconstruction loss \cite{shen2017style,zhao2018adversarially,fu2018style,chen2018adversarial, logeswaran2018content, yin2019utilizing, zhao2018language} used its autoencoder architecture is sufficient to preserve the semantics of the input sentences. Researchers have proposed many variations of the adversarial learning-based TST framework to improve content presentation. For example, studies have explored adding cycle-consistency loss~\cite{chen2018adversarial, logeswaran2018content, yin2019utilizing, zhao2018language, lai2019multiple,john2019disentangled} to \red{preserve the semantics of a sentence transmitted in style}. Fig.~\ref{fig:content_preservation_loss} \red{illustrates} the utilization of the two losses. \red{The loss of reconstruction forces the decoder}, which takes the content representation $\bm{z}$ and original style embedding $\bm{a}_o$ as input, to reconstruct the input sentence $\bm{x}$ itself. \red{This is a common approach applied in the architecture of TST models that utilize an auto-encoder}. A cycle-consistency loss is used to prevent model collapse during the TST operation. Specifically, when a style-transferred sentence $\bm{x}'$ is fed into the TST model to transfer the sentence back to its original style, the cycle-consistency loss is used to enforce the generated sentences in the original style to be similar to the input sentence.

Besides employing loss functions, other adversarial learning-based TST methods have also proposed auxiliary components to enhance TST operation. Yin et al. \cite{yin2019utilizing} presented two partial comparators to guide adversarial learning, a content comparator that judges whether the input sentence and the generated sentence share the same content to improve content preservation, and a style comparator that judges if they have different styles. Lai et al. \cite{lai2019multiple} combined the adversarial learning framework with a word-level conditional mechanism to preserve content information by retaining style-unrelated words while modifying the other style-related words. The unique component between their model and previous work is the condition architecture of attributes. Many TST works \cite{hu2017toward,shen2017style,logeswaran2018content} treat the attributes $a$ as part of the initial vector fed into the RNN cell in the decoder, while Lai et al. \cite{lai2019multiple} concatenated the attributes $a$ with the output of the RNN cell at each time step $t$. Yang et al. \cite{yang2018unsupervised} replaced the style classifier in the adversarial learning framework with a target domain language model as a discriminator to provide richer and more stable token-level feedback during the adversarial learning process.

The adversarial learning-based models have demonstrated remarkable performance in the TST task. However, it is not without criticism and weaknesses. Lample et al. \cite{lampleSSDRB19} argued that the adversarial training is not practical on the disentanglement of style and content. They conducted experiments and demonstrated that a classifier trained on the learned content representation could recover the original style easily, suggesting the adversarial learning process may not have disentangled the style from the content representation. Another limitation of the adversarial learning-based TST technique is its \red{dependence} on the style classifier. The style classifier's accuracy limits the disentanglement of content and style; a poorly trained style classifier cannot discriminate the style in a given text, thus affecting its disentanglement performance. Furthermore, large annotated datasets may be required to train the style classifier. Future adversarial learning-based TST models could explore solutions to address this limitation.

\subsection{Back-Translation} 
\label{sec:tax-backtranslation}

Back-translation has been applied in NMT to generate artificial corpora~\cite{sennrich2015improving}. This approach was also used to generate pseudo-parallel TST datasets~\cite{zhang2020parallel}. However, this section reviews how back-translation has been explored to learn a content representation devoid of style for TST \cite{prabhumoye2018style,zhang2019machine}. This line of approaches is inspired by Rabinovich et al. \cite{rabinovich-etal-2017-personalized}, who found that the author's traits, such as gender, were obfuscated in human and machine translation. However, \red{it should be noted that} (1) Rabinovich et al. only showed that gender information was obfuscated in human and machine translation, but text attributes such as sentiment, tense, politeness \red{were} not explored; (2) no quantitative experiment was conducted to quantify lost gender information.

\begin{figure*}[h]
    \centering
    \includegraphics[scale = 0.6]{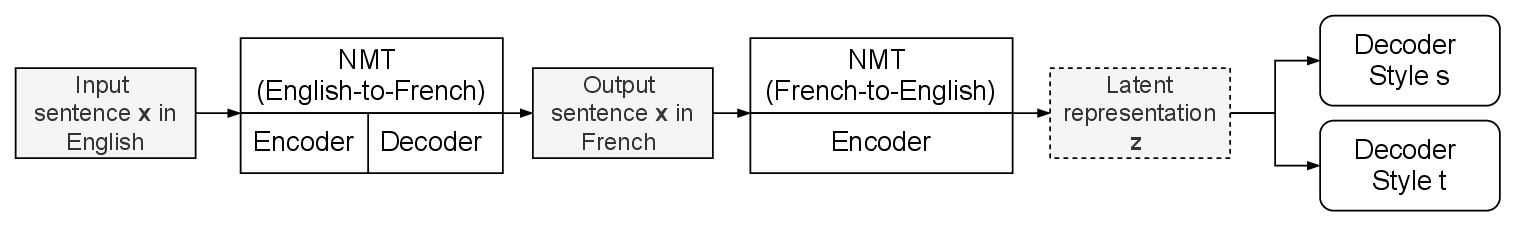}
    \caption{Back-translation framework for TST proposed by  Prabhumoye et al. \protect\cite{prabhumoye2018style}}
    \label{fig:back_translation}
\end{figure*}

Fig. \ref{fig:back_translation} shows a back translation TST method adopted by Prabhumoye et al. \cite{prabhumoye2018style}. The researchers attempted to use NMT models to rephrase the sentence and remove stylistic information from the text. Specifically,  an English text is first translated into French using an NMT model, and then the translated French text is translated back to English using another NMT model. The latent representation $\bm{z}$ learned by the NMT model is assumed to contain only content information devoid of stylistic properties. Finally, the latent representation $\bm{z}$ is used to generate text in a different style using the multi-decoder approach. Zhang et al. \cite{zhang2019machine} adopted an iterative back-translation pipeline to perform TST. The pipeline first learns a cross-domain word embedding to build an initial phrase-table, a set of corresponding phrases in the source and target styles. The phrase-table is then used to bootstrap an iterative back-translation model, which jointly trained two NMT systems to transfer text style.

\red{Similar to adversarial learning-based TST methods}, the main concern with back-translation TST methods is its effectiveness in content-style disentanglement; it is uncertain if the learned content representation is devoid of stylistic properties. It would be interesting to conduct a similar experiment proposed by Lample et al. \cite{lampleSSDRB19} to \red{investigate if it is possible to recover style information from the content representation}. Another potential improvement to the back-translation TST methods is to preserve semantics in the learned content representation. A possible solution may be to add cycle-consistency loss to existing back-translation TST methods \cite{prabhumoye2018style,zhang2019machine}.

\subsection{Attribute Control Generation}

Attribute control generation is an increasingly popular technique used in TST models~\cite{hu2017toward,tian2018Structured,lampleSSDRB19,dai2019style,zhang2018shaped,jain2019unsupervised,zhou2020exploring}. This technique often \red{learns} an attribute code $\bm{a}$ to control text generation in different styles. There are two broad strategies for applying attribute control generation to perform TST: (a) implicitly disentangling the latent representation $\bm{z}$ to contain only content information while learning the style attribute code; (b) without disentangling or constraining the latent representation $\bm{z}$ to contain only content information while learning the style attribute code. In both strategies, a classifier-guided loss is used to ensure that the generator $G$ generates a sentence $\bm{x'}$ \red{with} the desired style attribute. Specifically, the loss function minimizes the following: 


\begin{align*}
  L_{\text{Cla}}(\Theta_G,t) = - \mathbb{E}_{p(\bm{x})}[\log D(\bm{x'})],
\end{align*}
where $D$ is a style classifier pre-trained on real data $\bm{x}$. Similarly to adversarial training \red{on} generated sentences as introduced in Section \ref{sec:tax-adversarial}, $L_{\text{Cla}}$ can be trained with the Gumbel-softmax distribution \cite{DBLP:conf/iclr/JangGP17} or the policy gradient algorithm \cite{williams1992simple}.
\red{Note that adversarial training on generated sentences and classifier-guided loss share the same goal}:
to ensure that the transferred sentence $\bm{x'}$ carries the target attribute $t$. However, the two approaches used different loss functions.

As the attribute control generation technique also depended heavily on the style classifier, this approach shares similar limitations as the adversarial learning-based method. Specifically, the style classifier's accuracy limits learning a good style attribute code to perform TST. \red{However,} if the style attribute code is effectively learned, it could be potentially transferred to enhance other NLG tasks. For example, pre-trained style attribute code can be leveraged in machine translation to translate language and generate text in specific styles.



\subsubsection{Attribute Control Generation with Content-Style Disentanglement}
\label{sec:tax-controllable-disentanglement}

\begin{figure*}[h]
    \centering
    \includegraphics[scale = 0.6]{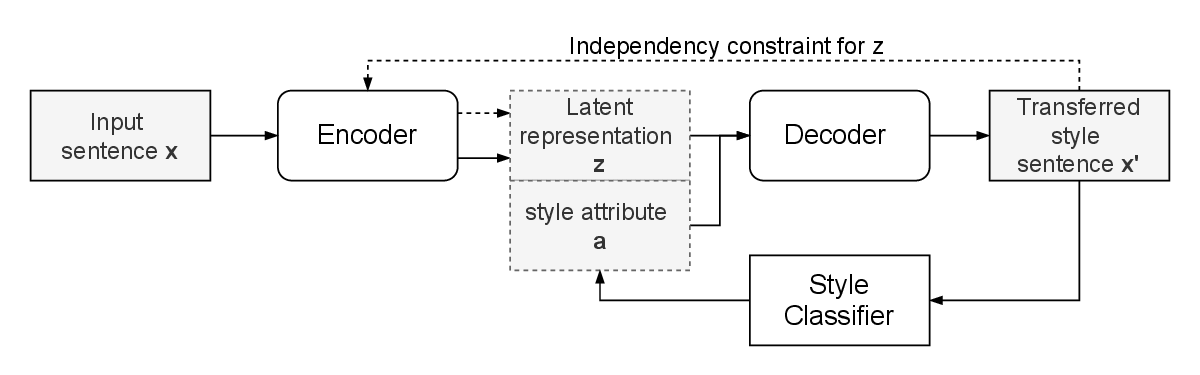}
    \caption{Attribute controlled generation proposed in by Hu et al \protect\cite{hu2017toward}}
    \label{fig:attribute_control}
\end{figure*}

The pivotal study by Hu et al. \cite{hu2017toward} is a representative work for applying the attribute control generation technique to perform content-style disentanglement for TST implicitly. Fig.~\ref{fig:attribute_control} illustrates the attribute control generation TST model of Hu et al. \cite{hu2017toward}. The proposed model \red{adopted} a variational autoencoder (VAE)~\cite{kingma2013auto} framework. Unlike an auto-encoder that learns a compressed representation for an input sequence, a VAE \cite{kingma2013auto} learns the parameters of a probability distribution that represents the data. The learned distribution can also be sampled to generate new data samples. Therefore, the generative nature of a VAE makes it widely explored and used in many natural language generation tasks \cite{gatt2018survey}. In \cite{hu2017toward}, they proposed a TST model that utilized VAE to learn a sentence's latent representation $\bm{z}$ and \red{leveraged} a style classifier to learn a style attribute vector $\bm{a}$. The probabilistic encoder of the VAE \red{captured} the variations of implicitly modeled aspects to guide the generator to avoid entanglement during attribute code manipulation. Finally, $\bm{z}$ and $\bm{a}$ \red{were} fed into a decoder to generate a sentence in the specific style. Specifically, the VAE loss function is shown as follows:
\begin{align*}
    L_{\text{VAE}}(\Theta_G,\Theta_E;\bm{x}) = \mathrm{KL}(q_E(\bm{z}|\bm{x})||p(\bm{z})) \\ - E_{q_E(\bm{z}|\bm{x})q_D(\bm{a}|\bm{x})}[\log p_G(\bm{x}|\bm{z},\bm{a})],
\end{align*}
where $\mathrm{KL}(\cdot || \cdot)$ is the KL-divergence, and $\Theta_E$ and $\Theta_G$ denote the parameters of the encoder and decoder, respectively. The conditional probabilistic encoder $E$, denoted as $q_E(\bm{z}|\bm{x})$, infers the latent representation $\bm{\bm{z}}$ given the input sentence $\bm{x}$, and $q_D(s|\bm{x})$ is the conditional distribution defined by the classifier $D$ for each structured variable in $\bm{a}$. \red{To ensure that $\bm{z}$ retains only the content information independent of style}, a \textit{independency constraint} is proposed to ensure that the latent representation $\bm{z}$ of the input sentence $\bm{x}$ and transferred style sentence $\bm{x'}$ remain close to each other. The added \textit{independency constraint} essentially ensures that the content information is disentangled from the input text and \red{encoded} in the latent representation $\bm{z}$.

\red{It should be noted that the training of sequential VAE models has proven to be very challenging due to the posterior collapse problem}~\cite{chen2016variational,bowman2016generating}. Annealing techniques are generally used to address this problem. Nevertheless, reconstructions from these models tend to differ from the input sequence. To address this limitation, Tian et al. \cite{tian2018Structured} \red{improved} the content presentation of \cite{hu2017toward} by adding more constraints to preserve style-independent content using part-of-speech (POS) information, and \red{a conditional language model of content}. Specifically, the researchers computed the distance between POS information of the input sentence and the output sentence as an error signal to the generator. This approach also assumed that the nouns of the sentence capture the content information. Therefore, they enforced the decoder to generate sentences with similar nouns.

\subsubsection{Attribute Control Generation without Content-Style Disentanglement}
\label{sec:tax-controllable-without}

\begin{figure*}[h]
    \centering
    \includegraphics[scale = 0.6]{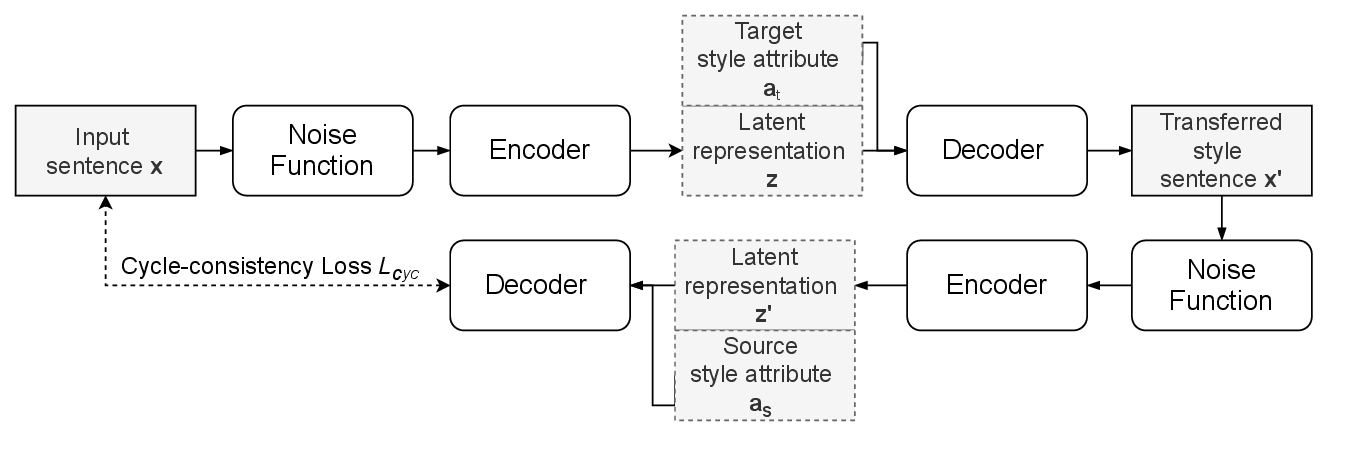}
    \caption{Attribute controlled generation with back-translation proposed in by Lample et al. \protect\cite{lampleSSDRB19}}
    \label{fig:lample}
\end{figure*}

As discussed in section \ref{sec:tax-adversarial}, Lample et al. \cite{lampleSSDRB19} argued that it is challenging to disentangle content and style in text. Therefore, researchers have proposed a new research direction to perform TST without content-style disentanglement. Fig. \ref{fig:lample} illustrates \red{the model proposed} by Lample et al. \cite{lampleSSDRB19}. The model employed denoising autoencoder \cite{vincent2010stacked} and back-translation \cite{sennrich2015improving} to build a translation between different styles. The noise function first corrupted the input sentence $\bm{x}$ by performing word drops and word order shuffling before feeding the corrupted output into an encoder to generate the latent representation $\bm{z}$. Next, $\bm{z}$ was subsequently combined with a trainable target-style attribute $\bm{a}_t$ and input into a decoder to generate a sentence $\bm{x'}$ in target style. Finally, a back-translation process \red{was} initiated to have the generated sentence fed into the sample encoder-decoder process to reconstruct the original sentence using the latent representation $\bm{z'}$ and the original style attributes $\bm{a}_s$. \red{The underlying intuition is that the decoder is forced to leverage on the target attribute $\bm{a}_s$ and $\bm{a}_t$ to generate sentence $\bm{x}$ and $\bm{x'}$ as latent representations $\bm{z}$ and $\bm{z'}$ only capture information from corrupted input sentences}. It is worth noting that the proposed model did not contest or constrain that the latent representation $\bm{z}$ only captures style-independent content information. Therefore, Lample et al. \cite{lampleSSDRB19} argued that this approach could perform TST without \red{disentangling content and style information in texts}.


Other variants of attribute-controlled approaches perform TST without content and style disentanglement. For example, Dai et al. \cite{dai2019style} adopted a Transformer-based autoencoder \cite{vaswani2017attention} to perform TST with a trainable style attribute. The model's goal is to leverage the power of the attention mechanism in the Transformer to achieve better style transfer and better content preservation. Zhang et al. \cite{zhang2018shaped} proposed a shared-private encoder-decoder (SHAPED) framework that learns the style attributes to transfer the text style. Li et al. \cite{li2019domain} extended the attribute-controlled TST works and proposed a domain-adaptive TST that enables style transfer to be performed in a domain-aware manner. Specifically, the proposed model also learned domain vectors of the text in the source and target domains in addition to the latent style attributes. The domain vectors represent different domains, such as movie reviews and restaurant reviews. This encourages the model to perform style transfer in a domain-aware manner rather than directly sharing the style transfer model. Intuitively, this design effectively avoids generations such as "The pizza is dramatic!" when domains are unspecified. The domain vectors are subsequently used with the style attributes and sentence's latent representations to perform TST across domains. Zhou et al. \cite{zhou2020exploring} proposed \red{a fine-grained attribute control method} to perform TST. The proposed model utilized an attentional Seq2Seq model that dynamically exploits each output word's relevance to the target style for text style transfer. The model included a carefully-designed objective function that fine-tuned the model's style transfer, style relevance consistency, content preservation, and fluency modeling loss terms.  Dathathri et al. \cite{DBLP:conf/iclr/DathathriMLHFMY20} proposed a language model for controllable language generation. The proposed method \red{combined} a pre-trained language model with one or more simple style classifiers that \red{guided} text generation without any further training of the language model.


\subsection{Entangled Latent Representation Editing}
\begin{figure*}[h]
    \centering
    \includegraphics[scale = 0.6]{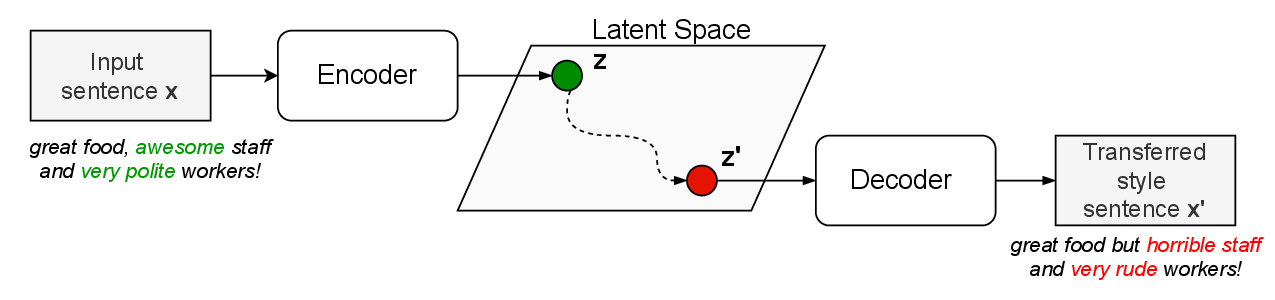}
    \caption{Common framework for editing text's latent representation for TST.}
    \label{fig:edit}
\end{figure*}

Another line of work that attempted to perform TST without any content and style disentanglement is directly editing the latent representation learned using the autoencoder-based models. Fig. \ref{fig:edit} shows a common framework adopted by the work that edits latent representations for TST. Typically, the latent representation $\bm{z}$ learned using an autoencoder is manipulated with various methods. Fig.\ref{fig:edit} shows a common framework used in editing text's latent representation for TST. A style classifier is jointly trained with the autoencoder, where the training iteratively updates the latent representation $\bm{z}$ in the constraint space to maximize the prediction confidence score for the target attribute label of this style classifier. Specifically, each update is calculated based on the gradient of style classifier loss with respect to $\bm{z}$. The manipulated latent representation $z'$ is then input into the decoder to generate text of the target style.


In earlier work, Mueller et al. \cite{mueller2017sequence} explored manipulating the hidden representation learned using VAEs to generate sentences that contain a particular style measured by a corresponding classifier. However, we note that there was no quantitative evaluation of the effectiveness of text style transfer in this earlier work. 

Xu et al. \cite{xu2019variational} conducted extensive experiments to investigate the latent vacancy in unsupervised learning of controllable representation when modeling text with VAEs. Similar to the study in \cite{mueller2017sequence}, Xu et al. studied the impact on text style when manipulating the factors in latent representations, and both works found that when a manipulation failed to decode correct sentences, it was due to the manipulated results in representation areas that the decoder never saw during training. \red{Both works proposed performing latent space manipulations within a constrained set to handle this issue}. Specifically, \cite{mueller2017sequence} \red{opined that} the distribution of natural sequences is geometrically simple in the latent space by endowing $\bm{z}$ with their simple $N(0,\bm{I})$ prior. Whereas in \cite{xu2019variational} the researchers proposed constraining the posterior mean to a learned probability simplex and only \red{perform} manipulation within the probability simplex.

\red{Similarly to the study in \cite{mueller2017sequence}, Liu et al. \cite{liu2020revision} adopted a gradient-based optimization in continuous space to manipulate the learned latent representation using VAE and style classifiers to perform TST}. Moreover, the proposed method naturally can simultaneously control multiple fine-grained attributes, such as sentence length and the presence of specific words, when performing TST tasks. Wang et al. \cite{wang2019controllable} adopted a similar approach and performed a fast-gradient-iterative-modification algorithm to edit \red{the learned latent representation} using a Transformer-based autoencoder until the generated text conforms to the target style.

\red{The main challenge of TST methods for editing latent representations is capturing the boundaries of manipulations}. As mentioned, both \cite{xu2019variational} and \cite{mueller2017sequence} observed that the decoder is unable to generate sentences in \red{target style} if the manipulated results go beyond the representation areas observed by the decoder during training. \red{The current solutions focused on limiting manipulation within a constrained latent space.} However, it is unclear how the constrained latent space affects TST performance. Future research would need to explore better approaches to address this challenge.

\subsection{Reinforcement Learning} 
\label{sec:tax-Reinforcement}

\begin{figure*}[h]
    \centering
    \includegraphics[scale = 0.6]{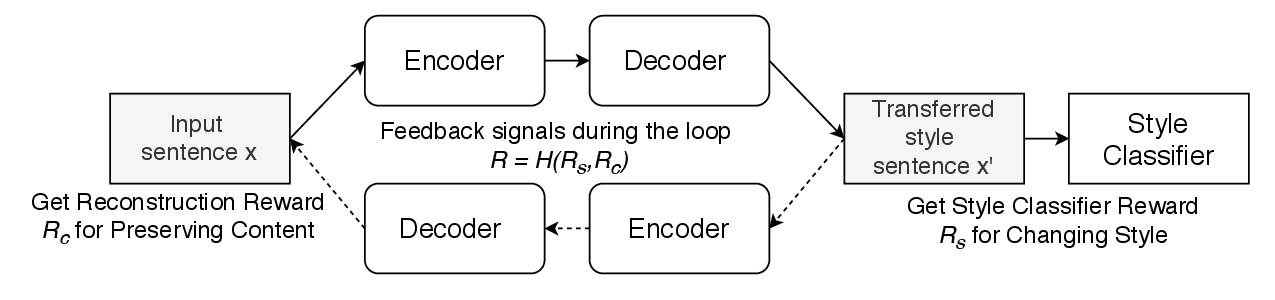}
    \caption{Dual reinforcement learning model for TST proposed in \protect\cite{luo2019dual}.}
    \label{fig:reinforce}
\end{figure*}

Reinforcement learning has also been applied to perform TST. The core idea of reinforcement-learning-based TST is using the specially designed reward functions to guide the TST process instead of the various loss functions applied to other TST methods. The policy gradient algorithm \cite{williams1992simple} is used to maximize the expected reward of the transferred text to optimize the parameters of a reinforcement learning TST model. The policy gradient algorithm makes training easier without worrying about the difficulty of discrete training caused by the automatic regression decoding process. However, due to the high variance of the sampling gradient, training with this method may be unstable. \red{For example, Luo et al. \cite{luo2019dual} proposed to learn two Seq2Seq models between two styles through reinforcement learning, without disentangling style and content}. Fig. \ref{fig:reinforce} illustrates the proposed dual reinforcement learning framework. The authors considered the learning of source-to-target style and target-to-source style as a dual-task. The style classifier reward, $R_s$, and reconstruction reward, $R_c$, are designed to encourage style transfer accuracy and content preservation. The overall reward is the harmonic mean of the two rewards, and it was used as the feedback signal to guide learning in the dual-task structure. As such, the model can be trained via reinforcement learning without parallel data or content-style disentanglement. 

Gong et al. \cite{gong2019reinforcement} \red{proposed a generator-evaluator framework based on reinforcement learning to perform TST}. The proposed model employs an attention-based encoder-decoder model to transfer and generate sentences in a target style like previous TST works. \red{However, unlike previous models that utilized a style classifier to guide the generation process, the proposed model employed a style classifier, a semantic model, and a language model to provide style, semantic, and fluency rewards to guide text generation. The fundamental intuition is that text style transfer should ensure style transfer, content preservation, and the generation of fluent sentences.}  

Reinforcement learning is a promising research direction for TST. Although the existing reinforcement learning methods still heavily relied on style classifiers to drive the TST process, the technique offers the possibility to design other reward functions to guide the transfer process. Computational linguistics researchers could design reward functions based on linguistics discourse theories to score text style and guide the TST process.

\subsection{Purely Unsupervised Methods}
Most of the TST studies discussed above are based on the assumption that style-specific corpora (i.e., parallel or non-parallel) are available. This section introduces another breed of TST models that performed style transfer in a purely unsupervised setting where only a mixed corpus of unspecified style is available. 

\red{There are relatively fewer proposed works to perform TST in a purely unsupervised setting} \cite{radford2017learning,jain2019unsupervised,xu2020variational,shen2020educating}. In an earlier study, Radford et al. \cite{radford2017learning} \red{explored the properties of recurrent language models at byte level where an LSTM model is trained on a text preprocessed as a sequence of UTF-8 encoded bytes in an unsupervised fashion.} Interestingly, the researchers discovered a single neuron unit within the trained LSTM model that \red{corresponds directly} to sentiment and manipulated neuron units to transfer sentiments in sentences. Xu et al. \cite{xu2020variational} conducted a similar study and successfully detected a latent dimension responsible for sentiment with 90+\% accuracy. Subsequently, they used unsupervised representation learning to separate style and content from a mixed corpus of unspecified styles and achieved satisfactory results in sentiment transfer tasks.

Jain et al. \cite{jain2019unsupervised} \red{proposed} an unsupervised training scheme to perform text formalization with unlabeled data. To train the encoder-decoder framework in an unsupervised manner, they \red{employed} external scorers to provide style information. Based on off-the-shelf language processing tools, these scorers \red{decided} the learning scheme of the encoder-decoder based on its actions. The authors labeled the formality level of sentences based on the scores given by the external scorers to help the TST model capture formality information.

Shen et al. \cite{shen2020educating} extended adversarial auto-encoders (AAE) with a denoising objective where original sentences are reconstructed from perturbed versions. The proposed denoising AAE model will map similar sentences to similar latent representations and make the boundaries of different representation clusters more obvious. To perform sentiment transfer, they \red{computed} a single “sentiment vector” by averaging the latent code $\bm{z}$ separately for 100 (non-parallel) positive sentences and negative sentences in the development set, and then \red{calculated} the difference between the two. Given a sentence from the test set, they \red{attempted} to change its sentiment from positive to negative or negative to positive through simple addition/subtraction of the sentiment vector.

The competitive performances of the purely unsupervised TST methods are encouraging. However, we noted that most of the aforementioned methods were evaluated on the sentiment transfer task. Specifically, in \cite{radford2017learning} and \cite{xu2020variational}, which have identified the latent attribute that corresponds to sentiment in the text, it is unclear \red{whether} the methods can be applied to other stylistic properties, such as text formality. More studies would need to be conducted to evaluate if the purely unsupervised methods can be generalized to other TST tasks.

\subsection{Other Peculiar Methods}
\label{sec:tax-other}
TST is a relatively new research area that is constantly evolving, and there are particular methods that we cannot sort into the techniques defined in our taxonomy framework. This section covers some of these TST methods.

He et al. \cite{He2020A} proposed a probabilistic deep generative model to infer the latent representations of sentences for TST. The proposed model hypothesizes a parallel latent sequence that generates each observed sequence, and the model learns to transform sequences from one domain to another in a non-parallel supervised setting. Specifically, the model combines a recurrent language model with an encoder-decoder transducer to infer the sentence's latent representations in the assumed parallel style corpus. The inferred latent representation is then used to generate a sentence of a specific style via a decoder. This was the first work that applied the text transduction method on the TST task to the best of our knowledge.

Unlike the other adversarial learning-based and attribute control generation methods that implicitly disentangled content and style information for TST, Cheng et al. \cite{DBLP:conf/acl/ChengMSMZLC20} leveraged mutual information to achieve the same goal. Specifically, the researchers \red{minimized} the mutual information between style and content embedding to induce style and content embeddings into two independent low-dimensional spaces. Concurrently, they maximized mutual information between content embedding and input sentence and style embedding and style label to ensure that content embedding sufficiently captures content information from the input sentence and style embedding are strongly correlated to style label. Finally, to perform the TST task, two sentences are encoded into a disentangled representation, and the style embedding from one sentence and the content embedding from another are combined to generate a new sentence.

Syed et al. \cite{syed2020adapting} proposed an interesting framework that adapts language models for non-parallel author-stylized rewriting. The researchers first pre-trained a language model on a large English corpus containing 141 authors' articles and Wikipedia articles in an unsupervised fashion using masked language modeling. Subsequently, they \red{cascaded} the pre-trained language models with an encoder-decoder framework for text generation. The encoder-decoder was fine-tuned separately on each of the target author's corpus using Denoising Auto-Encoder loss:

\begin{equation}
    L_{DAE} = \bm{E}_{\bm{x} \sim S}[-logP(\bm{x}|C(\bm{x}))]
\end{equation}

\red{where $C(\bm{x})$ is the noisy version obtained by dropping and replacing works with a special token $[BLANK]$ of the input sentence $\bm{x}$ and $S$ are the sentences in the corpus of the target author.} While performing TST of authors' writing style, the fine-tuned author-specific encoder-decoders were used to generate sentences in the target author's style.








\section{Resources and Evaluation Methods}
\label{sec:evaluation}
As TST is a relatively new research area, new methods will need to be designed to evaluate the TST algorithms. This section summarizes the downstream tasks and existing datasets used to evaluate the TST models. Next, we discuss the automated and human evaluation methods used to assess the quality of TST algorithms.

\subsection{Tasks and Datasets}
\textbf{Datasets}. Table \ref{tbl:all_dataset_statistics} summarizes the datasets used in existing studies to evaluate the performance of TST algorithms. These corpora often contain texts labeled with two or more attributes. For example, the Yelp dataset contains review text records labeled with binary sentiment class (i.e., positive or negative), and the Caption dataset contains caption-text records labeled with \textit{romantic}, \textit{humorous}, and \textit{factual} classes. Most of these datasets are non-parallel datasets, i.e., there are no matching text pairs in the different attribute classes, except for Shakespearean-Modern and GYAFC. It is also interesting to note that while the GYAFC corpus is a parallel dataset, most of the existing TST studies assume a non-parallel setting when training the TST models with this dataset. 

Many downstream tasks have been proposed to leverage these datasets to evaluate TST models. In the rest of this section, we will review these downstream tasks in greater detail. 



\begin{table*}
\centering
\caption{Dataset statistics for text style transfer.}
\label{tbl:all_dataset_statistics}
\begin{tabular}{|c|c|c|c|}
\hline
Dataset & Subset & Attributes & \#Text records \\
\hline
\multirow{2}*{Shakespearean-Modern \cite{jhamtani2017shakespearizing}} & \multirow{2}*{-} & Shakespeare & 21,076 \\
\cline{3-4}
 & & Modern & 21,076 \\
\hline
\multirow{2}*{Yelp \cite{shen2017style}}  & \multirow{2}*{-} & Positive & 381,911 \\
\cline{3-4}
 & & Negative & 252,343 \\
 \hline
\multirow{2}*{IMDb \cite{dai2019style} }  & \multirow{2}*{-} & Positive & 181,869 \\
\cline{3-4}
 & & Negative & 190,597 \\
\hline
\multirow{2}*{Amazon \cite{he2016ups}}  & \multirow{2}*{-} & Positive & 278,713 \\
\cline{3-4}
 & & Negative & 279,284 \\
\hline
\multirow{4}*{GYAFC \cite{rao2018dear}}  & \multirow{2}*{F\&R} & Informal & 56,087 \\
\cline{3-4}
&  & Formal & 55,233 \\
\cline{2-4}
 & \multirow{2}*{E\&M} & Informal & 56,888 \\
\cline{3-4}
&  & Formal & 56,033 \\
\hline
\multirow{2}*{PNTD \cite{fu2018style}} & \multirow{2}*{-} & Paper & 107,538 \\
\cline{3-4}
 & & News & 108,503 \\
\hline
\multirow{3}*{Caption \cite{li2018delete}} & \multirow{3}*{-} & Romantic & 6,300  \\
\cline{3-4}
 & & Humorous & 6,300  \\
 \cline{3-4}
 & & Factual & 300 \\
 \hline
\multirow{2}*{Gender \cite{prabhumoye2018style}} & \multirow{2}*{-} & Male & 1,604,068 \\
\cline{3-4}
 & & Female & 1,604,068 \\
  \hline
\multirow{2}*{Political \cite{prabhumoye2018style} } & \multirow{2}*{-} & Democracy & 298,961 \\
\cline{3-4}
 & & Republican & 298,961 \\
  \hline
\multirow{4}*{Offensive \cite{dos2018fighting}} & \multirow{2}*{Twitter} & Offensive & 74,218 \\
\cline{3-4}
&  & Non-offensive & 1,962,224 \\
\cline{2-4}
 & \multirow{2}*{Reddit} & Offensive & 266,785 \\
\cline{3-4}
&  & Non-offensive & 7,096,473 \\
\hline
\end{tabular}
\end{table*}

\textbf{Author Imitation.} Author imitation is the task of paraphrasing a sentence to match a specific author's style. To perform this task, Xu et al. \cite{xu2012paraphrasing} collected a parallel dataset, which captures the line-by-line \red{modern interpretations} of 16 of Shakespeare's 36 plays (\textit{Antony \& Cleopatra, As You Like It, Comedy of Errors, Hamlet, Henry V, etc.}) using the educational site \textit{Sparknotes} \footnote{www.sparknotes.com}. The goal was to imitate Shakespeare's text style by transferring modern English sentences into Shakespearean-style sentences. This dataset is publicly available\footnote{ http://tinyurl.com/ycdd3v6h} and was also used in other TST studies \cite{jhamtani2017shakespearizing,He2020A}.

The imitation of the author's writing style is an exciting task in TST. There are many interesting industrial applications, such as transferring famous novel authors' writing styles into other stories and unifying multiple authors' writing styles to a single author in a collaborative setting. However, the Shakespearean-Modern dataset is the only known corpus that facilitates author imitation in TST studies. There are also some apparent limitations in the corpus; the dataset size is small and the approach is limited to transfer to only one author's style. An interesting future work may be collecting text written by various authors and transferring the text style among multiple authors.




\textbf{Sentiment transfer.} Sentiment transfer is a \red{common assessment} task adopted in many TST studies. The task involves modifying the sentiment of a sentence while preserving its original contextual content. Table \ref{tbl:yelpexample} shows an example of the sentiment transfer task. Given the input sentence with positive sentiment, ``Everything is fresh and so delicious!'', the goal of the TST model is to convert the sentence into negative sentiment while preserving the contextual content information. In this example, the word ``Everything'' represents the content information preserved during the style transfer operation. The examples also reveal an interesting aspect of the sentiment transfer task: while the sentence's style is transferred and the contextual content is preserved, the semantics of the sentence has also changed. For example, a sentence supporting a particular political party can semantically change to a negative opinion during the sentiment transfer process. \red{However, this task is widely used to evaluate TST models and three popular datasets have been proposed for this task:}


\begin{itemize}
    \item \textit{Yelp}\footnote{https://github.com/shentianxiao/language-style-transfer} is a corpus of restaurant reviews from Yelp collected in \cite{shen2017style}. Original Yelp reviews are on a 5-point rating scale. As part of data preprocessing, reviews with 3 points and above ratings are labeled positive, while those below 3 points are labeled negative. The reviews with an exact 3 points rating are considered neutral and excluded from this dataset. 
    \item \textit{Amazon}\footnote{https://github.com/lijuncen/Sentiment-and-Style-Transfer\label{lijuncen_git}} \red{is Amazon's product review dataset collected in \cite{he2016ups}.} It is preprocessed using the same method as in the Yelp dataset. 
    \item \textit{IMDb}\footnote{https://github.com/fastnlp/nlp-dataset} is a movie reviews dataset. Dai et al. \cite{dai2019style} constructed this dataset by performing similar data preprocessing methods on a publicly available and popular movie reviews dataset \cite{maas2011learning}.
\end{itemize}

\begin{table}[t]
\caption{Sentiment transfer examples.}
\label{tbl:yelpexample}
\begin{tabular}{cl}
\hline
& {\color{red}\textbf{Positive}} \textbf{Sentiment} $\longleftrightarrow$ {\color{green}\textbf{Negative}} \textbf{Sentiment} \\
\hline
\textbf{input} & Everything is fresh and so delicious! \\
\hline
Ref-0 & Everything was so stale. \\
Ref-1 & Everything is rotten and not so delicious. \\
Ref-2 & Everything is stale and horrible. \\
Ref-3 & Everything is stale and tastes bad. \\
\hline
\end{tabular}
\end{table}

\textbf{Formality transfer.} The formality transfer task involves modifying the formality of a given sentence. Typically, an informal sentence is transformed into its formal form, and vice versa. Formality transfer is presumably more complex than sentiment transfer, as multiple attributes may affect text formality. For instance, the modification of the sentence structure, text length, punctuation, and capitalization, may influence text formality.  Table \ref{tbl:GYAFCexample} shows an example of transferring an informal sentence to its formal form. We illustrate that simple keyword replacement methods cannot transfer the formality of a sentence from the four transferred formal sentences. After an informal sentence is transferred to its formal form, the length of the sentence (as shown in Ref-0) and punctuation may be changed (e.g., replacement of ellipsis with a full stop). Furthermore, unlike sentiments, a sentence's formality is highly subjective; individuals may perceive a sentence's degree of formality differently.  

\textit{GYAFC}\footnote{https://github.com/raosudha89/ GYAFC-corpus} is the largest human-labeled parallel dataset proposed for the formality transfer task \cite{rao2018dear}. The authors extracted informal sentences from Entertainment\&Music (E\&M) and Family\&Relationship (F\&R) domains of the Yahoo Answers L6 corpus \footnote{https://webscope.sandbox.yahoo.com/catalog.php?datatype=l}. The collected dataset was further preprocessed to remove too short or long sentences. Finally, the authors crowd-sourced workers to manually rewrite the informal sentences to their formal form, resulting in a parallel formality dataset. This dataset was also widely used to evaluate the recent TST models.

\begin{table*}[t]
\centering
\caption{Formality transfer examples.}
\label{tbl:GYAFCexample}
\begin{tabular}{cl}
\hline
& {\color{red}\textbf{Informal}} \textbf{Formality} $\longleftrightarrow$ {\color{green}\textbf{formal}} \textbf{Formality} \\
\hline
\textbf{input} & He loves you, too, girl...Time will tell. \\
\hline
Ref-0 & He loves you as well, but only time can tell what will happen. \\
Ref-1 & He loves you too, lady...time will tell. \\
Ref-2 & He loves you, as well. Time will tell. \\
Ref-3 & He loves you too and time will tell. \\
\hline
\end{tabular}
\end{table*}

\textbf{Paper-news title transfer.} Paper-news title transfer was the task of transferring a title to a different type while preserving the content. Fu et al. \cite{fu2018style} collected the \textit{PNTD}\footnote{https://github.com/fuzhenxin/textstyletransferdata} dataset, which consisted of paper titles retrieved from academic publication archive websites such ACM Digital Library, Arxiv, Sprint, ScienceDirect, and Nature, and news titles from the UC Irvine Machine Learning Repository.



\textbf{Captions style transfer.} Li et al. 2018 \cite{li2018delete}  proposed the task of transferring factual formal captions into romantic and humorous styles. The researchers collected the caption dataset\footref{lijuncen_git}, where each sentence was labeled as factual, romantic, or humorous. This is also the smallest TST dataset. 


\textbf{Gender style transfer.} The differences between male and female writing styles is a widely studied research topic in sociolinguistics. \red{Prabhumoye et al. \cite{prabhumoye2018style} extended the sociolinguistic studies to perform TST between text written by different genders}, that is, transfer a text written by a male to a female writing style and vice versa. The researchers constructed the gender dataset\footnote{https://github.com/shrimai/Style-Transfer-Through-Back-Translation\label{back_translation}} by first preprocessing a Yelp review dataset annotated with the gender of the reviewers \cite{reddy2016obfuscating} by splitting the reviews into sentences and preserving the gender label for each sentence. \red{The sentences that are deemed gender neutral are removed from the dataset.} 


\textbf{Political slant transfer.} Political slant transfer modifies the writer's political affiliation writing style while preserving the content. Prabhumoye et al. \cite{prabhumoye2018style} collected comments of Facebook posts from 412 members of the United States Senate and House who have public Facebook pages. The comments are annotated with the congressperson's political party affiliation: democracy or republican. Table \ref{tbl:politicalexample} shows examples of comments collected in the dataset\footref{back_translation}.


\begin{table*}[t]
\centering
\caption{Political slant examples.}
\label{tbl:politicalexample}
\begin{tabular}{cl}
\hline
Republican & defund them all, especially when it comes to the illegal immigrants. \\
Republican & thank u james, praying for all the work u do .  \\
\hline
Democracy& on behalf of the hard-working nh public school teachers- thank you! \\
Democracy & we need more strong voices like yours fighting for gun control . \\
\hline
\end{tabular}
\end{table*}

\textbf{Offensive language correction.} Offensive and abusive language is a growing problem on social media. 
The offensive language correction task aims to transfer offensive sentences into non-offensive ones. Santos et al. \cite{dos2018fighting} collected posts from Twitter and Reddit. The posts were subsequently classified into ``offensive'' and ``non-offensive'' classes using a classifier pre-trained on an annotated offensive language dataset. 


\textbf{Multiple-attribute style transfer.} Thus far, the tasks we have discussed involve transferring text between two style attributes. Lample et al. \cite{lampleSSDRB19} proposed a multiple-attribute style transfer task and collected multi-style attribute datasets based on Yelp and Amazon review datasets. Table \ref{tbl:multi_dataset_statistics} summarizes the statistics of three datasets collected in their studies. The goal is to transfer a text with specific multiple style attributes, such as sentiments and gender of the author. The specification of multiple attributes makes the TST task more complex and realistic as the text style should be multi-faceted. For instance, the gender and sentiment of the author could both affect the style of the text. We postulate that multiple-attribute style transfer may be one of the TST research's future directions, and we will discuss this further in Section \ref{sec:challanges}.


\begin{table*}
\centering
\caption{Dataset statistics for multiple-attribute transfer datasets.}
\label{tbl:multi_dataset_statistics}
\setlength{\tabcolsep}{0.4mm}{
\begin{tabular}{|c|cc|cc|ccccc|}
\hline
Dataset & \multicolumn{2}{|c|}{Sentiment} & \multicolumn{2}{|c|}{Gender} & \multicolumn{5}{|c|}{Category} \\
\hline
\multirow{2}*{FYelp} & Positive & Negative & Male & Female & American & Asian & Bar & Dessert & Mexican \\
& 2,056,132 & 639,272 & 1,218,068 & 1,477,366 & 904,026 & 518,370 & 595,681 & 431,225 & 246,102 \\
\hline
\multirow{2}*{Amazon} & Positive & Negative & - & - & Book & Clothing & Electronics & Movies & Music \\
 & 64,251,073 & 10,944,310 & - & - & 26,208,872 & 14,192,554 & 25,894,877 & 4,324,913 & 4,574,167 \\
\hline
\multirow{2}*{Social Media Content} & Relaxed & Annoyed & Male & Female & age:18-24 & age:65+ & & & \\
& 7,682,688 & 17,823,468 & 14,501,958 & 18,463,789 & 12,628,250 & 7,629,505 & & &\\
\hline
\end{tabular}}
\end{table*}


\subsection{Automated Evaluation}
Several automated evaluation metrics have been proposed to measure the effectiveness of TST models \cite{pang2019towards,pang2019daunting,pang2019unsupervised,mir2019evaluating}. In general, these metrics evaluate the TST algorithms on three criteria:
\begin{enumerate}
    \item The ability to transfer the text style.
    \item The amount of original content preserved after the TST operation.
    \item The fluency of the transferred style sentence.
\end{enumerate}

A TST algorithm \red{that fails to meet any of these three criteria} is considered ineffective for performing the TST task. For example, a TST algorithm transfers a negative sentiment sentence, ``the pasta tastes bad!'', to a positive one, ``the movie is great!''. While the algorithm can transfer the input text style, i.e. from negative to positive sentiment, it fails to preserve the original statement's content, i.e. describing the pasta. Like many other tasks of natural language generation, the transferred sentence will also have to achieve a certain level of fluency for the TST algorithm to be \red{useful} in real-world applications. Therefore, an effective TST algorithm will have to perform well in all three evaluation criteria.

\textbf{Transfer strength.} The transfer strength of a TST model or its ability to transfer the text style is commonly measured using \textit{Style Transfer Accuracy} \cite{hu2017toward,shen2017style,fu2018style,luo2019dual,john2019disentangled}. Typically, a binary style classifier TextCNN \cite{DBLP:conf/emnlp/2014} is first pre-trained separately to predict the style label of the input sentence. The style classifier is then used to approximate the style transfer accuracy of the transferred style sentences by considering the target style as the ground truth. It is also important to note that the style classifier is not perfect. For instance, when pre-trained on the Yelp and GYAFC datasets and applied to classify tweets on their respective validation dataset, the style classifier can achieve 97.2\% accuracy on the Yelp dataset, while it is only able to achieve 83.4\% accuracy on the GYAFC dataset. 
An alternative metric is to measure the \textit{Earth Mover's Distance} \cite{rubner1998metric} between the style distributions of the input text and the transferred text. The Earth Mover's distance can be interpreted as a ``cost'' to turn one distribution into the other, or how ``intense'' the transfer is \cite{mir2019evaluating}.

\textbf{Content preservation.} To quantitatively measure the amount of original content preserved after the style transfer operation, TST studies have borrowed three automated evaluation metrics that are commonly used in other natural language generation tasks:

\begin{itemize}
    \item \textit{BLEU}: The BLEU score \cite{papineni2002bleu} was originally designed to evaluate the quality of a machine-translated text. The BLEU score was one of the first metrics to claim a high correlation with human judgment on \red{the quality of the translated text.} To calculate the BLEU score, machine-translated text is compared with a set of good quality reference translations. Similarly, in TST, the BLEU is computed when parallel TST datasets or human references are available. Specifically, we compute the BLEU score between the transferred sentences and the parallel references available to evaluate content preservation.
    \item \textit{source-BLEU (sBLEU)}:  Nevertheless, most of the TST tasks assume a non-parallel setting, and matching references of \red{style-transferred} sentences are not always available. Therefore, \red{TST studies often apply a modified score \textit{source-BLEU (sBLEU)} when the transferred sentence is compared to its input sentence from the source}. The intuition is that the content is assumed to be preserved, and the transferred sentence would share many overlap n-grams with the original sentence.
    \item \textit{Cosine Similarity}: Fu et al. \cite{fu2018style} calculated the cosine similarity between the original sentence embeddings and the transferred sentence embeddings. The intuition is that the embeddings of the two sentences should be close to preserving the semantics of the transferred sentence.    
    \item \textit{Word Overlap}: Vineet et al. \cite{john2019disentangled} argued that the cosine similarity is not a sensitive metric as the original and transferred sentences may have high cosine similarity scores even the content of the sentences are different. Thus, they used a simple metric that counts the unigram word overlap rate of the original and style-transferred sentences. Note that stop words and style-attributed words (e.g., sentiment words) are excluded from the word overlap calculation.
\end{itemize}

\textbf{Fluency.} Generating fluent sentences is a common goal for almost all natural language generation models. A common approach to measuring \red{the fluency of a sentence}is using a trigram Kneser-Ney language model \cite{479394}. The Kneser-Ney language model is pre-trained to estimate the empirical distribution of trigrams in a training corpus. Subsequently, the \textit{perplexity score} of a generated sentence is calculated by comparing the sentence's trigram and the estimated trigram distribution. The intuition is that a generated sentence with a lower perplexity score is considered to be more ``aligned'' to the training corpus and therefore more fluent. In TST tasks, the language model is similarly trained on the TST datasets, and the perplexity scores of the style-transferred sentences are computed to evaluate the sentences' fluency.

\subsection{Human Evaluation}
Few TST studies have performed human evaluations on their proposed TST algorithms \cite{shen2017style,li2018delete} as such evaluations are often expensive and laborious. Human workers are crowd-sourced in a typical human evaluation setting to rate how the style transferred sentence fair using a range scale based on the three evaluation criteria. For example, given a pair of original and transferred sentences, a human worker is asked to rate the degree to which the content is preserved in the transferred sentence on a scale of 1 to 5 points, with 5 points being ``\textit{very well preserved}''. Multiple human workers are asked to evaluate a given pair of original and transferred sentences, and the average scores are reported to reduce individual bias. Although researchers have made great efforts to ensure the quality of human evaluation in TST tasks, the evaluation approach has proven to be very challenging, as interpretation of the text style is subjective and can vary between individuals \cite{pang2019towards,pang2019daunting,mir2019evaluating}. Nevertheless, human evaluations still offer valuable insights into how well TST algorithms can transfer style and generate sentences acceptable by human standards.

\section{Reproducibility Study}
\label{sec:experiment}
Although most of the existing TST methods were evaluated in the original work using the downstream tasks discussed in Section \ref{sec:evaluation}, \red{the experiments were often carried out with few or no baselines.} Therefore, we conducted a reproducibility study\footnote{Code implementations of the reproduced models is compiled in this repository: \url{https://gitlab.com/bottle_shop/style/tst_survey}} and benchmarked 19 TST models on two popular corpora: Yelp reviews and GYAFC, representing the sentiment transfer task and the formality transfer task, respectively. To the best of our knowledge, this is the first time that so many TST models have been evaluated on the same datasets. \red{Specifically, the experimental results from this study provide new insights into how each TST algorithm performs and compares with each other} in terms of \textit{transfer strength}, \textit{content preservation}, and \textit{fluency}. This section is organized as follows: We first describe the experimental setup of our reproducibility study. Next, we discuss the experimental results for the tasks \textit{sentiment transfer} and \textit{formality transfer}. We also perform the trade-off analyses to investigate how the relationships between multiple evaluation criteria influence the TST model performance. Finally, we perform a human evaluation on a subset of representative TST models and report the results.

\subsection{Experimental Setup}
\textbf{Environment Settings.} The experiments were performed on an Ubuntu 18.04.4 LTS system with 24 cores, 128 GB RAM and a clock speed of 2.9 GHz. The GPU used for deep neural network-based models was Nvidia GTX 2080Ti. We followed the environmental requirements and hyperparameter settings of the released code implementations of the TST models to reproduce the experimental results. Table \ref{tbl:dataset_statistics} shows the training, validation, and test splits of the Yelp and GYAFC datasets used in our experiments.  

 \begin{table}[t]
\centering
\caption{Dataset statistics for Yelp and GYAFC.}
\label{tbl:dataset_statistics}
\begin{tabular}{|c|c|c|c|c|c|}
\hline
Dataset & Subset & Attributes & Train & Dev & Test \\
\hline
\multirow{2}*{Yelp} & \multirow{2}*{-} & Positive & 267,314 & 38,205 & 76,392 \\
\cline{3-6}
 & & Negative & 176,787 & 25,278 & 50,278 \\
\hline
\multirow{2}*{GYAFC} & \multirow{2}*{F\&R} & Informal & 51,967 & 2,788 & 1,332 \\
\cline{3-6}
&  & Formal & 51,967 & 2,247 & 1,019 \\
\hline
\end{tabular}
\end{table}

\textbf{Evaluation Metrics.} We adopt the evaluation metrics discussed in Section \ref{sec:evaluation} to measure the performance of the TST models. Specifically, we apply Style Transfer Accuracy (\textit{ACC}) to measure the transfer strength. To measure content preservation, we adopt \textit{BLEU}, \textit{sBLEU}, Cosine Similarity (\textit{CS}), and Word Overlap (\textit{WO}). Noted that the \textit{BLEU} score is only computed for the GYAFC dataset as the human references of the sentences in test set are available. We compute the perplexity score (\textit{PPL}) to quantify the fluency of the transferred sentences. In practice, we use the script \textbf{multi-bleu.perl} \footnote{\url{https://github.com/OpenNMT/OpenNMT/blob/master/benchmark/3rdParty/multi-bleu.perl}} to calculate \textit{BLEU} and \textit{sBLEU} scores. For \textit{WO}, we exclude both stopwords and
sentiment words \footnote{\url{https://www.cs.uic.edu/~liub/FBS/sentiment-analysis.html}}. Finally, we compute two average metrics that consider \red{all the aspects of evaluation:}
   
\begin{itemize}
    \item Geometric Mean (\textit{G-Score}): We compute the geometric mean of \textit{ACC} (transfer strength), \textit{sBLEU} (content preservation), \textit{WO} (content preservation) and \textit{1/PPL} (fluency). We excluded the measure \textit{CS} in the mean calculation due to its insensitivity and take the inverse of the calculated perplexity score because a lower \textit{PPL} score corresponds to better fluency.
    \item Harmonic Mean (\textit{H-Score}): Different averaging methods reflect different priorities. Thus, we also compute the harmonic mean of \textit{ACC}, \textit{sBLEU}, \textit{WO}, and \textit{1/PPL}.
\end{itemize}

\textbf{Reproduced Models.} We limit our reproducibility study to these 19 TST models as they had published their implementation codes. We hope to encourage fellow researchers to publish their codes and datasets to promote the development of this field. We have also grouped and color-coded the models according to the strategies discussed in Section \ref{sec:taxonomy}. \orange{Orange} represents the explicit methods of disentanglement of style and content, \green{green} denotes the methods that applied the implicit disentanglement strategy, and \blue{blue} represents the methods that performed TST without disentanglement of style and content. Specifically, we reproduced and implemented the following TST models:

\orange{\textbf{Explicit Style-Content Disentanglement}}

\begin{itemize}

    \item \orange{\textbf{DeleteOnly; Template; Del\&Retri}} \cite{li2018delete}: A style keyword replacement method, which disentangles the style and content of a sentence explicitly by keyword replacement. The authors proposed three variants of their mode: \orange{DeleteOnly}, which first remove the removed style attributed keywords from the source sentence. Subsequently, the latent representation of the source sentence is combined with the target-style attribute and input into a sequence model to generate the sentence in the target style. The \orange{Template} model simply replaces deleted style-attributed keywords with target style keywords. The \orange{Del\&Retri} model first performed the same keyword removal as \orange{DeleteOnly} method. Next, it retrieves a new sentence associated with the target attribute. Lastly, the keyword-removed source and retrieved sentences are input into a sequence model to generate the transferred style sentence.
    
    \item \orange{\textbf{B-GST; G-GST}} \cite{sudhakar2019transforming}: A TST model that extended the work in \cite{li2018delete} and  proposed the Generative Style Transformer (GST) to perform text style transfer.  There are two variants of the GST model: the Blind Generative Style Transformer (\orange{B-GST}) and the Guided Generative Style Transformer (\orange{G-GST}).
    
    \item \orange{\textbf{PTO}} \cite{wu2019hierarchical}:  A style keyword replacement TST model that applied reinforcement-learning to hierarchically reinforced a \textit{Point-Then-Operate} (\orange{PTO}) sequence operation. The \orange{PTO} operation has two agents: a high-level agent that iteratively proposes operation positions and a low-level agent that alters the sentence based on the high-level proposals. The style-attributed keywords are \red{explicitly replaced} to perform TST using this reinforcement framework.
    
    \item \orange{\textbf{UST}} \cite{xu2018unpaired}: A style keyword replacement method, which utilized cycled reinforcement learning to iteratively replace style attributed keywords while maintaining the content in text for text style transfer. This model was originally implemented for the sentiment transfer task. 
    
    \item \orange{\textbf{SMAE}} \cite{zhang2018learning}:  A style keyword replacement model to perform TST by disentangling style and content explicitly. The model was originally designed for sentiment transfer. The sentiment attribute words are first detected, and a sentiment-memory-based auto-encoder model is subsequently used to perform sentiment modification without parallel data.

All the explicit style-content disentanglement methods are discussed in detail in Section \ref{sec:tax-explicit}.

\end{itemize}

\green{\textbf{Implicit Style-Content Disentanglement}}

\begin{itemize}
    
    \item \green{\textbf{DRLST}} \cite{john2019disentangled}: An adversarial learning TST model that incorporates auxiliary multi-task and adversarial objectives for style prediction and bag-of-word prediction respectively to perform text style transfer. This method is discussed in Section \ref{sec:tax-adversarial}.
    
    \item \green{\textbf{BST}} \cite{prabhumoye2018style}: A back-translation based TST model that employed a pre-trained back translation model to rephrase a sentence while reducing its stylistic characteristics. Subsequently, separate style-specific decoders were used for style transfer. This method is discussed in Section \ref{sec:tax-backtranslation}.
    
    \item \green{\textbf{CAAE}} \cite{shen2017style}: An adversarial learning TST model that implicitly disentangle the text's style. Specifically, the model assumes a shared latent content distribution across different text corpora and proposes a method that leverages refined alignment of latent representations to perform TST. This method is discussed in Section \ref{sec:tax-adversarial}.
    
     \item \green{\textbf{Ctrl-Gen}} \cite{hu2017toward}: An attribute-controlled TST model that used variational auto-encoders and style classifier to guide the learning of a style attribute to control the generation of text in different styles. This method is discussed in Section \ref{sec:tax-controllable-disentanglement}.
    
    \item \green{\textbf{ARAE}} \cite{zhao2018adversarially}: A generic natural language generation technique that utilizes adversarial learning to modify the specific attributes in text. TST is one of the model's applications proposed in its original paper. This method is discussed in Section \ref{sec:tax-adversarial}.
    
    \item \green{\textbf{Multi-Dec; Style-Emb}} \cite{fu2018style}: An adversarial learning TST model that utilized a style classifier to perform disentanglement of style and content representation for the style transfer task. Two variants of the model were proposed: The multi-decoder (\green{Multi-Dec}) model that uses different decoders to generate text with different styles. The style embedding (\green{Style-Emb}) model concatenates the style embedding vector with content representation to generate different style text with one decoder. This method is discussed in Section \ref{sec:tax-adversarial}.
    
\end{itemize}

\blue{\textbf{Without Style-Content Disentanglement}}
\begin{itemize}

    \item \blue{\textbf{DualRL}} \cite{luo2019dual}: A reinforcement learning-based TST model that utilized two seq2seq models to transfer between two text styles. Specifically, this model considers the learning of source-to-target style and target-to-source style as a dual-task that mutually reinforces each other to perform TST without disentangling style and content. This method is discussed in Section \ref{sec:tax-Reinforcement}. 
    
    \item \blue{\textbf{DAST; DAST-C}} \cite{li2019domain}: An attribute-controlled TST model that performs TST in a domain-aware manner. Two variants are proposed: The Domain Adaptation Style (\blue{DAST}) model and \blue{DAST} with generic content information (\blue{DAST-C}). Latent style attributes and domain vectors are learned in these models to perform TST across domains. This method is discussed in Section \ref{sec:tax-controllable-without}. 
    
    \item \blue{\textbf{PFST}} \cite{He2020A}: A probabilistic deep generative TST model. The model combines a language model prior and an encoder-decoder transducer to infer the sentence's latent representations in an assumed parallel-style corpus.  The inferred latent representations are subsequently used to generate a sentence of a specific style via a decoder. This method is discussed in Section \ref{sec:tax-other}.
    
\end{itemize}
    

\subsection{Sentiment Transfer}


Table \ref{tbl:resultyelp} shows the performance of various TST models in the sentiment transfer task. We observe that no TST models achieved the best performance in all evaluation metrics.  \blue{DualRL}, \orange{PTO}, \orange{B-GST}, and \blue{PFST} have achieved a well-balanced trade-off between text fluency, content preservation, and style transfer accuracy. \green{DRLST} has achieved the second-best transfer accuracy. However, the model also suffers from a very low \textit{sBLEU} score, suggesting the ineffectiveness of \green{DRLST} in preserving the content of the original sentence. Moreover, we observe that most implicit style-content disentanglement methods have poor performance on content preservation. A potential reason could be that some content information may be lost while performing disentanglement of style and content.
We also note that the different average methods, that is, \textit{G-Score} and \textit{H-Score}, weighted the different evaluation metrics differently. For example, \textit{H-score} gives a higher weight to the perplexity scores of the generated sentences. Thus, \green{DRLST}, which has the lowest \textit{PPL} score, also has the highest \textit{H-score}. \red{On the contrary}, the \orange{Template} model has the highest \textit{PPL} score and the lowest \textit{H-score}.

More interestingly, we observed that style keyword replacement methods such as \orange{DeleteOnly}, \orange{Template}, \orange{Del\&Retri}, \orange{B-GST}, \orange{G-GST}, \orange{UST}, \orange{PTO}, and \orange{SMAE}, have achieved good performance in sentiment transfer tasks. Methods have achieved a high transfer accuracy score while preserving content information, that is, high scores of \textit{sBLEU}, \textit{CS}, and \textit{WO}. A possible reason for the style keyword replacement methods' good performance might be due to the nature of the task; the sentiment of a sentence can be easily modified by replacing keywords related to the source sentiment. For example, replacing ``fresh'' with ``rotten'' would transform the sentiment from positive to negative. However, it is interesting to note that the \orange{Template} method, which is an algorithm that replaces the sentiment-related keywords, has a high perplexity score, which indicates bad performance in sentence fluency. This motivates more complex generative approaches that can prevent the generation of implausible sentences by simple keyword replacement.

\begin{table*}[t]
\centering
\caption{TST results in Yelp dataset (sentiment transfer task).}
\label{tbl:resultyelp}
\begin{tabular}{cccccccccc}
\hline
Model & \textit{ACC(\%)} & \textit{sBLEU} & \textit{CS} & \textit{WO} & \textit{PPL} & \textit{G-Score} & \textit{H-Score} & \#Params & FLOPs \\\hline
\orange{DeleteOnly} &84.2 &28.7 &0.893 &0.501  &115  &1.80  &0.034 & 9.18M  & 27.98G \\
\orange{Template}  &78.2 &48.1 &0.850 &0.603 &1959  &1.04  &0.002 & - & - \\
\orange{Del\&Retri}  &88.1 &30.0 &0.897 &0.464 &101  &1.87  &0.039 & 9.88M & 29.78G \\
\orange{B-GST} &89.2 &46.5 &0.959 &0.649  &216  &1.88  &0.018 & - & - \\
\orange{G-GST} &72.7  &52.0  &0.967  &0.617  &407  &1.55  &0.010 & - & - \\
\orange{PTO} &82.3 &57.4 &\textbf{0.982} &0.737  &245  &1.94  &0.016 & - & - \\
\orange{UST} &74.0  &41.0  &0.929  &0.448  &394  &1.36  &0.010 & 79.97M  & 2390.6G \\
\orange{SMAE} &84.4  &14.8  &0.907  &0.294  &210  &1.315  &0.019 &76.27M & 4418.5G \\
\hline
\green{DRLST} &91.2 &7.6 &0.904  &0.484  &\textbf{86} & 1.41 & \textbf{0.045} &64.0M &0.149G \\
\green{BST} &83.1 &2.3 &0.827 &0.076 &261  &0.49  &0.015 & - & - \\
\green{CAAE} &82.7 &11.2 &0.901 &0.277  &145  &1.15  &0.027 & 12.92M & 0.436G \\
\green{ARAE} &83.2 &18.0 &0.874 &0.270  &138  &1.31  &0.028 & 4.20M & 5.61G \\
\green{Ctrl-Gen} &89.6 &49.5 &0.953 &0.707  &384  &1.69 &0.010 & 13.7M & 0.153G \\
\green{Multi-Dec} &69.6 &17.2  &0.887  &0.244  &299  &0.99  &0.013 & - & - \\
\green{Style-Emb} &47.5 &31.4  &0.926  &0.433  &217  &1.31  &0.018 & - & - \\
\hline                                                                    
\blue{DualRL}  &79.0 &\textbf{58.3} &0.97 &\textbf{0.801} &134 & \textbf{2.29}  &0.030 & 44M & 0.862G \\
\blue{DAST} &90.7  &49.7  &0.961  &0.705  &323  &1.77  &0.012 &26.68M & 0.822G\\
\blue{DAST-C} &\textbf{93.6}  &41.2  &0.933  &0.560  &450  &1.48  &0.009 &24.51M & 0.508G\\
\blue{PFST} &85.3  &41.7  &0.902  &0.527  &104  &2.06  &0.038 & 34.32M & 76.19G \\
\hline
\end{tabular}
\end{table*}

\subsection{Formality Transfer}

Table \ref{tbl:resultGYAFC} shows the performance of various TST models in the formality transfer task. Similar to the observation in sentiment transfer, none of the TST models can score well on all evaluation metrics. We noted that the average style transfer accuracy in GYAFC is 52.9\%, which is significantly lower than Yelp's average score of 84.4\%. This highlights the difficulty of the formality transfer task. We also observe that most models performed worse on this task compared to the sentiment transfer task. It is not surprising that the keyword replacement methods did not perform well in the formality transfer task; most of these models achieved low style transfer accuracy. Some of the adversarial learning-based TST models, such as \green{CAAE} \cite{shen2017style} and \green{DRLST} \cite{john2019disentangled}, had achieved high style transfer accuracy, but very low content preservation, as these models lack the mechanism to control content preservation during the generative process. Interestingly, we observe that the attribute-controlled TST methods, i.e., \green{Ctrl-Gen} \cite{hu2017toward}, \blue{DAST} \cite{li2019domain}, and \green{DAST-C} \cite{li2019domain} have achieved good performance both style transfer accuracy and content preservation. Similar to sentiment transfer, we observe that the implicit style-content disentanglement methods, i.e., \green{DRLST}, \green{BST}, \green{CAAE}, \green{ARAE}, and  \green{Multi-Dec}, also performed poorly on content preservation for the formality transfer task. Specifically, the scores of content preservation metrics are even lower than that in the sentiment transfer task. Methods that perform TST tasks without style-content disentanglement have achieved the best performance for the formality transfer task.

The GYAFC dataset provided the performance of four human references performing the formality transfer task on the test dataset (as shown at the bottom of Table \ref{tbl:resultGYAFC}). On average, the human references had achieved 78.1\% style transfer accuracy. This is considered a good performance, since the pre-trained binary classifier only managed to achieve 83.4\% accuracy on the test set. Furthermore, text formality is subjective, and the four human references may have different opinions on the degree of formality in text. 

As the GYAFC dataset is a parallel dataset, i.e., there are matching sentences in source and target styles, we can compute the \textit{hBLEU} score between the transferred style sentence and the matching sentence in the target style. Unsurprisingly, human references have achieved the highest \textit{hBLEU} score, suggesting that the sentences generated by human references are quite similar to the matching sentences in the target style. In comparison, the TST models perform poorly on \textit{hBLEU} scores. We also observe that the TST models' average content preservation metrics scores in the formality transfer task are lower than the scores in the sentiment transfer task. For example, \textit{WO} scores in the sentiment transfer task are higher because only a few keywords need to be replaced to perform the style transfer. However, in the formality transfer case, a more drastic and complex modification of the text has to be performed for the style transfer. As such, there will be less word overlap between the original sentence and the transferred sentence, resulting in a lower \textit{WO} score. The limitation of existing metrics in measuring content preservation in formality transfer highlights the need to search for better evaluation methods for this challenging task.

\begin{table*}[t]
\centering
\caption{TST results in the GYAFC dataset (formality transfer task).}
\label{tbl:resultGYAFC}
\begin{tabular}{cccccccccccc}
\hline
Model & \textit{ACC(\%)} & \textit{sBLEU} & \textit{hBLEU} & \textit{CS} & \textit{WO} & \textit{PPL} & \textit{G-Score} & \textit{H-Score} & \#Params & FLOPs \\\hline
\orange{DeleteOnly} &26.0 &35.4 &16.2 &0.945 &0.431  &82  &1.48  &\textbf{0.047} & 9.19M & 27.89G  \\
\orange{Template}  &51.5 &45.1 &19.0 &0.943 &0.509 &111  &1.81  &0.035 & - & - \\
\orange{Del\&Retri}  &50.6 &22.1 &11.8 &0.934 &0.345 &94  &1.42  &0.041 & 9.99M & 29.78G \\
\orange{B-GST} &30.3 &22.5 &11.6 &\textbf{0.951} &\textbf{0.557}  &117  &1.34  &0.034 &- & - \\
\orange{G-GST} &31.0 &20.7 &10.2 &0.941  &0.556  &127  &1.29  &0.031 & - & - \\
\orange{UST} & 23.6  &0.5 &0.5 &0.881  &0.012  &\textbf{28}  &0.27  &0.035 &79.97M  & 2390.6G \\
\orange{SMAE} &21.6  &6.5 &1.2  &0.898  &0.079  &74  &0.62  &0.046 &76.27M & 4418.5G\\
\hline
\green{DRLST} &71.1 &4.2 &2.7 &0.909  &0.342  &86 & 1.04 & 0.045 & 64.0M & 0.149G \\
\green{BST} &69.7 &0.5 &0.5 &0.883 &0.04 &69  &0.38  &0.042 & - & - \\
\green{CAAE} &72.3 &1.8 &1.5 &0.896 &0.028  &55  &0.51  &0.044 & 12.92M & 0.436G \\
\green{ARAE} &76.2 &4.8 &2.2 &0.903 &0.042  &77 &0.67  &0.040  & 4.20M & 5.61G \\
\green{Ctrl-Gen} &73.1 &57.0 &15.6 &0.943 &0.446  &168  &1.82 &0.023 & 13.7M & 0.153G \\
\green{Multi-Dec} &22.2 &13.4 &5.9 &0.911  &0.168  &146  &0.76  &0.026 & - & - \\
\green{Style-Emb} &27.7  &8.3 &3.6 &0.897  &0.102  &136  &0.64  &0.027 & - & - \\
\hline
\blue{DualRL}  &56.7 &\textbf{61.6} &18.8 &0.944 &0.447 &122  &\textbf{1.89}  &0.032 & 44M & 0.862G \\
\blue{DAST} &73.1  &50.6 &14.3 &0.934  &0.350  &204  &1.59  &0.019 & 26.68M & 0.822G \\
\blue{DAST-C} &78.2  &48.5 &13.8 &0.927  &0.328  &308  &1.42  &0.013 & 24.51M & 0.508G \\
\blue{PFST} &50.8  &55.3 &16.5 &0.940  &0.466  &200  &0.51  &0.020 & 34.32M & 76.19G \\
\hline
Human0 &78.1  &20.5 &43.5 &0.942  &0.393  &80  &1.67  &0.048 & - & - \\
Human1 &\textbf{78.7}  &18.2 &43.2 &0.931  &0.342  &199  &1.25  &0.020 & - & - \\
Human2 &78.2  &18.6 &43.4 &0.932  &0.354  &192  &1.28  &0.021 & - & - \\
Human3 &77.4  &18.8 &43.5 &0.931  &0.354  &196  &1.27  &0.020 & - & - \\
\hline
\end{tabular}
\end{table*}

\subsection{Computational Efficiency Evaluation}

We have also reported the computational resources needed to train the TST models in Tables \ref{tbl:resultyelp} and \ref{tbl:resultGYAFC}. Specifically, we compute the number of parameters textit{\#Params} and the floating-point operations \textit{FLOPs} of the various TST models. For Tensorflow framework-based models, we use \textbf{tf.trainable\_variables()} function to compute the number of parameters, and \textbf{tf.profiler.profile()} function to compute the \textit{FLOPs}. For Pytorch framework-based models, we use a third-party library name THOP \footnote{https://github.com/Lyken17/pytorch-OpCounter} to computer \#Params and FLOPs. Note that the calculated \#Params and FLOPs are approximations, as the built-in functions of Tensorflow and THOP cannot evaluate the custom functions defined by the model developers. Furthermore, we cannot report the metrics for some of the TST models due to various technical constraints. For instance,  orange{Template} is rule-based model, \orange{B-GST} and \orange{G-GST} use the pre-trained language model GPT \cite{radford2018improving}, and the Pytorch version of \green{BST} is too old. We are also unable to retrieve the metrics for \green{Multi-Dec} and \green{Style-Emb} since the models used the deep learning framework Theano. 

For TST methods that explicitly disentangled style and content, we observed that \orange{DeleteOnly} and \orange{Del\&Retri} have relatively lower \#Params and FLOPs compared to .\orange{UST} and \orange{SMAE}. This could be attributed to the technique used to perform the explicit style and content disentanglement. \orange{DeleteOnly} and \orange{Del\&Retri} have used a token frequency-based method to remove style-attributed keywords from a given sentence before performing keyword replacement \cite{li2018delete}. On the other hand, \orange{UST} and \orange{SMAE} trained an LSTM model with an attention mechanism to identify and remove style-attributed keywords in a sentence. The sequence models to perform the style-attributed keyword removal process added more \#Params needed to train the TST two models. \green{DRLST} has a relatively higher \#Params compared to the other TST methods that performed TST methods that implicitly disentangled style and content. This high \#Params could be attributed to the multi-task learning objectives in \green{DRLST} \cite{john2019disentangled}. \blue{DualRL} has the highest \#Params among methods that perform TST without style-content disentanglement because of the complex neural network architecture; the model needs to learn and map two Seq2Seq models to perform TSTs \cite{luo2019dual}.

\begin{figure*}[t] 
	\centering
	\setlength{\tabcolsep}{0pt} 
	\renewcommand{\arraystretch}{0} 
	\begin{tabular}{cccc}
		\includegraphics[scale = 0.5]{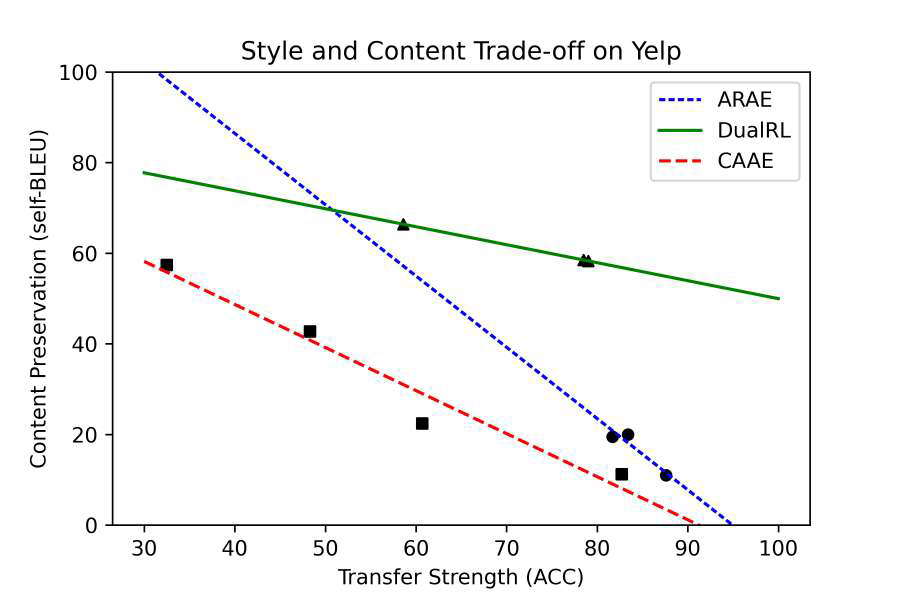} &  
        \includegraphics[scale = 0.5]{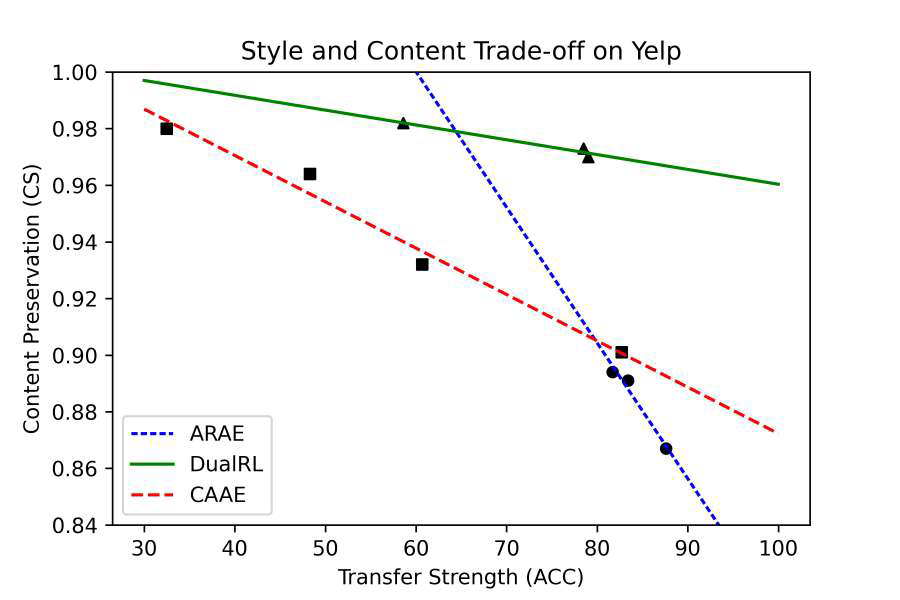} & \\
        \includegraphics[scale = 0.5]{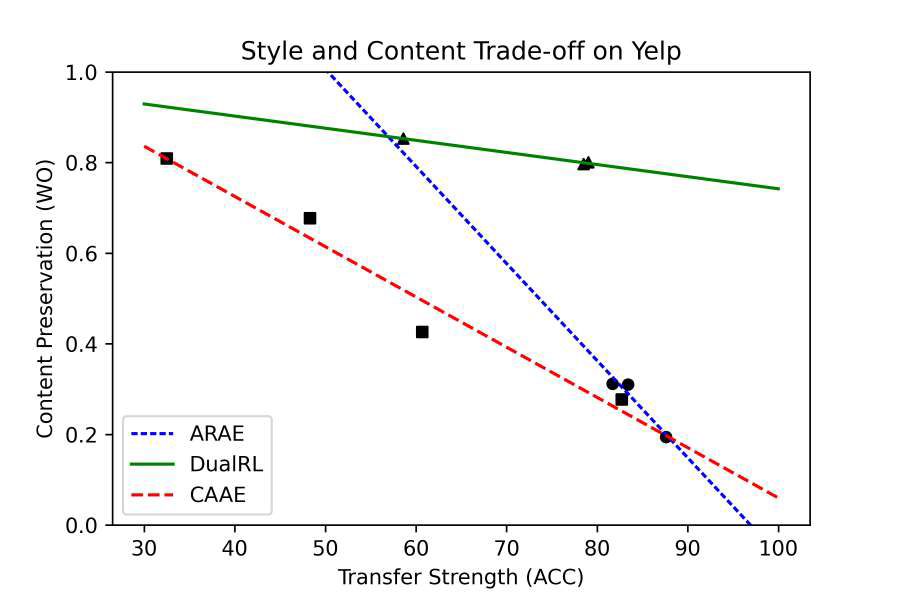} & 
        \includegraphics[scale = 0.5]{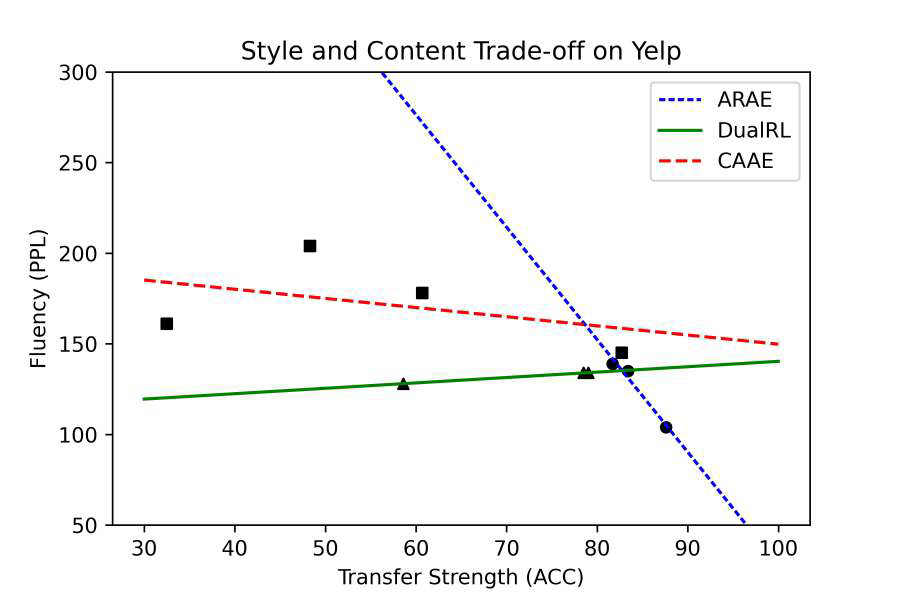} & \\
	\end{tabular}
	\caption{Metrics trade-off analysis for sentiment transfer on Yelp review dataset}
	\label{fig:transfer_strength_content_preservation_yelp}
\end{figure*}

\begin{figure*}[t] 
	\centering
	\setlength{\tabcolsep}{0pt} 
	\renewcommand{\arraystretch}{0} 
	\begin{tabular}{cccc}
		\includegraphics[scale = 0.5]{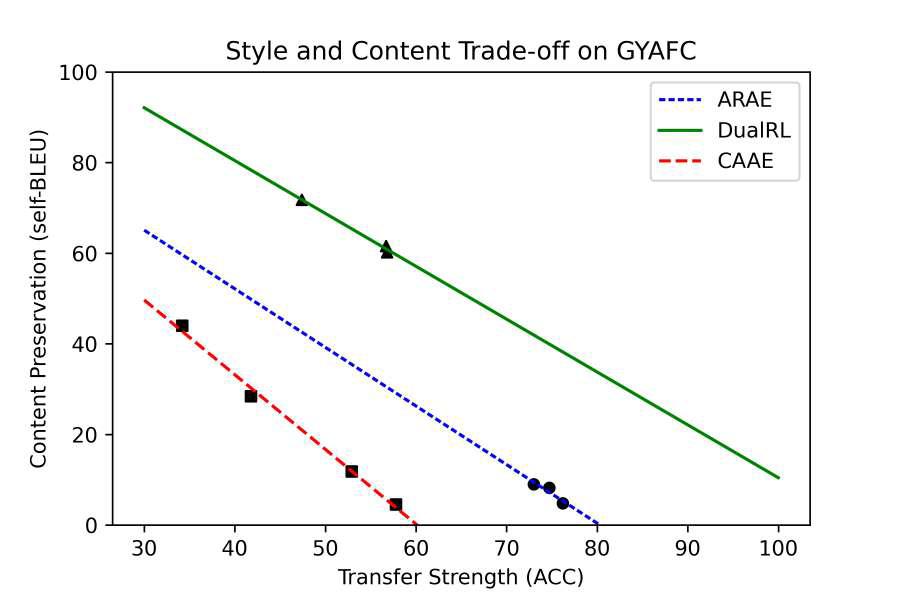} &  
        \includegraphics[scale = 0.5]{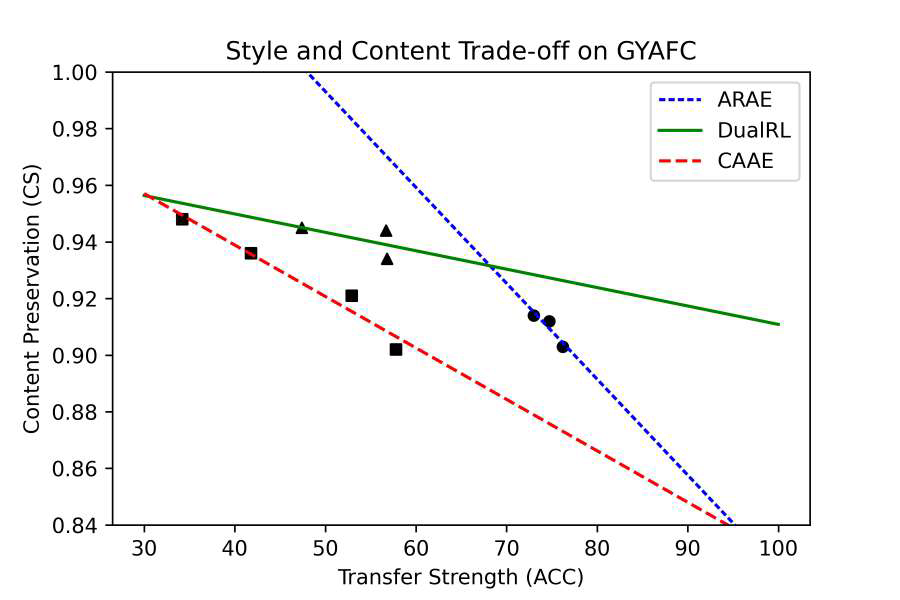} & \\
        \includegraphics[scale = 0.5]{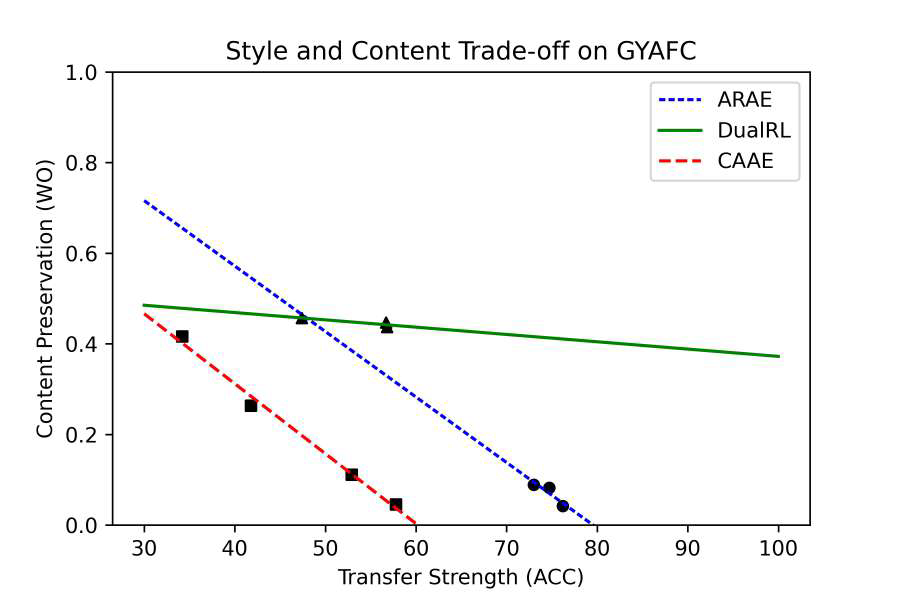} & 
        \includegraphics[scale = 0.5]{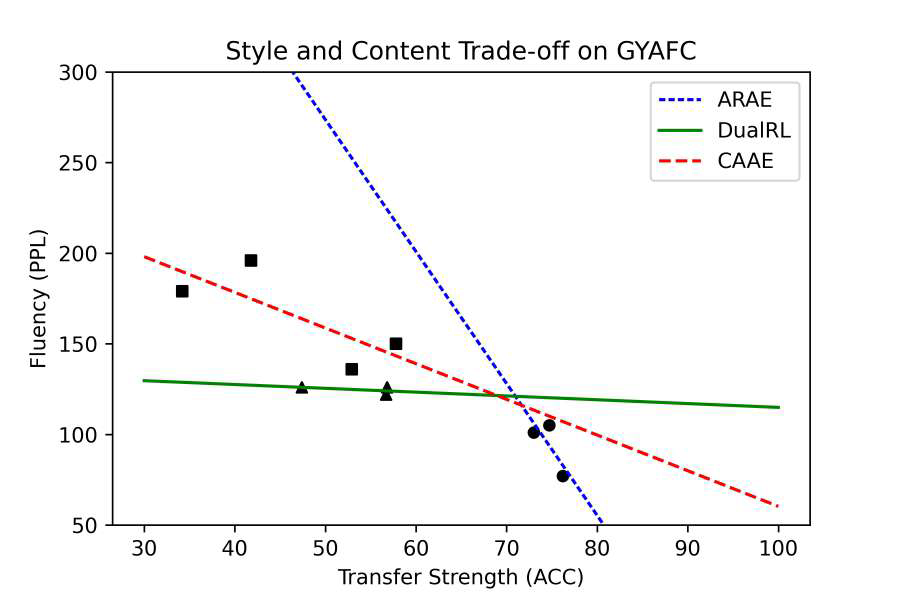} & \\
	\end{tabular}
	\caption{Metrics trade-off analysis for formality transfer on GYAFC review dataset}
	\label{fig:transfer_strength_content_preservation_GYAFC}
\end{figure*}

\subsection{Evaluation Metrics Trade-off Analysis}
Besides evaluating the TST models in the two style transfer tasks, we also reproduced the evaluation metrics trade-off analysis proposed in \cite{mir2019evaluating}. Specifically, we create variants of TST models by varying their hyperparameters and studying the trade-off effects between pairs of evaluation metrics. We selected models \green{ARAE}, \blue{DualRL}, and \green{CAAE} for our analysis. 

The trade-off between style transfer accuracy and content preservation has been discussed in existing TST studies \cite{fu2018style,mir2019evaluating}. For example, Fu et al. \cite{fu2018style} performed the trade-off analysis between style transfer accuracy and content preservation of \green{Multi-Dec} and \green{Style-Emb} for the sentiment transfer task. Similar experiments were also conducted in \cite{mir2019evaluating} to evaluate the trade-off between content preservation, style transfer accuracy, and fluency of the models \green{CAAE}, \green{ARAE}, and \orange{Del\&Retri} for the sentiment transfer task. However, no known study performs the trade-off analysis for the formality transfer task. \red{The purpose of this section is to fill this gap by performing and comparing trade-off analysis on the sentiment transfer and formality transfer tasks.}  


Fig. \ref{fig:transfer_strength_content_preservation_yelp} \red{shows the results of the trade-off analysis in the sentiment transfer task. Specifically}, Fig. \ref{fig:transfer_strength_content_preservation_yelp}A,B and C show the trade-off relationships between style transfer strength and content preservation metrics, while Fig. \ref{fig:transfer_strength_content_preservation_yelp}D shows the trade-off relationship between style transfer strength and fluency metric. Similar to the observations made in \cite{mir2019evaluating}, we notice that as the transfer strength increases, content preservation metrics decrease across the three models. However, the trade-off relationship between the transfer strength and sentence fluency is less obvious as we notice that \green{ARAE} can achieve lower \textit{PPL} when \textit{ACC} increases. Similar observations are made for the formality transfer task in Fig. \ref{fig:transfer_strength_content_preservation_GYAFC}. Interestingly, we noted that even when content preservation is reduced to the lowest,  it could not achieve a high style transfer accuracy for the formality transfer task.

The observations made in our trade-off analysis are consistent with previous observations with those made in previous studies \cite{fu2018style,mir2019evaluating}. Specifically, when transfer strength scores increase, content preservation scores decrease, and vice versa. A potential reason for observation might be the entanglement between semantic and stylistic properties in natural language; it is hard to separate the two properties, and changing one affects the other. Therefore, when optimizing to transfer the style in text, it is hard to maintain the sentence's semantic, i.e., the content information. 

\subsection{Human Evaluation}

We conducted a human-based evaluation study to further evaluate the TST models' performance. We focus our human evaluation on representative models from the three TST strategies, and the models are evaluated on sentiment transfer and formality transfer tasks.

We first randomly sampled 50 negative and 50 positive sentences from the Yelp dataset for the sentiment transfer task. Next, we perform TST for the sampled sentences using \orange{PTO}, \green{DRLST} and \blue{DualRL}, which are the best performing models from the various TST  strategies when evaluated using \textit{G-Score}. Similarly, we randomly sampled 50 formal and 50 informal sentences from the GYAFC dataset for the formality transfer task. Subsequently, we perform TST for the sampled sentences using orange{Template}, \green{DRLST} and \blue{DualRL}, which are the best performing representative models for the formality transfer task.

We recruited six linguistic researchers (i.e., evaluators) to evaluate the style-transferred sentences generated by the TST models.  Each evaluator is assigned to evaluate pairs of sentences: the source sentences and style-transferred sentences. The evaluators were asked to evaluate the style-transferred sentences on the three criteria discussed in the earlier section:

\begin{enumerate}
    \item \textbf{Transfer Strength:} The evaluators were asked to rate the style of the transferred sentences using a 5-point Likert scale. 1: strongly opined that the sentence is in source style (e.g., positive sentiment), and 5: strongly opined that the sentence is in target style (e.g., negative sentiment).
    \item \textbf{Content Preservation:} The evaluators were asked to compare the given source sentence and transferred sentence and rate the amount of content preserved in the transferred sentence using a 5-point Likert scale. 1:  source and transferred sentences cover absolutely different content, and 5: source and transferred sentences cover the absolutely same content.
    \item \textbf{Fluency:} The evaluators were asked to rate the fluency of the transferred sentences using a 5 Likert scale. 1: unreadable with too many grammatical errors, 5: perfect and fluent sentence.
\end{enumerate}

To minimize biases, we randomly assign every style-transferred sentence to two evaluators, and the models' names will not be displayed to the evaluators.

\begin{table*}[t]
\small
\caption{Human evaluation results of selected TST models in the Yelp and GYAFC datasets.}
\centering
\begin{tabular}{cccccc|ccccc}
\hline
&\multicolumn{5}{c}{Yelp}&\multicolumn{5}{c}{GYAFC}\\
\hline
Model & \textit{Style} & \textit{Content} & \textit{Fluency} & \textit{G-Score} & \textit{H-Score} & \textit{Style} & \textit{Content} & \textit{Fluency} & \textit{G-Score} & \textit{H-Score} \\
\hline
\orange{PTO} & 3.88 & 3.06 & 3.24 & 3.37 & 3.35 & - & - & -  & - & - \\
\orange{Template} & - & - & - & - & - & 3.03 & \textbf{3.74} & 3.29 & 3.34  & 3.33 \\
\green{DRLST} & \textbf{4.26} & 2.27 & 4.22 & 3.44 & 3.29 & 3.77 & 1.86 & \textbf{3.94} & 3.02 & 2.84 \\
\blue{DualRL} & 3.84 & \textbf{3.59} & \textbf{4.26} & \textbf{3.88} & \textbf{3.87} &\textbf{3.81} & 3.55 & 3.58 & \textbf{3.64} & \textbf{3.64} \\
\hline
\end{tabular}
\label{tbl:human_evaluation}
\end{table*}

Table ~\ref{tbl:human_evaluation} shows the results of our human evaluation experiment. We compute the models' average 5-point Likert scores for transfer strength, content preservation, and fluency criteria. We also report the geometric mean score (\textit{G-Score}) and harmonic mean score (\textit{H-Score}) of the three evaluated criteria.

Examining the results of the sentiment transfer task on the Yelp dataset, we observe that the \green{DualRL} model can achieve the best and balanced performance. This demonstrated \green{DualRL}'s ability to generate fluent sentences in the target style, while \red{maintaining a good balance between style transfer strength and content preservation. \green{DRLST} is observed to achieve the best transfer strength performance.} However, the model has performed poorly on content preservation, suggesting that the model can generate sentences in the target style, but the content differs significantly from the source sentence. This observation is consistent with the automatic evaluation, since \green{DRLST} has a low \textit{sBLEU} score. The explicit style keyword replacement model \orange{PTO} can achieve competitive performance on transfer strength and content preservation but can not generate fluent sentences as the other two models. 

For the formality transfer task on the GYAFC dataset, we observed that \orange{Template} had achieved the best content preservation among the compared TST models. However, the model has worse performance in transfer strength. \red{Similarly to} the sentiment transfer task, \blue{DualRL} achieved the best performance in transfer strength, \textit{G-Score}, and \textit{H-Score}. \green{DRLST} has good performance in transfer strength and fluency but is unable to preserve content information during the TST process.

\section{Applications}
\label{sec:application}
The research on TST algorithms has many industrial applications and could lead to many commercial benefits. In this section, we summarize these applications and present some potential usages. 

\subsection{Writing Tools}
One of the industrial applications of TST algorithms is the design of writing tools. \red{Academics in various domains have widely researched computer-aided writing tools,} and the industry has developed many applications of writing tools \cite{klahold2020word,klahold2020computer,silva2019computational,snyder1993writing,parra2019automated,macarthur2009reflections}. The TST methods can be applied as new valuable features in existing writing tool applications.

The utility of writing style has been widely studied by linguistic, and literacy education scholars \cite{can2004change,young2002technical,halliday1981linguistic,johnstone2009stance,ashok2013success,ivanivc2004discourses}. The TST algorithms enable writing tool applications to apply the insights from existing linguistics studies to improve the writings of users. For instance, applying TST algorithms enables writing tool users to switch between writing styles for different audiences while preserving their writing content. The style evaluation methods developed to evaluate TST algorithms can also be applied to analyze \red{user writing style} \cite{parra2019automated}. For instance, the writing tool could analyze the style of users' business email draft to be too informal and recommend the users modify their writing to make the writing style more formal. Previous studies \cite{kajiwara2016building,nisioi2017exploring,saggion2017automatic,nishida2019multi,cao2020expertise} have developed interesting real-world TST applications, where the texts are transferred between expert and layman styles. The underlying intuition is that expertise style transfer aims at improving the readability of a text by reducing the expertise level, such as explaining the complex terminology with simple phrases. On the other hand, it also aims to improve the expertise level based on context so that laymen's expressions can be more accurate and professional.

\subsection{Persuasive Communication and Marketing}
Studies have explored \red{the use of} persuasive text to influence the attitude or behaviors of people \cite{kaptein2012adaptive,chambliss1996adults}, and the insights gained from these studies have also been applied to improve marketing and advertising in the industry. The style of a text has an impact on its persuasiveness \cite{darani2014persuasive,muehlenhaus2012if,johnstone1989linguistic}, and the TST algorithms can be used to convert a text into a more persuasive style. Recent studies have also explored personalizing persuasive strategies \red{based on the user profile} \cite{kaptein2015personalizing}. Similarly, TST algorithms could also be used to structure the text in different persuasive text styles that best appeal \red{to} user profiles. For instance, TST algorithms can be applied to modify a marketing message into an authoritative style for users who appeal to authority. Jin et al. \cite{jin2020hooks} proposed a compelling use case \red{for using} TST methods to \red{make the} news headlines more attractive. Specifically, a TST algorithm is used to transfer news headlines between humor, romance, and clickbaity style.

\subsection{Chatbot Dialogue}
\red{The research and development of chatbots, i.e. intelligent dialogue systems that are able to engage in conversations with humans, have been one of the longest-standing goals in artificial intelligence} \cite{abdul2015survey,xu2017new,zhou2020design}. Kim et al. \cite{kim2019comparing} conducted a study on the impact of chatbot's conversational style on users.
They found that when an informal conversational style was adopted, participants in the experiment
were less likely to be persuaded to \red{take an action than
those conversing with a formal conversational style chatbot}. The encouraging results of the study suggest that the conversational styles of a chatbot can influence a user, and the TST algorithms could be exploited to \red{enhance chatbot flexibility in conversational styles.}.  
TST algorithms can be applied to equip chatbots with the ability to switch between conversational styles, \red{making chatbots more appealing and engaging to users}. \red{For example, a chatbot that recommends products} to customers may adopt a more persuasive conversational style, while the same chatbot may switch to a formal conversational style when \red{addressing customer complaints}.

\section{Ethical Considerations}
\label{sec:ethics}
With the recent increase of attention on artificial intelligence research's ethical issues, it is pertinent to discuss TST research's ethical considerations. This section highlights the potential ethical issues of TST applications and offers some guidelines to avoid ethical misconduct in future TST research.

\subsection{Negative Use-Cases}
The natural language processing (NLP) community has initiated thought-provoking discussions on the ethical issues of NLP research and applications. Hovy et al. \cite{hovy2016social} discussed the social implications of NLP technology and research and underlined their ethical significance. The ethical discussion on other NLP technology is also applicable to TST research. For instance, TST researchers need to be wary that the TST techniques created can have unintended negative consequences. While TST techniques can be applied to many exciting and beneficial settings mentioned in Section \ref{sec:application}, there are potential negative use-cases if the developed TST techniques are misused.

\textbf{Content Manipulation and Forgery.} TST method could be misused to perform malicious content manipulation and forgery. For example, TST methods can be applied to manipulate the polarity of reviews, such as Yelp and Amazon reviews; fraudulent sellers can apply the TST method to transfer negative reviews on their products to generate fake positive reviews and generate negative views on their competitors. Similarly, malicious users can apply TST techniques to forge documents and content based on a specific user's writing styles. This posts challenges to forensic investigation efforts in detecting document forgery \cite{iqbal2010mining,brocardo2013authorship,hu2020deepstyle}.

\textbf{Social Bots and Sock-Puppets.} TST methods could also be applied in social media. For example, the political slant transfer task~\cite{prabhumoye2018style} may raise ethical issues if applied in social bots to generate social media posts advocating certain political ideas and manipulating the political views of the masses. Similarly, TST could also enable sock-puppets or create an army of social bots with different persuasive styles to promote anti-social behaviors such as online hate speeches~\cite{fortuna2018survey} or cyberbullying against individuals or groups~\cite{salawu2017approaches}. 

\subsection{Ethical Guidelines}
Researchers and companies must raise awareness of these potential misuses of the developed TST techniques. Specifically, companies and research institutes should consider including an ethical review process to examine the TST research and development of TST applications. Leidner et al. \cite{leidner2017ethical} proposed a framework to encourage organizations to build NLP applications with ethical considerations, and this framework could also be applied to TST research. Specifically, companies should set up an Ethics Review Board (ERB) to examine the ethical issues before, during, and after executing the TST research and development. Most research institutes and universities have ERBs, and we encourage ERBs to include experts in artificial intelligence and NLP when examining ethical issues in the proposed TST research. By having ethics stakeholders participating in various stages of research and development, we hope that TST researchers could build ethics considerations into their studies and mitigate ethical concerns. 

\section{Future Research Direction and Open Issues}
\label{sec:challanges}
TST is a relatively new research area. While existing works have established a foundation for TST research, this section outlines several promising future research directions. We also discuss open issues that we believe are critical to the field's present state.

\subsection{Deeper Dive into Style-Content Disentanglement}
As discussed in Section 4.1.2, \red{text style and content disentanglement remains an open research question.} Researchers who intend to propose new TST methods that perform style and content disentanglement must design new experiments to demonstrate or quantify the extent of disentanglement. Style and content disentanglement should also be investigated in different TST tasks, e.g. sentiment and formality transfer tasks.

More studies could also be done on the style embeddings learned by the existing techniques. Currently, we have little understanding of the style representations learned using existing techniques; In addition to \red{knowing that} the style representation supposedly has some correlation with the style labels, we do not know much about the information preserved in the style representations. For example, in learning the style representations for formality transfer tasks, little is known about the preservation of the sentence structure in the representations, and the sentence structure may impact the formality of the text. To this end, a potential future research direction would be to conduct a deeper analysis of the style representations learned for the different tasks using existing techniques. We believe that this will provide new insights, which can guide the development of future TST techniques.

\subsection{Unsupervised Text Style Transfer}
While most of the emerging TST methods are developed for the non-parallel dataset setting, these techniques require a large amount of style-labeled data to guide the transfer of text styles. A promising research direction would be to explore unsupervised methods to perform TST with little or no labeled data. For instance, recent studies \cite{jain2019unsupervised,gong2019reinforcement} have explored guiding the transfer of text style by scoring the sentences' semantic relatedness, fluency, and readability grade instead of the style labels. We postulate that more aspects of the text, such as tone, brevity, sentence structures, etc., can be explored to train future TST models and reduce the dependence on the style labels.

\subsection{Going Beyond Transferring Between Two Styles}
Currently, most existing TST methods focus on transferring the text between two styles. We believe that TST studies should go beyond performing a binary-style transfer and explore more prosperous and more dynamic tasks. For example, Lai et al. \cite{lai2019multiple} \red{proposed a multiple attribute style transfer task where a text is transferred by specifying multiple attribute styles such as sentiment, author gender,} etc. Domain-aware TST method has also been explored where we consider the domain of the text (e.g., food or movie reviews) when transferring the text styles (e.g., from positive to negative sentiment). More dynamic TST tasks with better real-world applications will be a promising future research direction.

\subsection{Style in other languages}
Most of the existing TST models were applied to English corpora, neglecting \red{the potential of TST in other languages}. Different languages may have their own unique text style properties. Therefore, novel TST methods should be designed to capture the language-specific stylistic properties. Mizukami et al. \cite{mizukami2015linguistic} designed a dialogue system that modeled \red{the text style of individual Japanese writers} using a statistical machine translation-based technique. However, there is no recent exploration to perform TST in non-English languages, and we would encourage TST researchers to address this research gap. Moreover, exploring the style features in different languages may improve our understanding of text style and style representations.

\subsection{Automatic Evaluation for Text Style Transfer}
Our experimental evaluation in Section \ref{sec:experiment} has illustrated the challenges of evaluating the effectiveness of TST models. The existing evaluation methods have some limitations. First, the evaluation of text style transfer based on transfer accuracy is limited by the performance of the style classifier. Second, similar to previous studies \cite{fu2018style,pang2019unsupervised}, we note that the transfer strength is inversely proportional to content preservation, suggesting that these metrics may be complementary and challenging to optimize simultaneously. The limitations of existing evaluation metrics motivate the exploration of novel automatic evaluation metrics to evaluate TST models.

\section{Discussion and Conclusion}
\label{sec:conclusion}
Although TST is a relatively new branch of \red{the field of natural language processing}, considerable research on TST has recently been conducted. The explosive growth of TST research has generated many novel and interesting TST models. This survey aims to organize these novel TST models using a  taxonomy (cf.~Fig. \ref{fig:taxonomy}). We also summarize the common techniques used by modern TST models to transfer text styles. We also emphasize the critical TST research trends, such as the shift from TST models attempting to disentangle text style from content to aiming to perform TST without any style-content disentanglement. While we postulate that the trend to perform TST without any style-content disentanglement will continue, we believe that the study on style representation remains an interesting research direction that deserves further exploration.

In addition to discussing common TST techniques, we also conducted a large-scale reproducibility study where we replicated and benchmarked 19 state-of-the-art TST algorithms on two publicly available datasets. \red{To the best of our knowledge, this is the first large-scale reproducibility study on TST methods.} The results of our study show that none of the TST methods could dominate on all evaluation metrics. This suggests the complexity of the TST task, where different methods may have advantages in different aspects, and there is no simple way to declare a winner. The evaluation analysis in our reproducibility study also advocated the need to search for better TST evaluation metrics.

We believe that research on TST will continue to flourish, and the industry will continue to find more exciting applications for the existing TST methods. We hope that this survey can provide readers with a comprehensive understanding of the critical aspects of this field, clarify the crucial types of TST methods, and shed some light on future studies.

%
\balance
\bibliographystyle{abbrv}
\bibliography{ref}  

\begin{thebibliography}{100}

\bibitem{abdul2015survey}
S.~A. Abdul-Kader and J.~Woods.
\newblock Survey on chatbot design techniques in speech conversation systems.
\newblock {\em International Journal of Advanced Computer Science and
  Applications}, 6(7), 2015.

\bibitem{ashok2013success}
V.~G. Ashok, S.~Feng, and Y.~Choi.
\newblock Success with style: Using writing style to predict the success of
  novels.
\newblock In {\em Proceedings of the 2013 conference on empirical methods in
  natural language processing}, pages 1753--1764, 2013.

\bibitem{DBLP:journals/corr/BahdanauCB14}
D.~Bahdanau, K.~Cho, and Y.~Bengio.
\newblock Neural machine translation by jointly learning to align and
  translate.
\newblock In Y.~Bengio and Y.~LeCun, editors, {\em 3rd International Conference
  on Learning Representations, {ICLR} 2015, San Diego, CA, USA, May 7-9, 2015,
  Conference Track Proceedings}, 2015.

\bibitem{bowman2016generating}
S.~Bowman, L.~Vilnis, O.~Vinyals, A.~Dai, R.~Jozefowicz, and S.~Bengio.
\newblock Generating sentences from a continuous space.
\newblock In {\em Proceedings of The 20th SIGNLL Conference on Computational
  Natural Language Learning}, pages 10--21, 2016.

\bibitem{brocardo2013authorship}
M.~L. Brocardo, I.~Traore, S.~Saad, and I.~Woungang.
\newblock Authorship verification for short messages using stylometry.
\newblock In {\em 2013 International Conference on Computer, Information and
  Telecommunication Systems (CITS)}, pages 1--6. IEEE, 2013.

\bibitem{brown1990statistical}
P.~F. Brown, J.~Cocke, S.~A. Della~Pietra, V.~J. Della~Pietra, F.~Jelinek,
  J.~Lafferty, R.~L. Mercer, and P.~S. Roossin.
\newblock A statistical approach to machine translation.
\newblock {\em Computational linguistics}, 16(2):79--85, 1990.

\bibitem{DBLP:conf/nips/BrownMRSKDNSSAA20}
T.~B. Brown, B.~Mann, N.~Ryder, M.~Subbiah, J.~Kaplan, P.~Dhariwal,
  A.~Neelakantan, P.~Shyam, G.~Sastry, A.~Askell, S.~Agarwal,
  A.~Herbert{-}Voss, G.~Krueger, T.~Henighan, R.~Child, A.~Ramesh, D.~M.
  Ziegler, J.~Wu, C.~Winter, C.~Hesse, M.~Chen, E.~Sigler, M.~Litwin, S.~Gray,
  B.~Chess, J.~Clark, C.~Berner, S.~McCandlish, A.~Radford, I.~Sutskever, and
  D.~Amodei.
\newblock Language models are few-shot learners.
\newblock In H.~Larochelle, M.~Ranzato, R.~Hadsell, M.~Balcan, and H.~Lin,
  editors, {\em Advances in Neural Information Processing Systems 33: Annual
  Conference on Neural Information Processing Systems}, 2020.

\bibitem{can2004change}
F.~Can and J.~M. Patton.
\newblock Change of writing style with time.
\newblock {\em Computers and the Humanities}, 38(1):61--82, 2004.

\bibitem{cao2020expertise}
Y.~Cao, R.~Shui, L.~Pan, M.-Y. Kan, Z.~Liu, and T.-S. Chua.
\newblock Expertise style transfer: A new task towards better communication
  between experts and laymen.
\newblock In {\em Proceedings of the 58th Annual Meeting of the Association for
  Computational Linguistics}, 2020.

\bibitem{carlson2018evaluating}
K.~Carlson, A.~Riddell, and D.~Rockmore.
\newblock Evaluating prose style transfer with the bible.
\newblock {\em Royal Society open science}, 5(10):171920, 2018.

\bibitem{celikyilmaz2020evaluation}
A.~Celikyilmaz, E.~Clark, and J.~Gao.
\newblock Evaluation of text generation: A survey.
\newblock {\em arXiv preprint arXiv:2006.14799}, 2020.

\bibitem{chambliss1996adults}
M.~J. Chambliss and R.~Garner.
\newblock Do adults change their minds after reading persuasive text?
\newblock {\em Written Communication}, 13(3):291--313, 1996.

\bibitem{chen2018adversarial}
L.~Chen, S.~Dai, C.~Tao, H.~Zhang, Z.~Gan, D.~Shen, Y.~Zhang, G.~Wang,
  R.~Zhang, and L.~Carin.
\newblock Adversarial text generation via feature-mover's distance.
\newblock In {\em Advances in Neural Information Processing Systems}, pages
  4666--4677, 2018.

\bibitem{chen2016variational}
X.~Chen, D.~P. Kingma, T.~Salimans, Y.~Duan, P.~Dhariwal, J.~Schulman,
  I.~Sutskever, and P.~Abbeel.
\newblock Variational lossy autoencoder.
\newblock {\em arXiv preprint arXiv:1611.02731}, 2016.

\bibitem{DBLP:conf/acl/ChengMSMZLC20}
P.~Cheng, M.~R. Min, D.~Shen, C.~Malon, Y.~Zhang, Y.~Li, and L.~Carin.
\newblock Improving disentangled text representation learning with
  information-theoretic guidance.
\newblock In D.~Jurafsky, J.~Chai, N.~Schluter, and J.~R. Tetreault, editors,
  {\em Proceedings of the 58th Annual Meeting of the Association for
  Computational Linguistics, {ACL} 2020, Online, July 5-10, 2020}, pages
  7530--7541. Association for Computational Linguistics, 2020.

\bibitem{cho2014learning}
K.~Cho, B.~van Merri{\"e}nboer, C.~Gulcehre, D.~Bahdanau, F.~Bougares,
  H.~Schwenk, and Y.~Bengio.
\newblock Learning phrase representations using rnn encoder--decoder for
  statistical machine translation.
\newblock In {\em Proceedings of the 2014 Conference on Empirical Methods in
  Natural Language Processing (EMNLP)}, pages 1724--1734, 2014.

\bibitem{dai2019style}
N.~Dai, J.~Liang, X.~Qiu, and X.-J. Huang.
\newblock Style transformer: Unpaired text style transfer without disentangled
  latent representation.
\newblock In {\em Proceedings of the 57th Annual Meeting of the Association for
  Computational Linguistics}, pages 5997--6007, 2019.

\bibitem{darani2014persuasive}
L.~H. Darani.
\newblock Persuasive style and its realization through transitivity analysis: A
  sfl perspective.
\newblock {\em Procedia-social and behavioral sciences}, 158:179--186, 2014.

\bibitem{DBLP:conf/iclr/DathathriMLHFMY20}
S.~Dathathri, A.~Madotto, J.~Lan, J.~Hung, E.~Frank, P.~Molino, J.~Yosinski,
  and R.~Liu.
\newblock Plug and play language models: {A} simple approach to controlled text
  generation.
\newblock In {\em 8th International Conference on Learning Representations,
  {ICLR} 2020, Addis Ababa, Ethiopia, April 26-30, 2020}. OpenReview.net, 2020.

\bibitem{devlin-etal-2019-bert}
J.~Devlin, M.-W. Chang, K.~Lee, and K.~Toutanova.
\newblock {BERT}: Pre-training of deep bidirectional transformers for language
  understanding.
\newblock In {\em Proceedings of the 2019 Conference of the North {A}merican
  Chapter of the Association for Computational Linguistics: Human Language
  Technologies, Volume 1 (Long and Short Papers)}, pages 4171--4186,
  Minneapolis, Minnesota, June 2019. Association for Computational Linguistics.

\bibitem{dos2018fighting}
C.~dos Santos, I.~Melnyk, and I.~Padhi.
\newblock Fighting offensive language on social media with unsupervised text
  style transfer.
\newblock In {\em Proceedings of the 56th Annual Meeting of the Association for
  Computational Linguistics (Volume 2: Short Papers)}, pages 189--194, 2018.

\bibitem{dziri-etal-2019-augmenting}
N.~Dziri, E.~Kamalloo, K.~Mathewson, and O.~Zaiane.
\newblock Augmenting neural response generation with context-aware topical
  attention.
\newblock In {\em Proceedings of the First Workshop on NLP for Conversational
  AI}, pages 18--31, Florence, Italy, Aug. 2019. Association for Computational
  Linguistics.

\bibitem{enkvist2016linguistic}
N.~E. Enkvist.
\newblock {\em Linguistic stylistics}, volume~5.
\newblock Walter de Gruyter GmbH \& Co KG, 2016.

\bibitem{feng2018topic}
X.~Feng, M.~Liu, J.~Liu, B.~Qin, Y.~Sun, and T.~Liu.
\newblock Topic-to-essay generation with neural networks.
\newblock In {\em IJCAI}, pages 4078--4084, 2018.

\bibitem{fortuna2018survey}
P.~Fortuna and S.~Nunes.
\newblock A survey on automatic detection of hate speech in text.
\newblock {\em ACM Computing Surveys (CSUR)}, 51(4):1--30, 2018.

\bibitem{NEURIPS2019_5e2b6675}
Y.~Fu, Y.~Feng, and J.~P. Cunningham.
\newblock Paraphrase generation with latent bag of words.
\newblock In H.~Wallach, H.~Larochelle, A.~Beygelzimer, F.~d\textquotesingle
  Alch\'{e}-Buc, E.~Fox, and R.~Garnett, editors, {\em Advances in Neural
  Information Processing Systems}, volume~32, pages 13645--13656. Curran
  Associates, Inc., 2019.

\bibitem{fu2018style}
Z.~Fu, X.~Tan, N.~Peng, D.~Zhao, and R.~Yan.
\newblock Style transfer in text: Exploration and evaluation.
\newblock In {\em Thirty-Second AAAI Conference on Artificial Intelligence},
  2018.

\bibitem{gatt2018survey}
A.~Gatt and E.~Krahmer.
\newblock Survey of the state of the art in natural language generation: Core
  tasks, applications and evaluation.
\newblock {\em Journal of Artificial Intelligence Research}, 61:65--170, 2018.

\bibitem{gatys2015neural}
L.~A. Gatys, A.~S. Ecker, and M.~Bethge.
\newblock A neural algorithm of artistic style.
\newblock {\em CoRR}, abs/1508.06576, 2015.

\bibitem{9257174}
A.~{Gidiotis} and G.~{Tsoumakas}.
\newblock A divide-and-conquer approach to the summarization of long documents.
\newblock {\em IEEE/ACM Transactions on Audio, Speech, and Language
  Processing}, 28:3029--3040, 2020.

\bibitem{gong2019reinforcement}
H.~Gong, S.~Bhat, L.~Wu, J.~Xiong, and W.-m. Hwu.
\newblock Reinforcement learning based text style transfer without parallel
  training corpus.
\newblock In {\em Proceedings of the 2019 Conference of the North American
  Chapter of the Association for Computational Linguistics: Human Language
  Technologies, Volume 1 (Long and Short Papers)}, pages 3168--3180, 2019.

\bibitem{goodfellow2020generative}
I.~Goodfellow, J.~Pouget-Abadie, M.~Mirza, B.~Xu, D.~Warde-Farley, S.~Ozair,
  A.~Courville, and Y.~Bengio.
\newblock Generative adversarial networks.
\newblock {\em Communications of the ACM}, 63(11):139--144, 2020.

\bibitem{DBLP:conf/nips/GoodfellowPMXWOCB14}
I.~J. Goodfellow, J.~Pouget{-}Abadie, M.~Mirza, B.~Xu, D.~Warde{-}Farley,
  S.~Ozair, A.~C. Courville, and Y.~Bengio.
\newblock Generative adversarial nets.
\newblock In Z.~Ghahramani, M.~Welling, C.~Cortes, N.~D. Lawrence, and K.~Q.
  Weinberger, editors, {\em Advances in Neural Information Processing Systems
  27: Annual Conference on Neural Information Processing Systems 2014, December
  8-13 2014, Montreal, Quebec, Canada}, pages 2672--2680, 2014.

\bibitem{halliday1981linguistic}
M.~Halliday.
\newblock Linguistic function and literary style: An inquiry into the language
  of william golding’s the inheritors.
\newblock {\em Essays in Modern Stylistics}, pages 325--60, 1981.

\bibitem{He2020A}
J.~He, X.~Wang, G.~Neubig, and T.~Berg-Kirkpatrick.
\newblock A probabilistic formulation of unsupervised text style transfer.
\newblock In {\em International Conference on Learning Representations (ICLR)},
  2020.

\bibitem{he2016ups}
R.~He and J.~McAuley.
\newblock Ups and downs: Modeling the visual evolution of fashion trends with
  one-class collaborative filtering.
\newblock In {\em proceedings of the 25th international conference on world
  wide web}, pages 507--517, 2016.

\bibitem{hovy2016social}
D.~Hovy and S.~L. Spruit.
\newblock The social impact of natural language processing.
\newblock In {\em Proceedings of the 54th Annual Meeting of the Association for
  Computational Linguistics (Volume 2: Short Papers)}, pages 591--598, 2016.

\bibitem{hovy1987generating}
E.~Hovy.
\newblock Generating natural language under pragmatic constraints.
\newblock {\em Journal of Pragmatics}, 11(6):689--719, 1987.

\bibitem{hovy-lin-1998-automated}
E.~Hovy and C.-Y. Lin.
\newblock Automated text summarization and the {S}ummarist system.
\newblock In {\em TIPSTER TEXT PROGRAM PHASE III: Proceedings of a Workshop
  held at Baltimore, {M}aryland, October 13-15, 1998}, pages 197--214,
  Baltimore, Maryland, USA, Oct. 1998. Association for Computational
  Linguistics.

\bibitem{hu2020deepstyle}
Z.~Hu, R.~K.-W. Lee, L.~Wang, E.-p. Lim, and B.~Dai.
\newblock Deepstyle: User style embedding for authorship attribution of short
  texts.
\newblock In {\em Asia-Pacific Web (APWeb) and Web-Age Information Management
  (WAIM) Joint International Conference on Web and Big Data}, pages 221--229.
  Springer, 2020.

\bibitem{hu2017toward}
Z.~Hu, Z.~Yang, X.~Liang, R.~Salakhutdinov, and E.~P. Xing.
\newblock Toward controlled generation of text.
\newblock In {\em Proceedings of the 34th International Conference on Machine
  Learning-Volume 70}, pages 1587--1596. JMLR. org, 2017.

\bibitem{hu2018unifying}
Z.~Hu, Z.~Yang, R.~Salakhutdinov, and E.~P. Xing.
\newblock On unifying deep generative models.
\newblock In {\em International Conference on Learning Representations}, 2018.

\bibitem{huang-etal-2016-visual}
T.-H.~K. Huang, F.~Ferraro, N.~Mostafazadeh, I.~Misra, A.~Agrawal, J.~Devlin,
  R.~Girshick, X.~He, P.~Kohli, D.~Batra, C.~L. Zitnick, D.~Parikh,
  L.~Vanderwende, M.~Galley, and M.~Mitchell.
\newblock Visual storytelling.
\newblock In {\em Proceedings of the 2016 Conference of the North {A}merican
  Chapter of the Association for Computational Linguistics: Human Language
  Technologies}, pages 1233--1239, San Diego, California, June 2016.
  Association for Computational Linguistics.

\bibitem{iqbal2010mining}
F.~Iqbal, H.~Binsalleeh, B.~C. Fung, and M.~Debbabi.
\newblock Mining writeprints from anonymous e-mails for forensic investigation.
\newblock {\em digital investigation}, 7(1-2):56--64, 2010.

\bibitem{ivanivc2004discourses}
R.~Ivani{\v{c}}.
\newblock Discourses of writing and learning to write.
\newblock {\em Language and education}, 18(3):220--245, 2004.

\bibitem{jain2019unsupervised}
P.~Jain, A.~Mishra, A.~P. Azad, and K.~Sankaranarayanan.
\newblock Unsupervised controllable text formalization.
\newblock In {\em Proceedings of the AAAI Conference on Artificial
  Intelligence}, volume~33, pages 6554--6561, 2019.

\bibitem{DBLP:conf/iclr/JangGP17}
E.~Jang, S.~Gu, and B.~Poole.
\newblock Categorical reparameterization with gumbel-softmax.
\newblock In {\em 5th International Conference on Learning Representations,
  {ICLR} 2017, Toulon, France, April 24-26, 2017, Conference Track
  Proceedings}. OpenReview.net, 2017.

\bibitem{jhamtani2017shakespearizing}
H.~Jhamtani, V.~Gangal, E.~Hovy, and E.~Nyberg.
\newblock Shakespearizing modern language using copy-enriched sequence to
  sequence models.
\newblock In {\em Proceedings of the Workshop on Stylistic Variation}, pages
  10--19, 2017.

\bibitem{jin2020hooks}
D.~Jin, Z.~Jin, J.~T. Zhou, L.~Orii, and P.~Szolovits.
\newblock Hooks in the headline: Learning to generate headlines with controlled
  styles.
\newblock In {\em Proceedings of the 58th Annual Meeting of the Association for
  Computational Linguistics}, 2020.

\bibitem{jin2019imat}
Z.~Jin, D.~Jin, J.~Mueller, N.~Matthews, and E.~Santus.
\newblock Imat: Unsupervised text attribute transfer via iterative matching and
  translation.
\newblock In {\em Proceedings of the 2019 Conference on Empirical Methods in
  Natural Language Processing and the 9th International Joint Conference on
  Natural Language Processing (EMNLP-IJCNLP)}, pages 3088--3100, 2019.

\bibitem{jing2019neural}
Y.~Jing, Y.~Yang, Z.~Feng, J.~Ye, Y.~Yu, and M.~Song.
\newblock Neural style transfer: A review.
\newblock {\em IEEE transactions on visualization and computer graphics}, 2019.

\bibitem{john2019disentangled}
V.~John, L.~Mou, H.~Bahuleyan, and O.~Vechtomova.
\newblock Disentangled representation learning for non-parallel text style
  transfer.
\newblock In {\em Proceedings of the 57th Annual Meeting of the Association for
  Computational Linguistics}, pages 424--434, 2019.

\bibitem{johnstone1989linguistic}
B.~Johnstone.
\newblock Linguistic strategies and cultural styles for persuasive discourse.
\newblock 1989.

\bibitem{johnstone2009stance}
B.~Johnstone.
\newblock Stance, style, and the linguistic individual.
\newblock {\em Stance: sociolinguistic perspectives}, pages 29--52, 2009.

\bibitem{kajiwara2016building}
T.~Kajiwara and M.~Komachi.
\newblock Building a monolingual parallel corpus for text simplification using
  sentence similarity based on alignment between word embeddings.
\newblock In {\em Proceedings of COLING 2016, the 26th International Conference
  on Computational Linguistics: Technical Papers}, pages 1147--1158, 2016.

\bibitem{kaptein2012adaptive}
M.~Kaptein, B.~De~Ruyter, P.~Markopoulos, and E.~Aarts.
\newblock Adaptive persuasive systems: a study of tailored persuasive text
  messages to reduce snacking.
\newblock {\em ACM Transactions on Interactive Intelligent Systems (TiiS)},
  2(2):1--25, 2012.

\bibitem{kaptein2015personalizing}
M.~Kaptein, P.~Markopoulos, B.~De~Ruyter, and E.~Aarts.
\newblock Personalizing persuasive technologies: Explicit and implicit
  personalization using persuasion profiles.
\newblock {\em International Journal of Human-Computer Studies}, 77:38--51,
  2015.

\bibitem{kim2019comparing}
S.~Kim, J.~Lee, and G.~Gweon.
\newblock Comparing data from chatbot and web surveys: Effects of platform and
  conversational style on survey response quality.
\newblock In {\em Proceedings of the 2019 CHI Conference on Human Factors in
  Computing Systems}, pages 1--12, 2019.

\bibitem{kingma2013auto}
D.~P. Kingma and M.~Welling.
\newblock Auto-encoding variational bayes.
\newblock {\em arXiv preprint arXiv:1312.6114}, 2013.

\bibitem{klahold2020computer}
A.~Klahold and M.~Fathi.
\newblock {\em Computer Aided Writing}.
\newblock Springer, 2020.

\bibitem{klahold2020word}
A.~Klahold and M.~Fathi.
\newblock Word processing as writing support.
\newblock In {\em Computer Aided Writing}, pages 21--29. Springer, 2020.

\bibitem{479394}
R.~{Kneser} and H.~{Ney}.
\newblock Improved backing-off for m-gram language modeling.
\newblock In {\em 1995 International Conference on Acoustics, Speech, and
  Signal Processing}, volume~1, pages 181--184 vol.1, 1995.

\bibitem{koehn2007moses}
P.~Koehn, H.~Hoang, A.~Birch, C.~Callison-Burch, M.~Federico, N.~Bertoldi,
  B.~Cowan, W.~Shen, C.~Moran, R.~Zens, et~al.
\newblock Moses: Open source toolkit for statistical machine translation.
\newblock In {\em Proceedings of the 45th annual meeting of the association for
  computational linguistics companion volume proceedings of the demo and poster
  sessions}, pages 177--180, 2007.

\bibitem{kong2019adversarial}
X.~Kong, B.~Li, G.~Neubig, E.~Hovy, and Y.~Yang.
\newblock An adversarial approach to high-quality, sentiment-controlled neural
  dialogue generation.
\newblock {\em arXiv preprint arXiv:1901.07129}, 2019.

\bibitem{lai2019multiple}
C.-T. Lai, Y.-T. Hong, H.-Y. Chen, C.-J. Lu, and S.-D. Lin.
\newblock Multiple text style transfer by using word-level conditional
  generative adversarial network with two-phase training.
\newblock In {\em Proceedings of the 2019 Conference on Empirical Methods in
  Natural Language Processing and the 9th International Joint Conference on
  Natural Language Processing (EMNLP-IJCNLP)}, pages 3570--3575, 2019.

\bibitem{lamb2016professor}
A.~M. Lamb, A.~G. ALIAS PARTH~GOYAL, Y.~Zhang, S.~Zhang, A.~C. Courville, and
  Y.~Bengio.
\newblock Professor forcing: A new algorithm for training recurrent networks.
\newblock {\em Advances in neural information processing systems},
  29:4601--4609, 2016.

\bibitem{lampleSSDRB19}
G.~Lample, S.~Subramanian, E.~M. Smith, L.~Denoyer, M.~Ranzato, and Y.~Boureau.
\newblock Multiple-attribute text rewriting.
\newblock In {\em 7th International Conference on Learning Representations,
  {ICLR} 2019, New Orleans, LA, USA, May 6-9, 2019}, 2019.

\bibitem{LeeftinkS19}
W.~Leeftink and G.~Spanakis.
\newblock Towards controlled transformation of sentiment in sentences.
\newblock In {\em Proceedings of the 11th International Conference on Agents
  and Artificial Intelligence, {ICAART} 2019, Volume 2, Prague, Czech Republic,
  February 19-21, 2019}, pages 809--816, 2019.

\bibitem{leidner2017ethical}
J.~L. Leidner and V.~Plachouras.
\newblock Ethical by design: Ethics best practices for natural language
  processing.
\newblock In {\em Proceedings of the First ACL Workshop on Ethics in Natural
  Language Processing}, pages 30--40, 2017.

\bibitem{li2019domain}
D.~Li, Y.~Zhang, Z.~Gan, Y.~Cheng, C.~Brockett, B.~Dolan, and M.-T. Sun.
\newblock Domain adaptive text style transfer.
\newblock In {\em Proceedings of the 2019 Conference on Empirical Methods in
  Natural Language Processing and the 9th International Joint Conference on
  Natural Language Processing (EMNLP-IJCNLP)}, pages 3295--3304, 2019.

\bibitem{li-etal-2016-persona}
J.~Li, M.~Galley, C.~Brockett, G.~Spithourakis, J.~Gao, and B.~Dolan.
\newblock A persona-based neural conversation model.
\newblock In {\em Proceedings of the 54th Annual Meeting of the Association for
  Computational Linguistics (Volume 1: Long Papers)}, pages 994--1003, Berlin,
  Germany, Aug. 2016. Association for Computational Linguistics.

\bibitem{li2018delete}
J.~Li, R.~Jia, H.~He, and P.~Liang.
\newblock Delete, retrieve, generate: a simple approach to sentiment and style
  transfer.
\newblock In {\em Proceedings of the 2018 Conference of the North American
  Chapter of the Association for Computational Linguistics: Human Language
  Technologies, Volume 1 (Long Papers)}, pages 1865--1874, 2018.

\bibitem{liao2018quase}
Y.~Liao, L.~Bing, P.~Li, S.~Shi, W.~Lam, and T.~Zhang.
\newblock Quase: Sequence editing under quantifiable guidance.
\newblock In {\em Proceedings of the 2018 Conference on Empirical Methods in
  Natural Language Processing}, pages 3855--3864, 2018.

\bibitem{liu2020revision}
D.~Liu, J.~Fu, Y.~Zhang, C.~Pal, and J.~Lv.
\newblock Revision in continuous space: Fine-grained control of text style
  transfer.
\newblock 2020.

\bibitem{liu-etal-2020-impress}
Q.~Liu, Y.~Chen, B.~Chen, J.-G. Lou, Z.~Chen, B.~Zhou, and D.~Zhang.
\newblock You impress me: Dialogue generation via mutual persona perception.
\newblock In {\em Proceedings of the 58th Annual Meeting of the Association for
  Computational Linguistics}, pages 1417--1427, Online, July 2020. Association
  for Computational Linguistics.

\bibitem{logeswaran2018content}
L.~Logeswaran, H.~Lee, and S.~Bengio.
\newblock Content preserving text generation with attribute controls.
\newblock In {\em Advances in Neural Information Processing Systems}, pages
  5103--5113, 2018.

\bibitem{luan-etal-2017-multi}
Y.~Luan, C.~Brockett, B.~Dolan, J.~Gao, and M.~Galley.
\newblock Multi-task learning for speaker-role adaptation in neural
  conversation models.
\newblock In {\em Proceedings of the Eighth International Joint Conference on
  Natural Language Processing (Volume 1: Long Papers)}, pages 605--614, Taipei,
  Taiwan, Nov. 2017. Asian Federation of Natural Language Processing.

\bibitem{luo2019dual}
F.~Luo, P.~Li, J.~Zhou, P.~Yang, B.~Chang, X.~Sun, and Z.~Sui.
\newblock A dual reinforcement learning framework for unsupervised text style
  transfer.
\newblock In {\em Proceedings of the 28th International Joint Conference on
  Artificial Intelligence}, pages 5116--5122. AAAI Press, 2019.

\bibitem{ma-etal-2018-query}
S.~Ma, X.~Sun, W.~Li, S.~Li, W.~Li, and X.~Ren.
\newblock Query and output: Generating words by querying distributed word
  representations for paraphrase generation.
\newblock In {\em Proceedings of the 2018 Conference of the North {A}merican
  Chapter of the Association for Computational Linguistics: Human Language
  Technologies, Volume 1 (Long Papers)}, pages 196--206, New Orleans,
  Louisiana, June 2018. Association for Computational Linguistics.

\bibitem{maas2011learning}
A.~L. Maas, R.~E. Daly, P.~T. Pham, D.~Huang, A.~Y. Ng, and C.~Potts.
\newblock Learning word vectors for sentiment analysis.
\newblock In {\em Proceedings of the 49th annual meeting of the association for
  computational linguistics: Human language technologies-volume 1}, pages
  142--150. Association for Computational Linguistics, 2011.

\bibitem{macarthur2009reflections}
C.~A. MacArthur.
\newblock Reflections on research on writing and technology for struggling
  writers.
\newblock {\em Learning Disabilities Research \& Practice}, 24(2):93--103,
  2009.

\bibitem{mcdonald-pustejovsky-1985-computational}
D.~D. McDonald and J.~D. Pustejovsky.
\newblock A computational theory of prose style for natural language
  generation.
\newblock In {\em Second Conference of the {E}uropean Chapter of the
  Association for Computational Linguistics}, Geneva, Switzerland, Mar. 1985.
  Association for Computational Linguistics.

\bibitem{miao2016language}
Y.~Miao and P.~Blunsom.
\newblock Language as a latent variable: Discrete generative models for
  sentence compression.
\newblock In {\em Proceedings of the 2016 Conference on Empirical Methods in
  Natural Language Processing}, pages 319--328, 2016.

\bibitem{mir2019evaluating}
R.~Mir, B.~Felbo, N.~Obradovich, and I.~Rahwan.
\newblock Evaluating style transfer for text.
\newblock In {\em Proceedings of the 2019 Conference of the North American
  Chapter of the Association for Computational Linguistics: Human Language
  Technologies, Volume 1 (Long and Short Papers)}, pages 495--504, 2019.

\bibitem{mizukami2015linguistic}
M.~Mizukami, G.~Neubig, S.~Sakti, T.~Toda, and S.~Nakamura.
\newblock Linguistic individuality transformation for spoken language.
\newblock In {\em Natural Language Dialog Systems and Intelligent Assistants},
  pages 129--143. Springer, 2015.

\bibitem{DBLP:conf/emnlp/2014}
A.~Moschitti, B.~Pang, and W.~Daelemans, editors.
\newblock {\em Proceedings of the 2014 Conference on Empirical Methods in
  Natural Language Processing, {EMNLP} 2014, October 25-29, 2014, Doha, Qatar,
  {A} meeting of SIGDAT, a Special Interest Group of the {ACL}}. {ACL}, 2014.

\bibitem{muehlenhaus2012if}
I.~Muehlenhaus.
\newblock If looks could kill: The impact of different rhetorical styles on
  persuasive geocommunication.
\newblock {\em The Cartographic Journal}, 49(4):361--375, 2012.

\bibitem{mueller2017sequence}
J.~Mueller, D.~Gifford, and T.~Jaakkola.
\newblock Sequence to better sequence: continuous revision of combinatorial
  structures.
\newblock In {\em Proceedings of the 34th International Conference on Machine
  Learning-Volume 70}, pages 2536--2544, 2017.

\bibitem{nikolov2018large}
N.~I. Nikolov and R.~H.~R. Hahnloser.
\newblock Large-scale hierarchical alignment for author style transfer.
\newblock {\em CoRR}, abs/1810.08237, 2018.

\bibitem{nishida2019multi}
K.~Nishida, I.~Saito, K.~Nishida, K.~Shinoda, A.~Otsuka, H.~Asano, and
  J.~Tomita.
\newblock Multi-style generative reading comprehension.
\newblock In {\em Proceedings of the 57th Annual Meeting of the Association for
  Computational Linguistics}, pages 2273--2284, 2019.

\bibitem{nisioi2017exploring}
S.~Nisioi, S.~{\v{S}}tajner, S.~P. Ponzetto, and L.~P. Dinu.
\newblock Exploring neural text simplification models.
\newblock In {\em Proceedings of the 55th Annual Meeting of the Association for
  Computational Linguistics (Volume 2: Short Papers)}, pages 85--91, 2017.

\bibitem{pang2019daunting}
R.~Y. Pang.
\newblock The daunting task of real-world textual style transfer
  auto-evaluation.
\newblock {\em CoRR}, abs/1910.03747, 2019.

\bibitem{pang2019towards}
R.~Y. Pang.
\newblock Towards actual (not operational) textual style transfer
  auto-evaluation.
\newblock In {\em Proceedings of the 5th Workshop on Noisy User-generated Text
  (W-NUT 2019)}, pages 444--445, 2019.

\bibitem{pang2019unsupervised}
R.~Y. Pang and K.~Gimpel.
\newblock Unsupervised evaluation metrics and learning criteria for
  non-parallel textual transfer.
\newblock In {\em Proceedings of the 3rd Workshop on Neural Generation and
  Translation}, pages 138--147, 2019.

\bibitem{papineni2002bleu}
K.~Papineni, S.~Roukos, T.~Ward, and W.-J. Zhu.
\newblock Bleu: a method for automatic evaluation of machine translation.
\newblock In {\em Proceedings of the 40th annual meeting on association for
  computational linguistics}, pages 311--318. Association for Computational
  Linguistics, 2002.

\bibitem{park2019paraphrase}
S.~Park, S.-w. Hwang, F.~Chen, J.~Choo, J.-W. Ha, S.~Kim, and J.~Yim.
\newblock Paraphrase diversification using counterfactual debiasing.
\newblock In {\em Proceedings of the AAAI Conference on Artificial
  Intelligence}, volume~33, pages 6883--6891, 2019.

\bibitem{parra2019automated}
G.~Parra et~al.
\newblock Automated writing evaluation tools in the improvement of the writing
  skill.
\newblock {\em International Journal of Instruction}, 12(2):209--226, 2019.

\bibitem{prabhumoye2018style}
S.~Prabhumoye, Y.~Tsvetkov, R.~Salakhutdinov, and A.~W. Black.
\newblock Style transfer through back-translation.
\newblock In {\em Proceedings of the 56th Annual Meeting of the Association for
  Computational Linguistics (Volume 1: Long Papers)}, pages 866--876, 2018.

\bibitem{rabinovich-etal-2017-personalized}
E.~Rabinovich, R.~N. Patel, S.~Mirkin, L.~Specia, and S.~Wintner.
\newblock Personalized machine translation: Preserving original author traits.
\newblock In {\em Proceedings of the 15th Conference of the {E}uropean Chapter
  of the Association for Computational Linguistics: Volume 1, Long Papers},
  pages 1074--1084, Valencia, Spain, Apr. 2017. Association for Computational
  Linguistics.

\bibitem{radford2017learning}
A.~Radford, R.~Jozefowicz, and I.~Sutskever.
\newblock Learning to generate reviews and discovering sentiment.
\newblock {\em arXiv preprint arXiv:1704.01444}, 2017.

\bibitem{radford2018improving}
A.~Radford, K.~Narasimhan, T.~Salimans, and I.~Sutskever.
\newblock Improving language understanding by generative pre-training.
\newblock 2018.

\bibitem{radford2019language}
A.~Radford, J.~Wu, R.~Child, D.~Luan, D.~Amodei, and I.~Sutskever.
\newblock Language models are unsupervised multitask learners.
\newblock {\em OpenAI Blog}, 1(8):9, 2019.

\bibitem{rao2018dear}
S.~Rao and J.~Tetreault.
\newblock Dear sir or madam, may i introduce the gyafc dataset: Corpus,
  benchmarks and metrics for formality style transfer.
\newblock In {\em Proceedings of the 2018 Conference of the North American
  Chapter of the Association for Computational Linguistics: Human Language
  Technologies, Volume 1 (Long Papers)}, pages 129--140, 2018.

\bibitem{reddy2016obfuscating}
S.~Reddy and K.~Knight.
\newblock Obfuscating gender in social media writing.
\newblock In {\em Proceedings of the First Workshop on NLP and Computational
  Social Science}, pages 17--26, 2016.

\bibitem{rubner1998metric}
Y.~Rubner, C.~Tomasi, and L.~J. Guibas.
\newblock A metric for distributions with applications to image databases.
\newblock In {\em Sixth International Conference on Computer Vision (IEEE Cat.
  No. 98CH36271)}, pages 59--66. IEEE, 1998.

\bibitem{saggion2017automatic}
H.~Saggion.
\newblock Automatic text simplification.
\newblock {\em Synthesis Lectures on Human Language Technologies},
  10(1):1--137, 2017.

\bibitem{salawu2017approaches}
S.~Salawu, Y.~He, and J.~Lumsden.
\newblock Approaches to automated detection of cyberbullying: A survey.
\newblock {\em IEEE Transactions on Affective Computing}, 11(1):3--24, 2017.

\bibitem{sennrich2015improving}
R.~Sennrich, B.~Haddow, and A.~Birch.
\newblock Improving neural machine translation models with monolingual data.
\newblock In {\em Proceedings of the 54th Annual Meeting of the Association for
  Computational Linguistics (Volume 1: Long Papers)}, pages 86--96, 2016.

\bibitem{serban2016building}
I.~Serban, A.~Sordoni, Y.~Bengio, A.~Courville, and J.~Pineau.
\newblock Building end-to-end dialogue systems using generative hierarchical
  neural network models.
\newblock In {\em Proceedings of the AAAI Conference on Artificial
  Intelligence}, volume~30, 2016.

\bibitem{DBLP:conf/aaai/SerbanKTTZBC17}
I.~V. Serban, T.~Klinger, G.~Tesauro, K.~Talamadupula, B.~Zhou, Y.~Bengio, and
  A.~C. Courville.
\newblock Multiresolution recurrent neural networks: An application to dialogue
  response generation.
\newblock In S.~P. Singh and S.~Markovitch, editors, {\em Proceedings of the
  Thirty-First {AAAI} Conference on Artificial Intelligence, February 4-9,
  2017, San Francisco, California, {USA}}, pages 3288--3294. {AAAI} Press,
  2017.

\bibitem{shang2019semi}
M.~Shang, P.~Li, Z.~Fu, L.~Bing, D.~Zhao, S.~Shi, and R.~Yan.
\newblock Semi-supervised text style transfer: Cross projection in latent
  space.
\newblock In {\em Proceedings of the 2019 Conference on Empirical Methods in
  Natural Language Processing and the 9th International Joint Conference on
  Natural Language Processing (EMNLP-IJCNLP)}, pages 4939--4948, 2019.

\bibitem{shen2017style}
T.~Shen, T.~Lei, R.~Barzilay, and T.~Jaakkola.
\newblock Style transfer from non-parallel text by cross-alignment.
\newblock In {\em Advances in neural information processing systems}, pages
  6830--6841, 2017.

\bibitem{shen2020educating}
T.~Shen, J.~Mueller, R.~Barzilay, and T.~Jaakkola.
\newblock Educating text autoencoders: Latent representation guidance via
  denoising.
\newblock In {\em International Conference on Machine Learning}, pages
  8719--8729. PMLR, 2020.

\bibitem{silva2019computational}
P.~C.~D. Silva, R.~L.~P. Teixeira, and V.~O. A.~V. Boas.
\newblock Computational linguistics: Analysis of the functional use of
  microsoft text word processor text corrector.
\newblock {\em International Journal of Linguistics, Literature and Culture,
  LLC}, page~23, 2019.

\bibitem{snyder1993writing}
I.~Snyder.
\newblock Writing with word processors: a research overview.
\newblock {\em Educational Research}, 35(1):49--68, 1993.

\bibitem{sordoni-etal-2015-neural}
A.~Sordoni, M.~Galley, M.~Auli, C.~Brockett, Y.~Ji, M.~Mitchell, J.-Y. Nie,
  J.~Gao, and B.~Dolan.
\newblock A neural network approach to context-sensitive generation of
  conversational responses.
\newblock In {\em Proceedings of the 2015 Conference of the North {A}merican
  Chapter of the Association for Computational Linguistics: Human Language
  Technologies}, pages 196--205, Denver, Colorado, May{--}June 2015.
  Association for Computational Linguistics.

\bibitem{sudhakar2019transforming}
A.~Sudhakar, B.~Upadhyay, and A.~Maheswaran.
\newblock “transforming” delete, retrieve, generate approach for controlled
  text style transfer.
\newblock In {\em Proceedings of the 2019 Conference on Empirical Methods in
  Natural Language Processing and the 9th International Joint Conference on
  Natural Language Processing (EMNLP-IJCNLP)}, pages 3260--3270, 2019.

\bibitem{DBLP:journals/inffus/SunLWLT20}
X.~Sun, J.~Li, X.~Wei, C.~Li, and J.~Tao.
\newblock Emotional editing constraint conversation content generation based on
  reinforcement learning.
\newblock {\em Inf. Fusion}, 56:70--80, 2020.

\bibitem{sutskever2014sequence}
I.~Sutskever, O.~Vinyals, and Q.~V. Le.
\newblock Sequence to sequence learning with neural networks.
\newblock In {\em Advances in neural information processing systems}, pages
  3104--3112, 2014.

\bibitem{syed2020adapting}
B.~Syed, G.~Verma, B.~V. Srinivasan, A.~Natarajan, and V.~Varma.
\newblock Adapting language models for non-parallel author-stylized rewriting.
\newblock In {\em Proceedings of the AAAI Conference on Artificial
  Intelligence}, volume~34, pages 9008--9015, 2020.

\bibitem{takase-okazaki-2019-positional}
S.~Takase and N.~Okazaki.
\newblock Positional encoding to control output sequence length.
\newblock In {\em Proceedings of the 2019 Conference of the North {A}merican
  Chapter of the Association for Computational Linguistics: Human Language
  Technologies, Volume 1 (Long and Short Papers)}, pages 3999--4004,
  Minneapolis, Minnesota, June 2019. Association for Computational Linguistics.

\bibitem{tian2018Structured}
Y.~Tian, Z.~Hu, and Z.~Yu.
\newblock Structured content preservation for unsupervised text style transfer.
\newblock {\em CoRR}, abs/1810.06526, 2018.

\bibitem{vaswani2017attention}
A.~Vaswani, N.~Shazeer, N.~Parmar, J.~Uszkoreit, L.~Jones, A.~N. Gomez,
  {\L}.~Kaiser, and I.~Polosukhin.
\newblock Attention is all you need.
\newblock In {\em Advances in neural information processing systems}, pages
  5998--6008, 2017.

\bibitem{vincent2010stacked}
P.~Vincent, H.~Larochelle, I.~Lajoie, Y.~Bengio, and P.-A. Manzagol.
\newblock Stacked denoising autoencoders: Learning useful representations in a
  deep network with a local denoising criterion.
\newblock {\em Journal of machine learning research}, 11(Dec):3371--3408, 2010.

\bibitem{vinyals2015pointer}
O.~Vinyals, M.~Fortunato, and N.~Jaitly.
\newblock Pointer networks.
\newblock In {\em Advances in neural information processing systems}, pages
  2692--2700, 2015.

\bibitem{voita-etal-2018-context}
E.~Voita, P.~Serdyukov, R.~Sennrich, and I.~Titov.
\newblock Context-aware neural machine translation learns anaphora resolution.
\newblock In {\em Proceedings of the 56th Annual Meeting of the Association for
  Computational Linguistics (Volume 1: Long Papers)}, pages 1264--1274,
  Melbourne, Australia, July 2018. Association for Computational Linguistics.

\bibitem{wang2019controllable}
K.~Wang, H.~Hua, and X.~Wan.
\newblock Controllable unsupervised text attribute transfer via editing
  entangled latent representation.
\newblock In {\em Advances in Neural Information Processing Systems}, pages
  11034--11044, 2019.

\bibitem{ijcai2018-619}
L.~Wang, J.~Yao, Y.~Tao, L.~Zhong, W.~Liu, and Q.~Du.
\newblock A reinforced topic-aware convolutional sequence-to-sequence model for
  abstractive text summarization.
\newblock In {\em Proceedings of the Twenty-Seventh International Joint
  Conference on Artificial Intelligence, {IJCAI-18}}, pages 4453--4460.
  International Joint Conferences on Artificial Intelligence Organization, 7
  2018.

\bibitem{wang-etal-2018-metrics}
X.~Wang, W.~Chen, Y.-F. Wang, and W.~Y. Wang.
\newblock No metrics are perfect: Adversarial reward learning for visual
  storytelling.
\newblock In {\em Proceedings of the 56th Annual Meeting of the Association for
  Computational Linguistics (Volume 1: Long Papers)}, pages 899--909,
  Melbourne, Australia, July 2018. Association for Computational Linguistics.

\bibitem{wang2019harnessing}
Y.~Wang, Y.~Wu, L.~Mou, Z.~Li, and W.~Chao.
\newblock Harnessing pre-trained neural networks with rules for formality style
  transfer.
\newblock In {\em Proceedings of the 2019 Conference on Empirical Methods in
  Natural Language Processing and the 9th International Joint Conference on
  Natural Language Processing (EMNLP-IJCNLP)}, pages 3564--3569, 2019.

\bibitem{williams1992simple}
R.~J. Williams.
\newblock Simple statistical gradient-following algorithms for connectionist
  reinforcement learning.
\newblock {\em Machine learning}, 8(3-4):229--256, 1992.

\bibitem{wu2019hierarchical}
C.~Wu, X.~Ren, F.~Luo, and X.~Sun.
\newblock A hierarchical reinforced sequence operation method for unsupervised
  text style transfer.
\newblock In {\em Proceedings of the 57th Annual Meeting of the Association for
  Computational Linguistics}, pages 4873--4883, 2019.

\bibitem{wu2019mask}
X.~Wu, T.~Zhang, L.~Zang, J.~Han, and S.~Hu.
\newblock Mask and infill: applying masked language model to sentiment
  transfer.
\newblock In {\em Proceedings of the 28th International Joint Conference on
  Artificial Intelligence}, pages 5271--5277. AAAI Press, 2019.

\bibitem{xing2017topic}
C.~Xing, W.~Wu, Y.~Wu, J.~Liu, Y.~Huang, M.~Zhou, and W.-Y. Ma.
\newblock Topic aware neural response generation.
\newblock In {\em Proceedings of the AAAI Conference on Artificial
  Intelligence}, volume~31, 2017.

\bibitem{xing2018hierarchical}
C.~Xing, Y.~Wu, W.~Wu, Y.~Huang, and M.~Zhou.
\newblock Hierarchical recurrent attention network for response generation.
\newblock In {\em Proceedings of the AAAI Conference on Artificial
  Intelligence}, volume~32, 2018.

\bibitem{xu2017new}
A.~Xu, Z.~Liu, Y.~Guo, V.~Sinha, and R.~Akkiraju.
\newblock A new chatbot for customer service on social media.
\newblock In {\em Proceedings of the 2017 CHI Conference on Human Factors in
  Computing Systems}, pages 3506--3510, 2017.

\bibitem{xu2018unpaired}
J.~Xu, X.~Sun, Q.~Zeng, X.~Zhang, X.~Ren, H.~Wang, and W.~Li.
\newblock Unpaired sentiment-to-sentiment translation: A cycled reinforcement
  learning approach.
\newblock In {\em Proceedings of the 56th Annual Meeting of the Association for
  Computational Linguistics (Volume 1: Long Papers)}, pages 979--988, 2018.

\bibitem{xu2019variational}
P.~Xu, Y.~Cao, and J.~C.~K. Cheung.
\newblock On variational learning of controllable representations for text
  without supervision.
\newblock {\em CoRR}, abs/1905.11975, 2019.

\bibitem{xu2020variational}
P.~Xu, J.~C.~K. Cheung, and Y.~Cao.
\newblock On variational learning of controllable representations for text
  without supervision.
\newblock In {\em International Conference on Machine Learning}, pages
  10534--10543. PMLR, 2020.

\bibitem{xu2019formality}
R.~Xu, T.~Ge, and F.~Wei.
\newblock Formality style transfer with hybrid textual annotations.
\newblock {\em CoRR}, abs/1903.06353, 2019.

\bibitem{xu2012paraphrasing}
W.~Xu, A.~Ritter, B.~Dolan, R.~Grishman, and C.~Cherry.
\newblock Paraphrasing for style.
\newblock In {\em Proceedings of COLING 2012}, pages 2899--2914, 2012.

\bibitem{yang2018investigating}
M.~Yang, Q.~Qu, K.~Lei, J.~Zhu, Z.~Zhao, X.~Chen, and J.~Z. Huang.
\newblock Investigating deep reinforcement learning techniques in personalized
  dialogue generation.
\newblock In {\em Proceedings of the 2018 SIAM International Conference on Data
  Mining}, pages 630--638. SIAM, 2018.

\bibitem{yang2017personalized}
M.~Yang, Z.~Zhao, W.~Zhao, X.~Chen, J.~Zhu, L.~Zhou, and Z.~Cao.
\newblock Personalized response generation via domain adaptation.
\newblock In {\em Proceedings of the 40th International ACM SIGIR Conference on
  Research and Development in Information Retrieval}, pages 1021--1024, 2017.

\bibitem{yang2018unsupervised}
Z.~Yang, Z.~Hu, C.~Dyer, E.~P. Xing, and T.~Berg-Kirkpatrick.
\newblock Unsupervised text style transfer using language models as
  discriminators.
\newblock In {\em Advances in Neural Information Processing Systems}, pages
  7287--7298, 2018.

\bibitem{yin2019utilizing}
D.~Yin, S.~Huang, X.-Y. Dai, and J.~Chen.
\newblock Utilizing non-parallel text for style transfer by making partial
  comparisons.
\newblock In {\em Proceedings of the 28th International Joint Conference on
  Artificial Intelligence}, pages 5379--5386. AAAI Press, 2019.

\bibitem{young2002technical}
M.~Young.
\newblock {\em The technical writer's handbook: writing with style and
  clarity}.
\newblock University Science Books, 2002.

\bibitem{zhang2018shaped}
Y.~Zhang, N.~Ding, and R.~Soricut.
\newblock Shaped: Shared-private encoder-decoder for text style adaptation.
\newblock In {\em Proceedings of NAACL-HLT}, pages 1528--1538, 2018.

\bibitem{zhang2020parallel}
Y.~Zhang, T.~Ge, and X.~Sun.
\newblock Parallel data augmentation for formality style transfer.
\newblock In {\em Proceedings of the 58th Annual Meeting of the Association for
  Computational Linguistics}, 2020.

\bibitem{zhang2018learning}
Y.~Zhang, J.~Xu, P.~Yang, and X.~Sun.
\newblock Learning sentiment memories for sentiment modification without
  parallel data.
\newblock In {\em Proceedings of the 2018 Conference on Empirical Methods in
  Natural Language Processing}, pages 1103--1108, 2018.

\bibitem{zhang2019machine}
Z.~Zhang, S.~Ren, S.~Liu, J.~Wang, P.~Chen, M.~Li, M.~Zhou, and E.~Chen.
\newblock Style transfer as unsupervised machine translation.
\newblock {\em CoRR}, abs/1808.07894, 2018.

\bibitem{zhao2018adversarially}
J.~Zhao, Y.~Kim, K.~Zhang, A.~M. Rush, and Y.~LeCun.
\newblock Adversarially regularized autoencoders.
\newblock In {\em 35th International Conference on Machine Learning, ICML
  2018}, pages 9405--9420. International Machine Learning Society (IMLS), 2018.

\bibitem{zhao2018language}
Y.~Zhao, W.~Bi, D.~Cai, X.~Liu, K.~Tu, and S.~Shi.
\newblock Language style transfer from sentences with arbitrary unknown styles.
\newblock {\em CoRR}, abs/1808.04071, 2018.

\bibitem{zhou2020exploring}
C.~Zhou, L.~Chen, J.~Liu, X.~Xiao, J.~Su, S.~Guo, and H.~Wu.
\newblock Exploring contextual word-level style relevance for unsupervised
  style transfer.
\newblock In {\em Proceedings of the 58th Annual Meeting of the Association for
  Computational Linguistics}, 2020.

\bibitem{zhou2018emotional}
H.~Zhou, M.~Huang, T.~Zhang, X.~Zhu, and B.~Liu.
\newblock Emotional chatting machine: Emotional conversation generation with
  internal and external memory.
\newblock In {\em Proceedings of the AAAI Conference on Artificial
  Intelligence}, volume~32, 2018.

\bibitem{zhou2020design}
L.~Zhou, J.~Gao, D.~Li, and H.-Y. Shum.
\newblock The design and implementation of xiaoice, an empathetic social
  chatbot.
\newblock {\em Computational Linguistics}, 46(1):53--93, 2020.

\bibitem{zhu2018texygen}
Y.~Zhu, S.~Lu, L.~Zheng, J.~Guo, W.~Zhang, J.~Wang, and Y.~Yu.
\newblock Texygen: A benchmarking platform for text generation models.
\newblock In {\em The 41st International ACM SIGIR Conference on Research \&
  Development in Information Retrieval}, pages 1097--1100, 2018.

\end{thebibliography}
%
%
\clearpage
\nobalance
\appendix
\section{Generated Samples}
We picked 4 test samples, including positive to negative, negative to positive, formal to informal, and informal to formal to show the quality of the sentences generated from different models intuitively.

\begin{table}[h]
\small
\centering
\caption{Example output in Yelp datasets.}
\label{tbl:case_study_yelp}
\begin{tabular}{c|c|c}
\hline
Model& From negative to positive (Yelp) & From positive to negative (Yelp) \\
\hline
Source & Food was cold and lacking of flavors . & Excellent service , ood food , and nice atmosphere . \\
\hline
\orange{DeleteOnly} &They were great , with a wide variety of flavors . &I would give \_num\_ stars if i could . \\
\orange{Template}  &Fantastic of flavors and fabulous . & Excellent service , good food , and nice atmosphere . \\
\orange{Del\&Retri}  &Of a huge fan of flavors . & A waste of time and money ! \\
\orange{B-GST} &Food was bland and of no flavors . & Great service , great food , and great atmosphere . \\
\orange{G-GST} &The food was cold and lacking of flavor . & The service , nice food , and atmosphere . \\
\orange{PTO} &Food was fresh and delicious of flavors . & Horrible service , horrible food , and horrible atmosphere . \\
\orange{UST} &- & Slow service , good food , and nice atmosphere . \\
\orange{SMAE} &- & Poor service , poor food , and poor atmosphere . \\
\hline
\green{DRLST} &Food was delicious and fresh & Service was ok but the food was not bad\\
\green{BST} &Food is delicious , and the staff . & Very poor , and no service and nothing . \\
\green{CAAE} &Food was fresh and good . & However , food was good , and poor , poor atmosphere .\\
\green{ARAE} &Food was cold and lacking . & Excellent service , good food , nice atmosphere and nice atmosphere .\\
\green{Ctrl-Gen} &Food was nicely and engaging of flavors . & Terrible service , not food , and 80s atmosphere . \\
\green{Multi-Dec} &Food was prepared , vegetarian and service . &Disgusting waiter , both came and dry selection at ! \\
\green{Style-Emb} &Food was cold and authentic which has . &Excellent service , good food , and nice atmosphere . \\
\hline                                                                    
\blue{DualRL} &Food was fresh and great of flavors . &Terrible service , bad food , and rude atmosphere . \\
\blue{DAST} &Food was great and clean of flavors . & Poor service , horrible food , and bland failure . \\
\blue{DAST-C} &Food was delicious and selection of flavors . &Terrible service , bad food , and tasteless atmosphere . \\
\blue{PFST} &Food was fresh and exceptional of flavors . & Horrible service , bad food , and poor service .\\
\hline
\hline
\end{tabular}

\end{table}

\begin{table}[h]
\small
\centering
\caption{Example output of GYAFC datasets.}
\label{tbl:case_study_GYAFC}
\begin{tabular}{c|c|c}
\hline
& From informal to formal (GYAFC) & From formal to informal (GYAFC) \\
\hline
Source &I can onli say...women r complicated...  &I would estimate approximately three or four .\\
\hline
\orange{DeleteOnly}& I can you say ... women r you ... & I would you you three or four .\\
\orange{Template}  &In my opinion , onli say ... women to engage in complicated ... & I would estimate jus three or four .\\
\orange{Del\&Retri}  &I can you say women r you as well . & I would you you each other four . \\
\orange{B-GST} &I can onli say . . . & I would estimate at least three . \\
\orange{G-GST} &I am not complicated . . . & I would estimate approximately three or four . \\
\orange{UST} &I don't think it is wrong. & I am a girl.\\
\orange{SMAE} &I can do anything at all. & I would not do it for you.\\
\hline
\green{DRLST} &I am not attracted to women &I would've been in a long hair \\
\green{BST} &If it is not a complicated ... & If i have him and ou\\
\green{CAAE} &I would have & I would dont cheat out with her\\
\green{ARAE} &I think that is a $<$unk$>$ & I am not sure that i am not.\\
\green{Ctrl-Gen} &I can secretly recall remain excited. &I would shouldnt quit three or guys! \\
\green{Multi-Dec} &I can say ... women r ... &I was me of guy . \\
\green{Style-Emb} &I can , i don ' t very ! &I would or two of him . \\
\hline                                                                    
\blue{DualRL} &I can $<$unk$>$ say women women complicated . &I would $<$unk$>$ thrilled ... \\
\blue{DAST} &I can $<$unk$>$ $<$unk$>$ family $<$unk$>$ & I would didn't in super or bs\\
\blue{DAST-C} &I can equally equally provide equally &I would luv 1 :) or honey \\
\blue{PFST} &I can $<$unk$>$ & I would $<$unk$>$ 6 or 6 $<$unk$>$\\
\hline
\hline
\end{tabular}

\end{table}

\end{document}